\documentclass[11pt]{article}

\usepackage[preprint]{tmlr}

\usepackage{amsmath,amssymb,amsfonts,amsthm,mathtools}
\usepackage{booktabs}
\usepackage{multirow}
\usepackage{graphicx}
\usepackage{subcaption}
\usepackage{enumitem}
\usepackage{microtype}
\usepackage{url}
\usepackage{hyperref}
\usepackage{natbib}
\usepackage{placeins}

\newcommand{\btheta}{\boldsymbol{\theta}}

\newcommand{\bx}{\boldsymbol{x}}
\newcommand{\bX}{\boldsymbol{X}}

\newcommand{\bY}{\boldsymbol{Y}}

\title{Gaussian Process-based learning with new MCMC-based implementation of Wishart prior on correlation matrix}
\author{
\name Kane Warrior \email gtq520@york.ac.uk \\
\addr Department of Mathematics \\
University of York
\AND
\name Dalia Chakrabarty \email dalia.chakrabarty@york.ac.uk \\
\addr Department of Mathematics \\
University of York
}
\begin{document}

\maketitle
\begin{abstract}
  
In probabilstic supervised learning of an input-output relationship - as a sample function of a Gaussian Process (GP) - priors are typically specified for the hyperparameters of the kernel that parametrises the covariance function of the GP, where the induced covariance matrix of the (resulting multivariate Normal) likelihood, governs the learning and prediction. When the sought function is highly multivariate, multiple lengthscale parameters must be learnt simultaneously, making inference difficult. We develop a ``self-assembled'' Wishart prior for the covariance matrix, while undertaking Bayesian inference on the kernel hyperparameters using MCMC. The construction uses a look-back window over recent MCMC iterations to define a time-step dependent scale matrix, thereby introducing adaptiveness to the chain. Results suggest that direct prior specification on the covariance matrix can be useful for diagnosing weakly informative inputs within the GP-based learning paradigm. We support our prior development with two distinct empirical illustrations - one on synthetic data, and another on a real-world dataset.

\end{abstract}

\section{Introduction}

Gaussian Process (GP) models are widely used as probabilistic models for nonlinear functions because they combine flexible function modelling with uncertainty quantification \citep{rasmussen2006gaussian,williams1996gaussian,mackay1994bayesian,neal1996bayesian}. Their predictive performance depends heavily on how kernel hyperparameters are learnt \citep{sundararajan2001predictive}. This becomes especially important in higher-dimensional multivariate settings, where many input-specific hyperparameters may be present and where only some inputs may contribute meaningful predictive structure \citep{mackay1994bayesian,neal1996bayesian,rasmussen2006gaussian,linkletter2006variable,paananen2019variable}.

In standard Bayesian formulations of GP learning, prior specification is usually imposed directly on kernel hyperparameters such as lengthscales, amplitude parameters, and noise terms \citep{rasmussen2006gaussian,williams1996gaussian}. This is natural from a modelling point of view, but it does not always give useful control over the covariance structure that those hyperparameters induce over the observed design points \citep{barnard2000modeling,gelman2006variancepriors,daniels1999nonconjugate,huang2013simple}. However, it is this induced covariance matrix that directly governs likelihood evaluation, numerical stability, and predictive behaviour \citep{rasmussen2006gaussian,stein1999interpolation}.

This distinction matters because proposed hyperparameter values can induce covariance matrices that are poorly conditioned, nearly singular, or otherwise difficult to work with numerically \citep{rasmussen2006gaussian,stein1999interpolation,ranjan2011computationally,gramacy2012cases}. Such difficulties become more pronounced when many hyperparameters must be learnt simultaneously \citep{mackay1994bayesian,neal1996bayesian,rasmussen2006gaussian} and when some inputs are only weakly informative or effectively irrelevant \citep{linkletter2006variable,paananen2019variable}. In these settings, a prior placed only on individual hyperparameters need not provide transparent control over the induced covariance structures that actually determine inference and prediction. This motivates a covariance-level perspective, in line with broader Bayesian work on prior specification for covariance matrices \citep{barnard2000modeling,daniels1999nonconjugate,huang2013simple}.

This motivates placing prior structure directly on the kernel-induced covariance matrix. Rather than increasing the flexibility of the kernel itself, we focus here on whether covariance-level prior information can improve GP hyperparameter learning in higher-dimensional multivariate settings. The idea is that, even when prior information about individual hyperparameters is weak, it may still be reasonable to constrain the covariance matrices they induce \citep{barnard2000modeling,daniels1999nonconjugate,huang2013simple}. This perspective is complementary to approaches that increase the flexibility of the covariance model itself, for example through nonstationary covariance constructions \citep{paciorek2006nonstationary} or richer kernel families \citep{wilson2013gaussian,duvenaud2013structure}.

In this paper, we develop a self-assembled Wishart prior that acts on the kernel-induced covariance matrix during Bayesian hyperparameter inference. At each iteration of the inference procedure, the current hyperparameter proposal induces a candidate covariance matrix, and a Wishart prior factor is evaluated directly on this matrix. The construction is \emph{self-assembled} in the sense that its scale matrix is not fixed in advance, but is instead built from a look-back window of recently accepted hyperparameter values. This produces an iteration-dependent covariance-level prior structure while retaining the kernel-parametric form of the model. Although related to the broader Wishart-process literature \citep{wilson2011generalised}, the construction is used here in a different role: not to define a stochastic process on covariance matrices over an external index set, but to introduce covariance-level prior structure into GP hyperparameter learning.

The contribution of the paper is both methodological and empirical. Methodologically, we formulate this covariance-level Wishart construction and show how it enters an MCMC inference scheme for GP hyperparameters. Empirically, we study its behaviour in sparse nonlinear synthetic problems with automatic relevance determination (ARD), and then in a real-data application to the T\'etouan City energy data. Across the synthetic experiments, the Wishart prior behaves very differently from priors placed only on individual hyperparameters: its main effect is not to make relevant-input lengthscales dramatically smaller, but to push irrelevant-input lengthscales much larger. This yields much sharper separation between relevant and nuisance dimensions, so that weakly informative inputs exert far less influence on the fitted predictor; in the settings considered here, this is accompanied by substantially better predictive accuracy.

The synthetic experiments also show that this behaviour is not explained by sparsity alone. The advantage of the Wishart prior varies across experiments with different numbers of relevant inputs and different signal structures, suggesting that the gain depends both on the amount of nuisance variation and on the complexity of the underlying covariance geometry. The real-data results complement this by showing that the same covariance-level behaviour can help identify an input that is only weakly informative under the assumed smooth GP structure, and that removing this input leads to a more effective reduced-input model. Taken together, the results support the view that prior specification directly on the induced covariance matrix can be a useful alternative to hyperparameter-level priors in higher-dimensional GP learning.

The main contributions of the paper are as follows:
\begin{enumerate}[leftmargin=1.5em]
    \item We develop a self-assembled Wishart prior that acts directly on the kernel-induced covariance matrix during Bayesian GP hyperparameter inference.
    \item We show how this prior can be incorporated into a Metropolis Hastings learning scheme while retaining a kernel-parametric covariance structure.
    \item We demonstrate in sparse nonlinear synthetic experiments that this covariance-level prior can yield much stronger relevance separation and substantially better prediction than priors placed only on individual hyperparameters.
    \item We show in a real-data application that the same covariance-level behaviour can help diagnose a weakly informative input and support a more effective reduced-input GP model.
\end{enumerate}

\section{Background and motivation}

\subsection{GP models with multi-input kernels}

The basic learning task here is the learning of $f(\cdot)$, where $Y = f(\bx)$, where the output variable $Y=y\in\mathbb{R}$, with the input $\bX = \mathbf{x}\in\mathcal{X}\subset\mathbb{R}^d$. We learn the random function $f(\cdot)$ using the training set ${\bf D} := \{(\mathbf{x}_i,y_i)\}_{i=1}^p$. 

With the random function $f(\cdot)$ modelled as a sample function of a zero-mean GP, the correlation function of which is parametrised with the kernel $k(\cdot,\cdot;\boldsymbol{\theta})$, the joint density of a finite number of realisations of $f(\cdot)$ is multivariate Normal. Then $\bY := \{Y_1=f(\bx_1),\ldots,Y_p=f(\bx_p)\}$ is such that (s.t.)
\[
\mathbf{Y}\mid \boldsymbol{\theta}\sim \mathcal{N}\!\bigl(\mathbf{0},\Sigma_{\boldsymbol{\theta}}\bigr),
\]
where
\[
\Sigma_{\boldsymbol{\theta}}
=
\bigl[k(\mathbf{x}_i,\mathbf{x}_j;\boldsymbol{\theta})\bigr]_{i,j=1}^p
\]
is the kernel-induced covariance matrix at the observed design points \citep{rasmussen2006gaussian,williams1996gaussian}.

In multi-input GP models, the kernel hyperparameter vector $\boldsymbol{\theta}$ often contains one or more input-specific parameters for each component of the input. A standard example is the squared exponential kernel with Automatic Relevance Determination (ARD),
\[
k(\mathbf{x},\mathbf{x}';\boldsymbol{\ell})
=
\exp\!\left(
-\frac{1}{2}
\sum_{q=1}^d \frac{(x_q-x'_q)^2}{\ell_q^2}
\right),
\]
where $\boldsymbol{\ell}=(\ell_1,\ldots,\ell_d)^\top$ is the vector of input-specific lengthscales. Under ARD, each input dimension is assigned its own lengthscale, so the covariance can respond differently to changes in different components of the input. These lengthscales are only interpretable relative to the scaling of the inputs; after standardisation of the inputs, a smaller learnt lengthscale means that the covariance decays more rapidly as that input changes, while a larger learnt lengthscale means that the model varies more slowly with respect to that input \citep{mackay1994bayesian,neal1996bayesian,rasmussen2006gaussian,paananen2019variable}.

ARD is useful in multi-input GP models because it permits input-specific lengthscales and hence a richer covariance structure than a single shared scale parameter. The resulting learning problem is then more demanding, since the number of hyperparameters grows with the input dimension and some of the associated inputs may be only weakly informative for prediction. In such settings, the learning problem is not only one of fitting a flexible nonlinear function, but also one of determining which inputs should retain influence in the covariance structure and which should be effectively smoothed out \citep{linkletter2006variable,paananen2019variable}. This is precisely the regime in which covariance structure becomes harder to control and posterior learning can become fragile.

\subsection{Why place prior structure on the covariance matrix?}

Standard Bayesian GP learning places a prior directly on the kernel hyperparameters and targets a posterior of the form
\[
\pi(\boldsymbol{\theta}\mid\mathcal{D})
\propto
\mathcal{L}(\boldsymbol{\theta};\mathcal{D})\,\pi_0(\boldsymbol{\theta}),
\]
where $\mathcal{L}(\boldsymbol{\theta};\mathcal{D})$ is the GP likelihood and $\pi_0(\boldsymbol{\theta})$ is a prior on the hyperparameters. This is natural from a modelling perspective, since the kernel is parameterised through $\boldsymbol{\theta}$ rather than directly through $\Sigma_{\boldsymbol{\theta}}$.

However, the object through which the likelihood is evaluated; predictions are computed; and numerical stability is determined, is the covariance matrix of the multivariate Normal likelihood that is induced by the GP \citep{rasmussen2006gaussian,stein1999interpolation}. Proposed hyperparameter values can lead to covariance matrices that are poorly conditioned, nearly singular, or difficult to handle numerically \citep{rasmussen2006gaussian,ranjan2011computationally,gramacy2012cases}. These difficulties can become more pronounced when many hyperparameters must be learnt simultaneously, and when some inputs are weakly informative or effectively irrelevant. In such cases, a prior placed only on an individual hyperparameter is typically weak, and does not provide transparent control over the covariance matrix that actually determines inference and prediction. This motivates a covariance matrix-level perspective, in line with broader Bayesian work on prior specification for covariance matrices \citep{barnard2000modeling,daniels1999nonconjugate,huang2013simple}. We therefore place a prior on $\Sigma_{\boldsymbol{\theta}}$ -
viewed alternatively as a joint prior on all kernel hyperparameters. As indicated above, this helps when $f(\cdot)$ is highly multivariate, where prior information on individual kernel hyperparameters is typically weak, though it may still be reasonable to constrain the covariance matrix that they induce.

\subsection{Relation to Wishart-process modelling}

A natural family of priors for covariance matrices comes from the Wishart distribution and extensions \citep{wishart1928,dawid1981,guptaNagar1999matrixvariate,anderson2003,muirhead1982}. In particular, \citet{wilson2011generalised} introduced the Generalised Wishart Process (GWP), realisations of which are positive semi-definite matrices indexed by an index such as time. This provides a flexible framework for modelling covariance matrices that vary over an index set.

Although our construction is closely related to Wishart-process modelling, the role of the Wishart prior here is different.
We remain within the GP-based model with a kernelised covariance function, and place the Wishart prior directly on the covariance matrix of the likelihood induced by the GP.
Since the covariance matrix is not the primary stochastic object of interest for us, (but the matrix through which the kernel hyperparameters determine likelihood evaluation and prediction), we aim to introduce the Wishart prior on the hyperparameter-induced covariance matrix, towards hyperparameter inference. Unlike \citet{wilson2011generalised}, we do not aim to build a general covariance-process model. 

\subsection{Why a self-assembled Wishart prior?}

A fixed Wishart or inverse-Wishart prior on \(\boldsymbol{\Sigma}_{\boldsymbol{\theta}}\) would require a fixed scale matrix chosen in advance, as in standard covariance-prior constructions for Bayesian multivariate models \citep{barnard2000modeling}. In some applications this may be reasonable, but in the GP-based approach within which we undertake hyperparameter learning, it is often difficult to specify a single covariance-level target {\it{a priori}}, that remains appropriate across the whole inference run, especially in highly multivariate ARD settings. The posterior structure may be highly uneven, and the covariance matrices induced by proposed kernel hyperparameters, may vary substantially across regions of the hyperparameter space.

For this reason, we use a \emph{self-assembled} Wishart construction that is implemented within the MCMC-based Bayesian inference that we undertake to learn the kernel hyperparameters given ${\bf D}$. At each iteration of the MCMC chain that we run to perform said inference, the scale matrix is built using hyperparameter values accepted during a ``look-back window'', i.e. an interval of time-steps in the immediate past of the current iteration. Hence, the resulting prior reflects covariance structures that the chain has recently regarded as plausible, rather than a single fixed covariance target chosen in advance. The Wishart prior is therefore assembled from the recent history of the inference itself.

The prior is then designed to keep the induced covariance matrices close to the covariance structures that the chain has recently regarded as plausible, while still allowing the GP to learn the relevant structure in the data. In highly multivariate ARD implementation, this provides a way to directly act on the covariance matrix, that in turn determines likelihood evaluation and prediction.

\section{A self-assembled Wishart prior on the induced covariance matrix}

This section formalises the proposed covariance-level prior construction. The aim is to place a Wishart prior directly on the covariance matrix induced by the kernel hyperparameters at the chosen design points, while constructing the Wishart scale matrix from hyperparameter values encountered over a look-back window in iteration space. The resulting prior remains kernel-parametric, but acts directly on the covariance structure through which kernel hyperparameters determine likelihood evaluation and prediction.

Let
\[
\boldsymbol{\theta}=(\theta_1,\ldots,\theta_m)^\top \in \Theta \subset \mathbb{R}^m
\]
denote the vector of kernel hyperparameters, where \(m\) is the total number of kernel hyperparameters in the model. For any \(\boldsymbol{\theta}\), let
\[
\boldsymbol{\Sigma}_{\boldsymbol{\theta}}
=
\bigl[k(\mathbf{x}_i,\mathbf{x}_j;\boldsymbol{\theta})\bigr]_{i,j=1}^p
\]
denote the \(p\times p\) covariance matrix induced by \(\boldsymbol{\theta}\) at the chosen design points \(\mathbf{x}_1,\ldots,\mathbf{x}_p\), where \(p\) is the size of the training set \({\bf D}\).

\subsection{Constructed scale matrix}

Bayesian inference on \(\boldsymbol{\theta}\) is undertaken using an MCMC chain of length \(N_{\mathrm{iter}}\). Let \(N_0\) denote the iteration at which the Wishart prior is first switched on, and assume that \(N_0>n\), so that a full look-back window of length \(n\) is available when the Wishart construction begins. For iterations \(t<N_0\), inference is driven only by the marginal priors on the individual hyperparameters together with the GP likelihood.

Now let \(t\in\{N_0,\ldots,N_{\mathrm{iter}}\}\). In the \(t\)-th iteration, let \(\boldsymbol{\theta}_t^{(*)}\) denote the proposed value of the hyperparameter vector, while \(\boldsymbol{\theta}_{t-1},\boldsymbol{\theta}_{t-2},\ldots\) denote previously accepted values. We define a look-back window of length \(n\) in iteration space, consisting of the indices
\[
\{t-n+1,\ldots,t-1,t\}.
\]
Within this window, the first \(n-1\) indices correspond to accepted hyperparameter values from the preceding iterations, while the final index \(t\) corresponds to the current proposal. Thus, in the \(t\)-th iteration, we use the collection
\[
\bigl\{\boldsymbol{\theta}_{t-n+1},\ldots,\boldsymbol{\theta}_{t-1},\boldsymbol{\theta}_t^{(*)}\bigr\}
\]
to define the mean hyperparameter vector
\begin{equation}
\bar{\boldsymbol{\theta}}_t
=
\frac{1}{n}
\left(
\boldsymbol{\theta}_t^{(*)}
+
\sum_{i=1}^{n-1}\boldsymbol{\theta}_{t-i}
\right).
\label{eq:lookback_mean}
\end{equation}

Using this mean, we define the constructed scale matrix
\begin{equation}
\mathbf{V}_t
=
\Bigl[
k(\mathbf{x}_j,\mathbf{x}_k;\bar{\boldsymbol{\theta}}_t)
\Bigr]_{j,k=1}^p.
\label{eq:scale_matrix}
\end{equation}

We also denote by
\begin{equation}
\boldsymbol{\Sigma}_t^{(*)}
=
\Bigl[
k(\mathbf{x}_j,\mathbf{x}_k;\boldsymbol{\theta}_t^{(*)})
\Bigr]_{j,k=1}^p
\label{eq:proposal_matrix}
\end{equation}
the covariance matrix induced by the proposed hyperparameter vector in the \(t\)-th iteration. Similarly, \(\boldsymbol{\Sigma}_t\) denotes the covariance matrix induced by the current hyperparameter vector in that iteration.

The matrix \(\mathbf{V}_t\) therefore serves as a constructed covariance-level reference matrix defined from the look-back window in iteration space.

\subsection{Wishart prior on the covariance matrix using the constructed scale matrix}

For each iteration \(t\in\{N_0,\ldots,N_{\mathrm{iter}}\}\), we evaluate a Wishart prior density at the proposed covariance matrix \(\boldsymbol{\Sigma}_t^{(*)}\), conditional on the constructed scale matrix \(\mathbf{V}_t\):
\begin{equation}
\pi_W\!\left(\boldsymbol{\Sigma}_t^{(*)}\mid \mathbf{V}_t\right)
=
\frac{
C\,\lvert \boldsymbol{\Sigma}_t^{(*)}\rvert^{\frac{n-p-1}{2}}
}{
\lvert \mathbf{V}_t\rvert^{\frac{n}{2}}
}
\exp\!\left(
-\frac{1}{2}\operatorname{tr}\!\left(\mathbf{V}_t^{-1}\boldsymbol{\Sigma}_t^{(*)}\right)
\right),
\label{eq:wishart_prior}
\end{equation}
where \(n\) is the Wishart degrees-of-freedom parameter, \(p\) is the dimension of the covariance matrix, and \(C\) is the Wishart normalising constant.

The two main terms in \eqref{eq:wishart_prior} play different roles. When \(n>p+1\), the determinant term penalises proposals approaching singularity through
\[
\lvert \boldsymbol{\Sigma}_t^{(*)}\rvert^{(n-p-1)/2}.
\]
The trace term
\[
\operatorname{tr}\!\left(\mathbf{V}_t^{-1}\boldsymbol{\Sigma}_t^{(*)}\right)
\]
penalises proposals whose induced covariance structure departs strongly from the reference covariance structure represented by \(\mathbf{V}_t\). Thus, for each \(t\in\{N_0,\ldots,N_{\mathrm{iter}}\}\), the Wishart prior acts directly on the covariance matrix induced by the proposed kernel hyperparameters in that iteration.

The prior does not replace kernel-based learning with an unconstrained covariance estimate. The covariance matrix remains fully determined by the kernel and the hyperparameter vector \(\boldsymbol{\theta}\). The role of the Wishart term is instead to impose covariance-level prior structure on the proposal-induced covariance matrix through the constructed scale matrix \(\mathbf{V}_t\).

\subsection{Posterior target and Metropolis Hastings update}

Let \(\mathcal{D}\) denote the observed data, and let
\[
\boldsymbol{\theta}=(\theta_1,\ldots,\theta_m)^\top
\]
denote the vector of kernel hyperparameters, where \(m\) is the total number of hyperparameters. Let \(\pi_0(\theta_k)\) denote the marginal prior on the \(k\)-th hyperparameter. Without the Wishart prior, the posterior target has the form
\[
\pi(\boldsymbol{\theta}\mid\mathcal{D})
\propto
\mathcal{L}(\boldsymbol{\theta};\mathcal{D})
\prod_{k=1}^{m}\pi_0(\theta_k).
\]

When the Wishart prior on the induced covariance matrix is included, the iteration-specific posterior target for \(t\in\{N_0,\ldots,N_{\mathrm{iter}}\}\) is
\begin{equation}
\tilde{\pi}_t(\boldsymbol{\theta}\mid\mathcal{D})
\propto
\mathcal{L}(\boldsymbol{\theta};\mathcal{D})\,
\Biggl[\prod_{k=1}^{m}\pi_0(\theta_k)\Biggr]\,
\pi_W\!\left(\boldsymbol{\Sigma}_{\boldsymbol{\theta}}\mid \mathbf{V}_t\right),
\label{eq:modified_target}
\end{equation}
where \(\boldsymbol{\Sigma}_{\boldsymbol{\theta}}\) is the covariance matrix induced by \(\boldsymbol{\theta}\), and \(\mathbf{V}_t\) is the constructed scale matrix in the \(t\)-th iteration.

At iteration \(t\), when the Wishart factor is included, the Metropolis Hastings acceptance probability for proposal \(\boldsymbol{\theta}_t^{(*)}\) is therefore
\begin{equation}
\alpha_t
=
\min\left\{
1,\;
\frac{
\mathcal{L}(\boldsymbol{\theta}_t^{(*)};\mathcal{D})\,
\Bigl[\prod_{k=1}^{m}\pi_0(\theta_{t,k}^{(*)})\Bigr]\,
\pi_W\!\left(\boldsymbol{\Sigma}_t^{(*)}\mid \mathbf{V}_t\right)
}{
\mathcal{L}(\boldsymbol{\theta}_{t-1};\mathcal{D})\,
\Bigl[\prod_{k=1}^{m}\pi_0(\theta_{t-1,k})\Bigr]\,
\pi_W\!\left(\boldsymbol{\Sigma}_{t-1}\mid \mathbf{V}_t\right)
}
\cdot
\frac{
q(\boldsymbol{\theta}_{t-1}\mid \boldsymbol{\theta}_t^{(*)})
}{
q(\boldsymbol{\theta}_t^{(*)}\mid \boldsymbol{\theta}_{t-1})
}
\right\},
\label{eq:mh_wishart}
\end{equation}
where \(q(\cdot\mid\cdot)\) is the proposal density. For symmetric random-walk proposals, the proposal ratio cancels. When the Wishart factor is not included, the corresponding Wishart ratio is omitted.

We consider the following specifications:
\begin{itemize}
    \item a \emph{Normal Prior} model, in which only the marginal prior
    \(\prod_{k=1}^{m}\pi_0(\theta_k)\) is used;
    \item a \emph{Wishart Prior} model, in which Uniform marginal priors are placed on the individual hyperparameters, and the Wishart prior
    \(\pi_W(\boldsymbol{\Sigma}_{\boldsymbol{\theta}}\mid \mathbf{V}_t)\) is included;
    \item a \emph{Wishart + Normal Prior} model, in which both
    \(\prod_{k=1}^{m}\pi_0(\theta_k)\) and the Wishart prior
    \(\pi_W(\boldsymbol{\Sigma}_{\boldsymbol{\theta}}\mid \mathbf{V}_t)\) are included.
\end{itemize}

\subsection{Adaptation}

For the iteration index \(t\in\{N_0,\ldots,N_{\mathrm{iter}}\}\), the
self-assembled Wishart prior introduces time dependence into the posterior
target through the constructed scale matrix \(\mathbf{V}_t\). Thus in the
\(t\)-th iteration, the posterior target is
\begin{equation}
\tilde{\pi}_t(\boldsymbol{\theta}\mid\mathcal{D})
\propto
\mathcal{L}(\boldsymbol{\theta};\mathcal{D})\,
\Biggl[\prod_{k=1}^{m}\pi_0(\theta_k)\Biggr]\,
\pi_W\!\left(\boldsymbol{\Sigma}_{\boldsymbol{\theta}}\mid \mathbf{V}_t\right),
\label{eq:adaptive_target}
\end{equation}
where \(\boldsymbol{\Sigma}_{\boldsymbol{\theta}}\) denotes the correlation
matrix induced by the kernel hyperparameter vector \(\boldsymbol{\theta}\), and
\(\mathbf{V}_t\) is the constructed scale matrix defined by collating the
hyperparameter values updated over the previous \(n\) iterations, i.e. over the
look-back window
\[
\boldsymbol{\theta}_{t-n+1},\ldots,
\boldsymbol{\theta}_{t-1},
\boldsymbol{\theta}_t^{(\star)}.
\]
Since \(\mathbf{V}_t\) varies with \(t\), the correlation structure of the
posterior target is iteration-dependent, and the MCMC chain used to learn
\(\boldsymbol{\theta}\) is therefore adaptive.

Thus, the posterior is rendered time-step dependent via the dependence of the Wishart prior $\pi_W\!\left(\boldsymbol{\Sigma}_{\boldsymbol{\theta}}\mid \mathbf{V}_t\right)$ on the dynamic scale matrix \(\mathbf{V}_t\). Hence, we realise that the adaptiveness of the chain is governed by \(\mathbf{V}_t\).

Now suppose that for some \(t_b\in\{N_0,\ldots,N_{\mathrm{iter}}\}\),
$\mathbf{V}_{t_b-1}$ approaches $\mathbf{V}_{t_b}$
in mean. Then the joint posterior density
$\tilde{\pi}_{t_b-1}(\boldsymbol{\theta}\mid\mathcal{D})$ computed at the hyperparameters that are current in the $t_b$-th iteration, (given the data ${\mathcal{D}}$), approaches
the posterior density 
$\tilde{\pi}_{t_b}(\boldsymbol{\theta}\mid\mathcal{D})$. Therefore $\theta_1, \ldots, \theta_m$ inferred upon under these two successive posterior targets, will approach each other in mean.
Hence, for \(\theta_k\), $\forall k\in\{1,\ldots,m\}$, the proposal generated in the \((t_b+1)\)-th iteration, conditional on the current value in the \(t_b\)-th iteration, is expected to approach the proposal generated in the \(t_b\)-th iteration conditional on the current value in the \((t_b-1)\)-th iteration, in mean. This in turn suggests that $\mathbf{V}_{t_b}$ approaches $\mathbf{V}_{t_b+1}$ in mean, (since $\mathbf{V}_t$ is defined using the hyperparameter values $\btheta_{t-n+1}, \ldots, \btheta_{t-1}, \btheta_t^{(\star)}$). 
Thus adaptation in the undertaken adaptive MCMC chain is diminishing.

At the same time, the state space of the undertaken MCMC chain is bound, since each hyperparameter is sampled from a Normal proposal that ensures that an accepted value of any hyperparameter is finite a.s.

From the above, it follows that our undertaken MCMC chain abides by diminishing adaptation and containment. Hence the chain converges \citep{roberts2007coupling}.

\subsection{Practical implementation choices}

Two practical choices determine how the self-assembled Wishart prior behaves. The first is the look-back window length \(n\), which controls how quickly the scale matrix \(\mathbf{V}_t\) responds to changes in recently accepted hyperparameter values. Larger values of \(n\) produce a more slowly varying reference covariance structure, while smaller values make \(\mathbf{V}_t\) more sensitive to local changes in the chain. In addition, for the Wishart density on \(p\times p\) covariance matrices to be non-degenerate, one requires \(n\ge p\).

The second is the iteration at which the Wishart prior is introduced within the chain. In early iterations, a full look-back window is not yet available, so in practice the Wishart term is switched on only once the required history has been formed. This ensures that the covariance-level prior is based on a complete recent window rather than on very early transient states.

Although \(n\ge p\) is sufficient for the Wishart density to be non-degenerate, in implementation we avoid the boundary choice \(n=p\). At this boundary, the determinant exponent \((n-p-1)/2\) is negative, so the determinant term places increasing weight on covariance matrices with very small determinants, corresponding to matrices close to singularity. Since near-singular covariance matrices are undesirable in GP likelihood evaluation and prediction, we use look-back windows with \(n>p\). In particular, \(n=p+1\) makes the determinant contribution neutral, while larger values of \(n\) increasingly penalise movement towards singularity.

\section{Synthetic experiments}

The synthetic study is designed to assess how the different priors behave when the underlying signal is nonlinear, embedded in a 15-dimensional input space, and accompanied by irrelevant inputs. The aim is practical: can the learning procedure distinguish relevant from irrelevant inputs, can it suppress nuisance inputs effectively, and does that improve prediction?

Throughout, each synthetic dataset is constructed in 15 input dimensions with 650 observations in total. Each input vector is drawn independently from
\[
[-1,1]^{15},
\]
and the full dataset is split randomly into 500 training observations and 150 test observations.

We compare the three prior specifications using root mean squared error (RMSE), mean absolute error (MAE), empirical 95\% predictive interval coverage, and a relevance ratio defined by
\[
\frac{\text{mean relevant lengthscale}}{\text{mean irrelevant lengthscale}}.
\]
Under ARD, after standardisation of inputs, smaller lengthscales indicate greater sensitivity to an input component, whereas larger lengthscales indicate heavier smoothing over that input. A relevance ratio below 1 therefore indicates that the model is assigning greater importance to the relevant inputs than to the irrelevant ones.

The synthetic study has one baseline experiment and four follow-up variants. The baseline problem has 6 relevant inputs and 9 nuisance inputs. The follow-up experiments then probe four related questions: the stability of the baseline pattern across repeated random seeds, behaviour in a much sparser additive setting with only 3 relevant inputs, behaviour in less sparse settings with 9 and 12 relevant inputs, and behaviour in a simplified 6-input signal with the same active inputs as the baseline but weaker interaction structure. Taken together, these experiments allow the effects of sparsity, nuisance-input suppression, and signal complexity to be examined separately.

\subsection{Synthetic designs}

\paragraph{Baseline 15-dimensional experiment.}
In the baseline setting, only 6 of the 15 inputs affect the response:
\[
X_1,\;X_3,\;X_5,\;X_8,\;X_{11},\;X_{14},
\]
while the remaining 9 inputs are nuisance variables. The latent signal is
\begin{align}
f(X)
&=
\sin(\pi X_1)
+0.8\cos(1.4\pi X_3)
+0.7\exp\!\bigl(-2.5(X_5+0.2)^2\bigr) \nonumber\\
&\quad
+0.6X_8X_{11}
-0.5X_{14}^2
+0.4\sin\!\bigl(\pi(X_1+X_{14})\bigr)
-0.35X_3X_8,
\end{align}
and the observed response is
\[
y=f(X)+\varepsilon,
\qquad
\varepsilon\sim\mathcal{N}(0,0.05^2).
\]
This makes the problem sparse, but also structurally nontrivial: the generator includes additive nonlinear terms, quadratic behaviour, and interaction effects.

\paragraph{Three relevant inputs.}
A much sparser additive variant uses only three relevant inputs,
\[
X_1,\;X_6,\;X_{12},
\]
with latent signal
\[
f(X)=1.0\sin(\pi X_1)+0.9\cos(1.4\pi X_6)+0.7\exp\!\bigl(-2.2(X_{12}+0.15)^2\bigr).
\]
This generator is structurally simpler than the baseline one, since it is purely additive and contains no interactions.

\paragraph{Nine relevant inputs.}
The next experiment makes the problem less sparse by increasing the number of relevant inputs to nine:
\[
X_1,\;X_2,\;X_4,\;X_5,\;X_7,\;X_9,\;X_{11},\;X_{13},\;X_{15}.
\]
This leaves only six nuisance inputs. The latent signal is
\begin{align}
f(X)
&=
0.9\sin(\pi X_1)
+0.7\cos(1.3\pi X_2)
+0.55\exp\!\bigl(-2.0(X_4-0.2)^2\bigr) \nonumber\\
&\quad
+0.45X_5
-0.4X_7^2
+0.35\sin(\pi X_9)
+0.3X_{11}X_{13}
+0.25\cos\!\bigl(\pi(X_{15}+X_1)\bigr).
\end{align}
This setting is less sparse than the baseline, but still contains mixed nonlinear effects, a quadratic term, and coupled structure.

\paragraph{Twelve relevant inputs.}
The least sparse setting considered here uses twelve relevant inputs,
\[
X_1,\;X_2,\;X_3,\;X_4,\;X_5,\;X_6,\;X_8,\;X_9,\;X_{10},\;X_{11},\;X_{13},\;X_{14},
\]
leaving only
\[
X_7,\;X_{12},\;X_{15}
\]
as nuisance variables. The latent signal is
\begin{align}
f(X)
&=
0.7\sin(\pi X_1)
+0.55\cos(1.2\pi X_2)
+0.45X_3
-0.35X_4^2 \nonumber\\
&\quad
+0.4\exp\!\bigl(-2.0(X_5+0.1)^2\bigr)
+0.35\sin(\pi X_6)
+0.3X_8X_9
+0.25\cos(\pi X_{10}) \nonumber\\
&\quad
-0.2X_{11}X_{13}
+0.2\sin\!\bigl(\pi(X_{14}+X_1)\bigr).
\end{align}
This is the least sparse version considered here, while still retaining mixed nonlinear structure and a small number of coupled terms.

\paragraph{Less interrelated six-input signal.}
To separate the effect of sparsity from the effect of signal complexity, we also considered a less interrelated variant of the baseline 6-input experiment, keeping the same active inputs
\[
X_1,\;X_3,\;X_5,\;X_8,\;X_{11},\;X_{14},
\]
but simplifying the signal structure relative to the original interaction-heavy case. The latent signal is
\[
f(X)
=
\sin(\pi X_1)
+0.8\cos(1.4\pi X_3)
+0.7\exp\!\bigl(-2.5(X_5+0.2)^2\bigr)
+0.5X_8
-0.45X_{11}^2
+0.35\sin(\pi X_{14}).
\]

\subsection{Summary results across synthetic experiments}

As an additional nonlinear prediction benchmark, Table~\ref{tab:synthetic_summary} also includes results from a fully connected feedforward deep neural network. The network used four hidden ReLU layers of widths \(128,128,64,\) and \(32\), followed by a single linear output unit, and was trained using AdamW with learning rate \(10^{-3}\), weight decay \(10^{-5}\), batch size \(256\), and \(2000\) epochs. Inputs and outputs were standardised using training-set statistics, and predictions were transformed back to the original response scale before evaluation. Unlike the GP experiments, which used \(500\) training observations and \(150\) test observations, the DNN benchmark was trained on \(2500\) observations and evaluated on a test set of \(750\) observations. Since the DNN is used here purely as a point-prediction benchmark, no entries are reported for coverage or relevance ratio.

Table~\ref{tab:synthetic_summary} summarises predictive performance and relevance separation across all synthetic experiments.

\begin{table}[!htbp]
\centering
\small
\begin{tabular}{llcccc}
\toprule
Experiment & Method & RMSE & MAE & 95\% coverage & Relevance ratio \\
\midrule
Baseline
& Normal & 0.289561 & 0.220585 & 0.973333 & 0.476766 \\
& Wishart & 0.136910 & 0.103927 & 0.920000 & 0.063423 \\
& Wishart + Normal & 0.289066 & 0.220156 & 0.966667 & 0.473727 \\
& DNN & 0.292010 & 0.230582 & -- & -- \\
\midrule
3 relevant
& Normal & 0.126915 & 0.098213 & 1.000000 & 0.281651 \\
& Wishart & 0.087020 & 0.073410 & 0.760000 & 0.046564 \\
& Wishart + Normal & 0.126673 & 0.098039 & 1.000000 & 0.280306 \\
& DNN & 0.190035 & 0.151494 & -- & -- \\
\midrule
9 relevant
& Normal & 0.297729 & 0.228554 & 0.993333 & 0.614073 \\
& Wishart & 0.188138 & 0.146355 & 0.926667 & 0.099666 \\
& Wishart + Normal & 0.296074 & 0.227007 & 0.993333 & 0.610118 \\
& DNN & 0.320970 & 0.254351 & -- & -- \\
\midrule
12 relevant
& Normal & 0.330038 & 0.262065 & 0.933333 & 0.690531 \\
& Wishart & 0.241785 & 0.191016 & 0.886667 & 0.149727 \\
& Wishart + Normal & 0.329965 & 0.262205 & 0.940000 & 0.690422 \\
& DNN & 0.275469 & 0.217273 & -- & -- \\
\midrule
Less interrelated 6-input
& Normal & 0.268220 & 0.210878 & 0.980000 & 0.494142 \\
& Wishart & 0.140665 & 0.104050 & 0.920000 & 0.089938 \\
& Wishart + Normal & 0.266795 & 0.209190 & 0.980000 & 0.497213 \\
& DNN & 0.289788 & 0.223757 & -- & -- \\
\bottomrule
\end{tabular}
\caption{Summary of predictive performance and, for the GP models, uncertainty coverage and relevance separation across the synthetic experiments. The DNN benchmark is included as a prediction-only comparator and therefore has no entries for coverage or relevance ratio.}
\label{tab:synthetic_summary}
\end{table}
\FloatBarrier

Several patterns are immediate. First, all three GP specifications recover some relevance structure, in the sense that all reported relevance ratios are below \(1\). The distinction is therefore not between complete failure and success, but between moderate and very strong separation. The Wishart prior produces much stronger separation than the other two GP specifications in every synthetic experiment considered here. This is the central mechanistic pattern in the synthetic study. Across all settings, the Wishart prior drives the relevance ratio down sharply relative to both the Normal prior and the Wishart + Normal prior, indicating that irrelevant inputs are being suppressed much more decisively.

Second, that stronger separation is reflected in prediction. Across the GP models, the Wishart prior gives the lowest RMSE and the lowest MAE in every synthetic experiment. The gain is particularly large in the baseline 6-input setting and in the less interrelated 6-input comparison, and remains clearly present in the 9- and 12-relevant-input experiments. The main trade-off is uncertainty calibration: in some settings, especially the 3-input additive case, improved point prediction is accompanied by lower empirical coverage.

Third, the DNN benchmark does not alter this picture. Despite being trained on a substantially larger dataset than the GP models, its predictive errors remain broadly comparable to those obtained under the Normal and Wishart + Normal GP priors, and do not match the stronger predictive performance achieved by the Wishart prior. In the baseline sparse setting, for example, the DNN RMSE is \(0.292010\), very close to the Normal and Wishart + Normal values of \(0.289561\) and \(0.289066\), whereas the Wishart prior attains RMSE \(0.136910\). A similar pattern appears in the less interrelated 6-input setting, and the same broad ranking persists in the 3-, 9-, and 12-relevant-input cases.

Taken together, Table~\ref{tab:synthetic_summary} suggests that the Wishart prior is helping primarily by suppressing nuisance inputs much more strongly than priors placed only on hyperparameters. The predictive advantage therefore appears to arise not from dramatically shrinking the relevant-input lengthscales relative to the competing methods, but from pushing irrelevant inputs far enough out that they exert much less influence on the fitted predictor. The DNN comparison strengthens this interpretation: a reasonably expressive nonlinear predictor trained on a larger sample still does not recover the same predictive gains, suggesting that the benefit of the Wishart construction is not simply a consequence of generic nonlinear flexibility, but is more specifically tied to covariance-level nuisance-input suppression.

\subsection{DNN permutation-importance diagnostics}

Table~\ref{tab:synthetic_summary} shows that the DNN benchmark attains predictive errors broadly comparable to those of the Normal and Wishart + Normal GP specifications, and does not recover the stronger predictive performance achieved by the Wishart prior. To examine whether this is due to a failure to identify the relevant inputs, we computed permutation-importance scores for the trained DNN in each synthetic experiment \citep{breiman2001random}.

\begin{figure}[!htbp]
    \centering
    \begin{subfigure}[t]{0.32\textwidth}
        \centering
        \includegraphics[width=\linewidth]{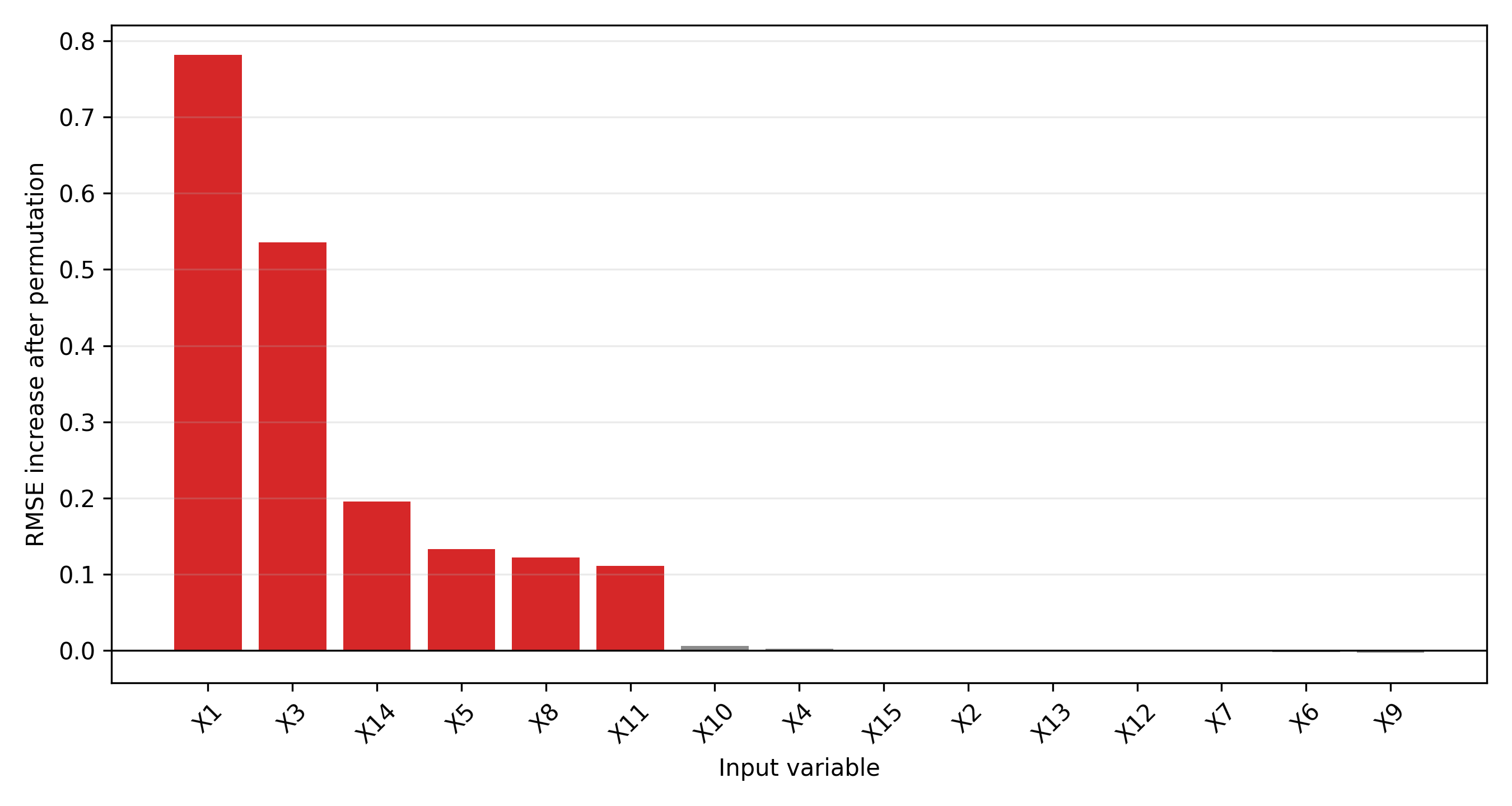}
        \caption{Baseline 6-relevant-input case.}
    \end{subfigure}\hfill
    \begin{subfigure}[t]{0.32\textwidth}
        \centering
        \includegraphics[width=\linewidth]{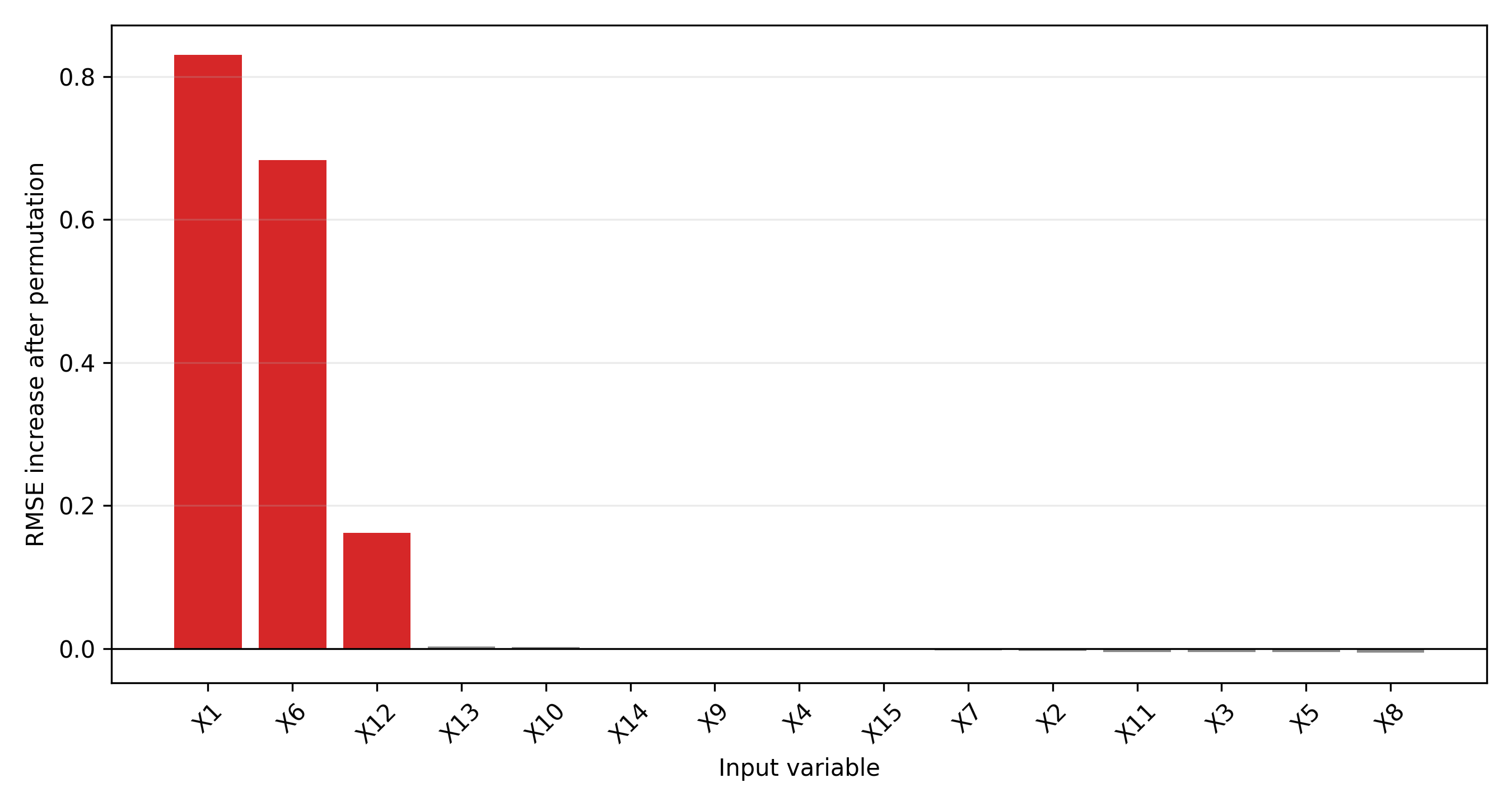}
        \caption{3 relevant inputs.}
    \end{subfigure}\hfill
    \begin{subfigure}[t]{0.32\textwidth}
        \centering
        \includegraphics[width=\linewidth]{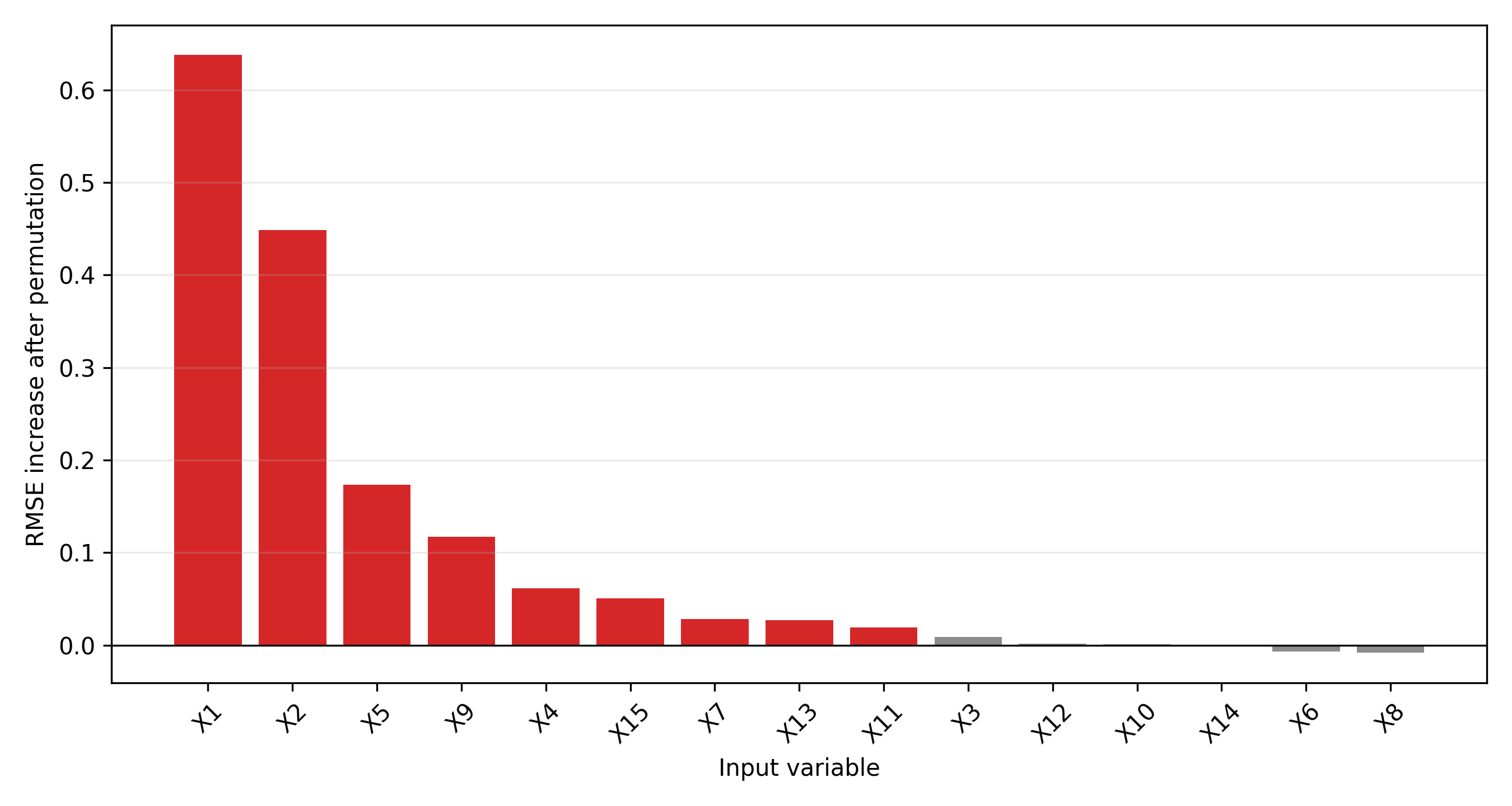}
        \caption{9 relevant inputs.}
    \end{subfigure}

    \vspace{0.8em}

    \begin{subfigure}[t]{0.32\textwidth}
        \centering
        \includegraphics[width=\linewidth]{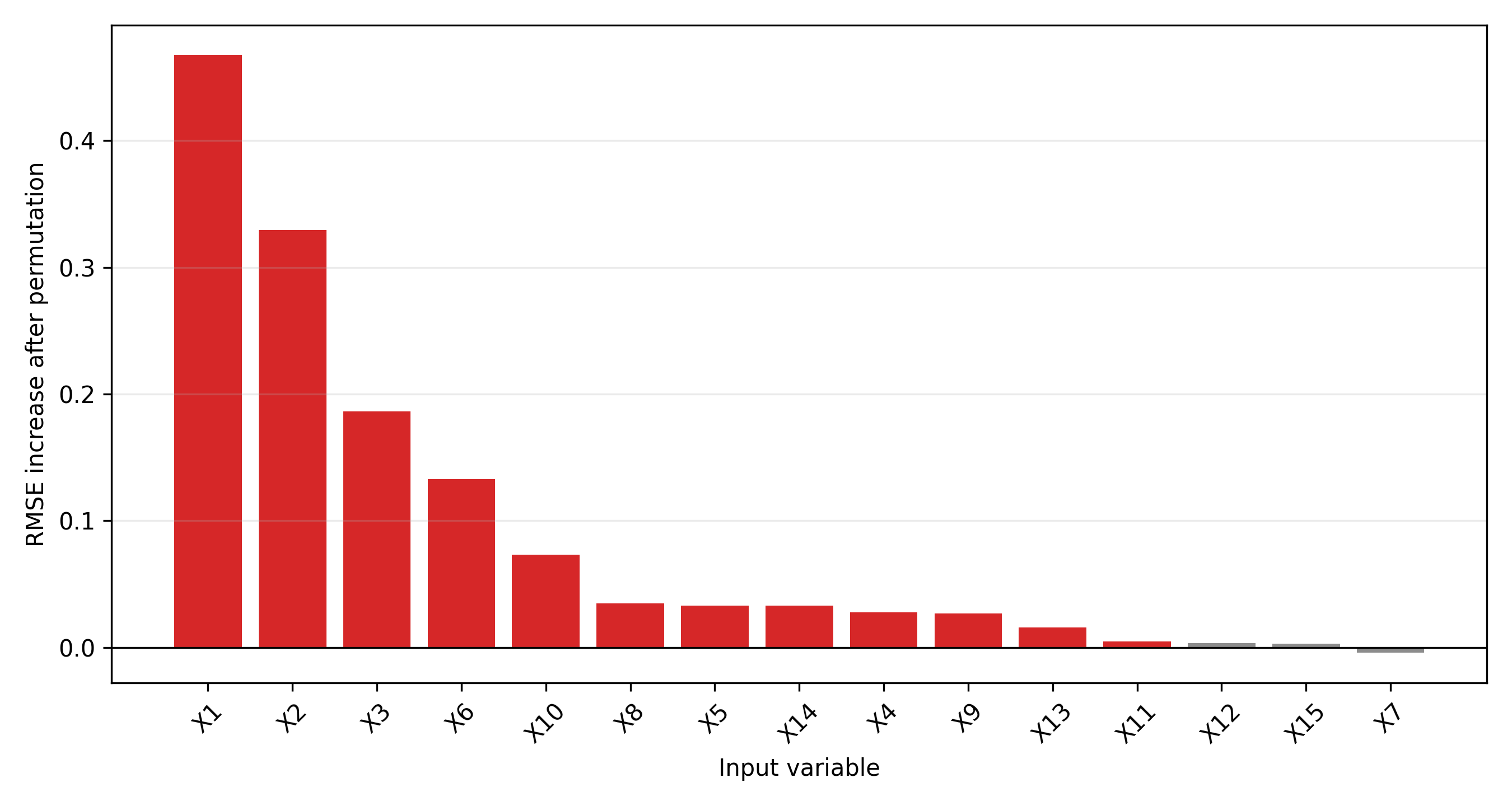}
        \caption{12 relevant inputs.}
    \end{subfigure}\hfill
    \begin{subfigure}[t]{0.32\textwidth}
        \centering
        \includegraphics[width=\linewidth]{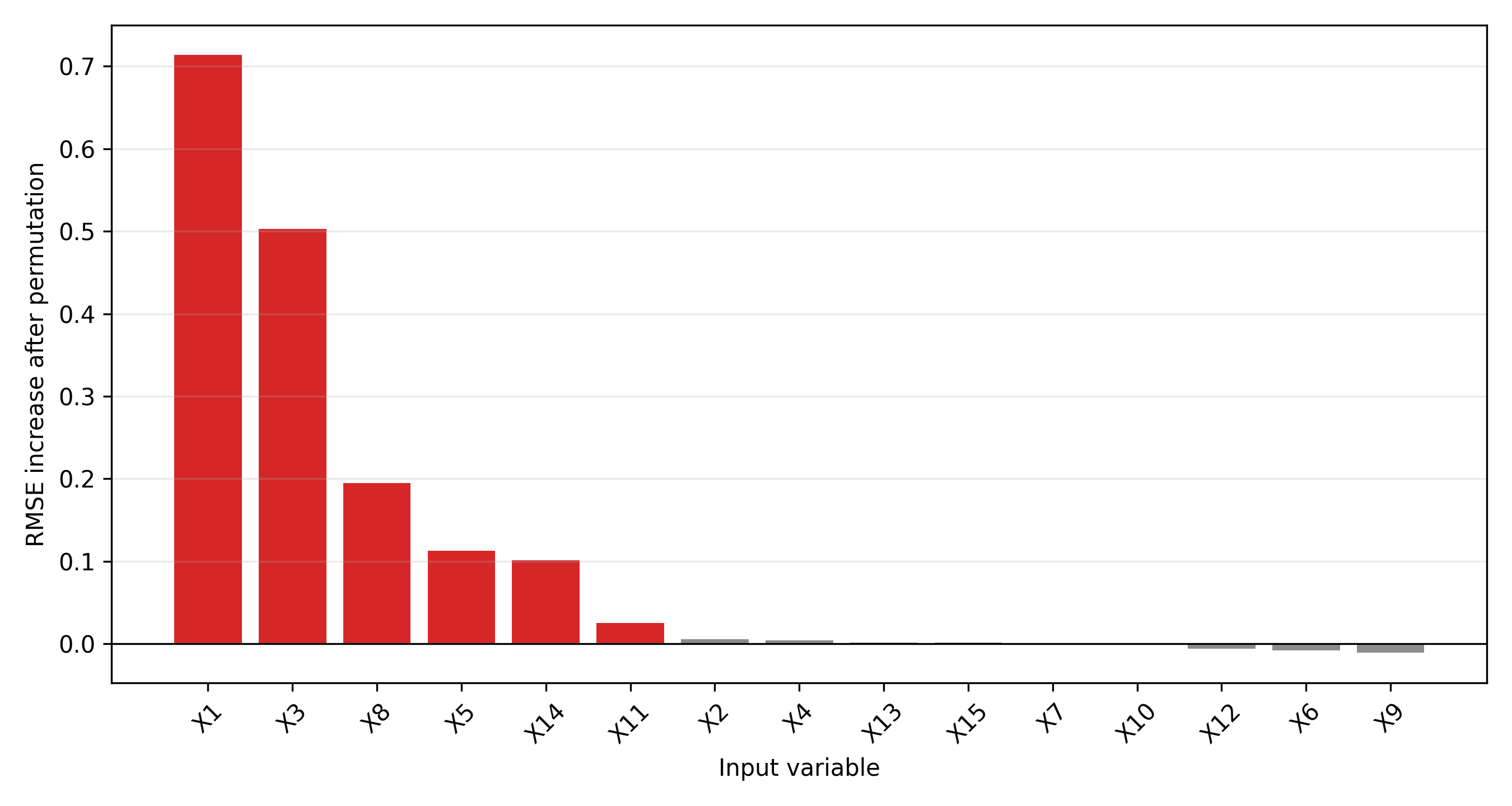}
        \caption{Less interrelated 6-input case.}
    \end{subfigure}

    \caption{Permutation-importance diagnostics for the DNN benchmark across the synthetic experiments. Larger values indicate greater deterioration in test RMSE after permutation of the corresponding input.}
    \label{fig:dnn_permutation_importance_synthetic}
\end{figure}

Figure~\ref{fig:dnn_permutation_importance_synthetic} shows that the DNN does recover substantial parts of the relevance structure. In the baseline setting, the largest importance scores are assigned to \(X_1\), \(X_3\), \(X_8\), \(X_5\), \(X_{14}\), and \(X_{11}\), which coincide with the truly relevant inputs, while the nuisance inputs have negligible or near-zero importance. The same broad pattern is seen in the less interrelated 6-input case. In the 3-relevant-input experiment, the three active inputs \(X_1\), \(X_6\), and \(X_{12}\) are sharply distinguished from the remaining inputs. In the 9- and 12-relevant-input cases, separation is weaker, but the highest-ranked variables still correspond predominantly to truly relevant inputs.

The permutation-importance diagnostics show that the DNN is learning the relevance structure reasonably well. However, this does not translate into the predictive gains achieved by the Wishart prior. Despite being trained on a substantially larger dataset, the DNN RMSE and MAE values remain broadly at the same level as those of the Normal and Wishart + Normal GP specifications. The difference therefore appears to lie not merely in whether informative inputs are identified, but in how strongly nuisance directions are suppressed in the induced predictive geometry. In the GP setting, the Wishart prior acts directly on the covariance structure and produces much sharper relevant--irrelevant separation than priors placed only on the hyperparameters. The DNN benchmark therefore suggests that recovering the relevant inputs is not, by itself, sufficient to attain the same predictive gains: the main advantage of the Wishart prior appears instead to lie in its stronger covariance-level suppression of nuisance inputs.

\FloatBarrier

\subsection{Repeated-seed stability}

To assess whether the baseline pattern is stable, the baseline experiment was repeated over five runs in total. The summary is shown in Table~\ref{tab:baseline_repeated}.

\begin{table}[!htbp]
\centering
\begin{tabular}{lccc}
\toprule
Method & Mean RMSE (sd) & Mean MAE (sd) & Mean relevance ratio (sd) \\
\midrule
Normal Prior & 0.274508 (0.013181) & 0.222939 (0.009739) & 0.498388 (0.028005) \\
Wishart Prior & 0.138322 (0.011924) & 0.103407 (0.005759) & 0.065716 (0.002147) \\
Wishart + Normal Prior & 0.274421 (0.013537) & 0.222957 (0.010049) & 0.468219 (0.018630) \\
\bottomrule
\end{tabular}
\caption{Repeated-seed summary for the baseline 15-dimensional synthetic experiment over five runs in total.}
\label{tab:baseline_repeated}
\end{table}

\FloatBarrier

The repeated-seed results are consistent with the single-run baseline pattern. The Normal prior and the Wishart + Normal prior remain almost indistinguishable from each other, while the Wishart prior again gives much lower prediction error and a dramatically smaller relevance ratio. The main conclusion from the single baseline run therefore persists across repeated random seeds.

\subsection{Illustrative multi-panel figures}

The numerical summaries above establish the main synthetic pattern, but the figures help show how it appears within individual runs. For each experiment, the corresponding multi-panel figure combines the three post-burn-in ARD lengthscale chain plots with predictive summaries. The first row shows the ARD lengthscale chains under the three GP prior specifications. The second row shows the corresponding GP posterior predictive summaries. The third row adds the DNN predictive summaries for the same experiment.

Since the DNN benchmark is not an ARD GP model, there are no lengthscale chains to display. The DNN is therefore shown only through its predictive behaviour, using the same two visual summaries: predicted values against test-point index and predicted values against the true test responses. This gives a direct visual comparison of predictive performance, without suggesting that the DNN has an analogous hyperparameter-relevance trace.

\begin{figure}[!htbp]
    \centering

    \begin{subfigure}[t]{0.32\textwidth}
        \centering
        \includegraphics[width=\linewidth]{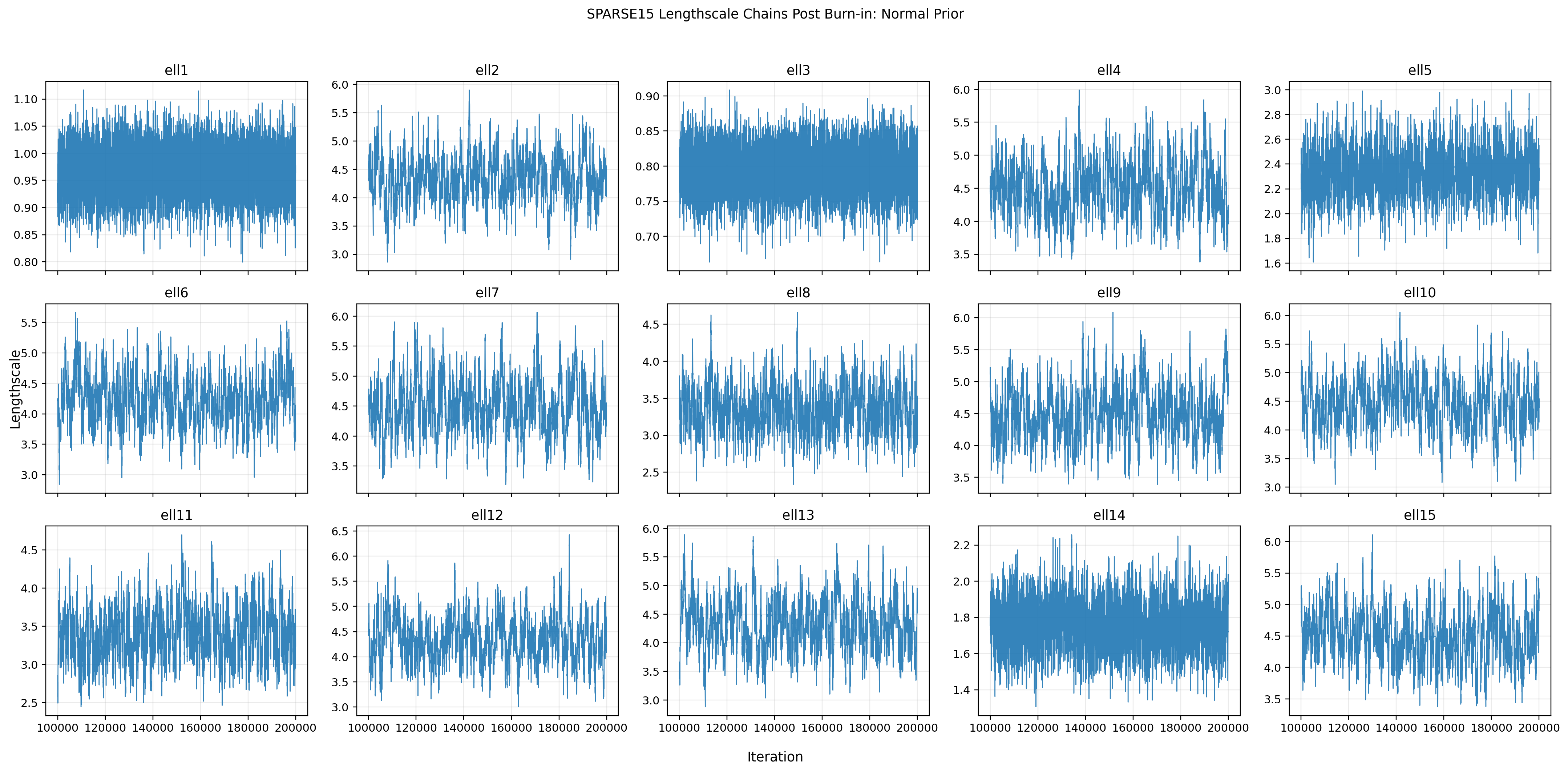}
        \caption{Normal prior.}
    \end{subfigure}\hfill
    \begin{subfigure}[t]{0.32\textwidth}
        \centering
        \includegraphics[width=\linewidth]{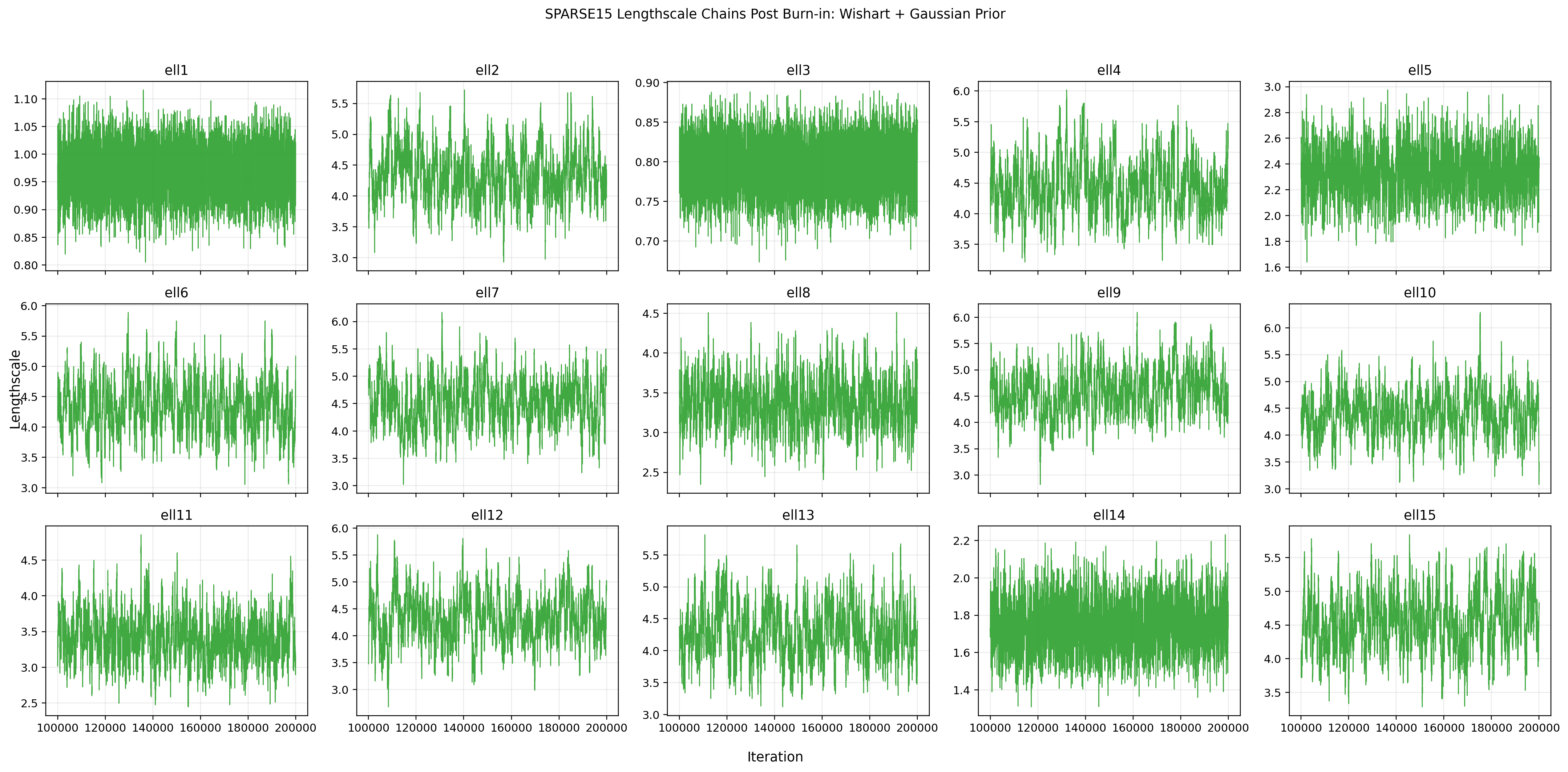}
        \caption{Wishart + Normal prior.}
    \end{subfigure}\hfill
    \begin{subfigure}[t]{0.32\textwidth}
        \centering
        \includegraphics[width=\linewidth]{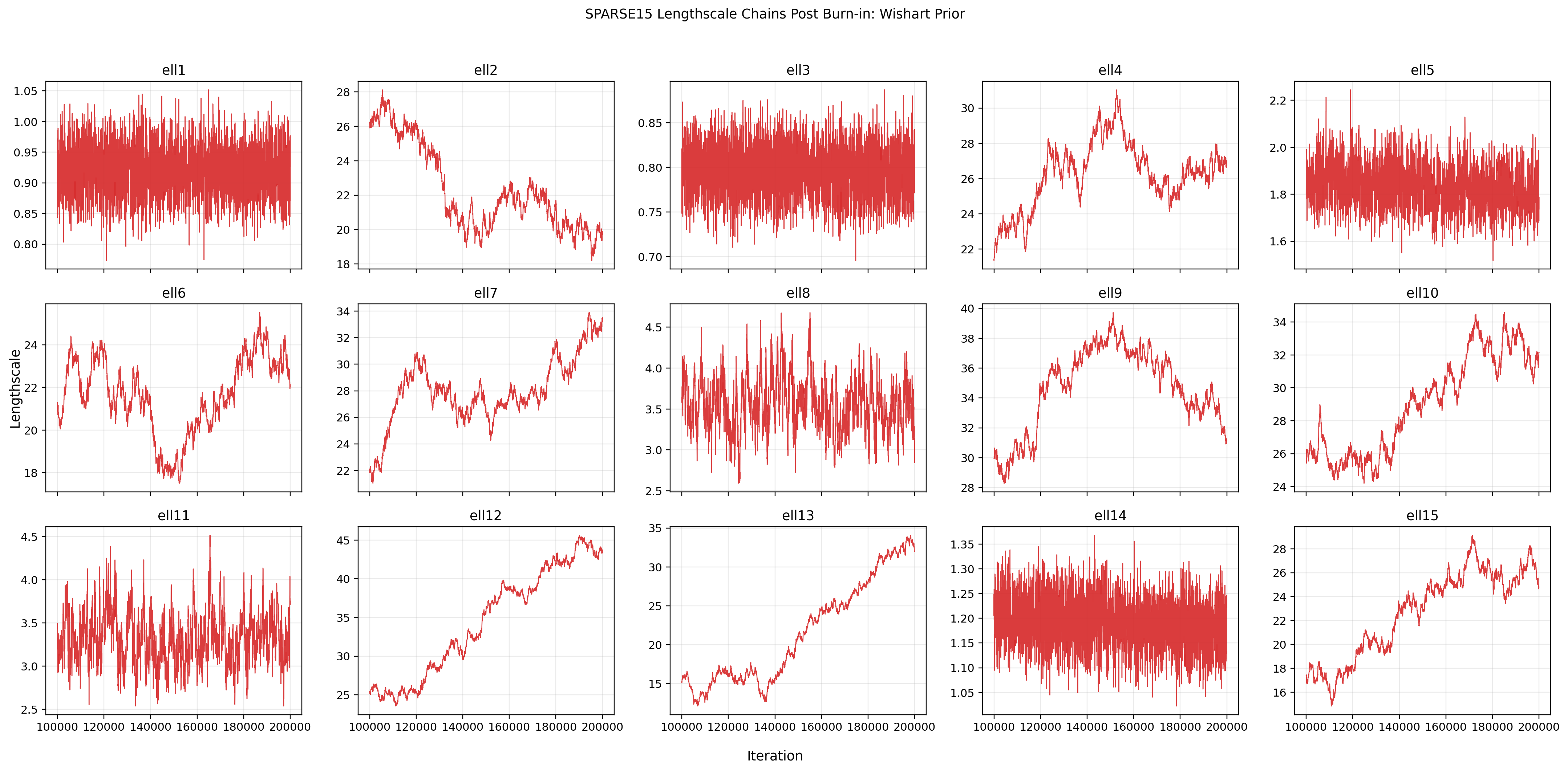}
        \caption{Wishart prior.}
    \end{subfigure}

    \vspace{0.8em}

    \begin{subfigure}[t]{0.49\textwidth}
        \centering
        \includegraphics[width=\linewidth]{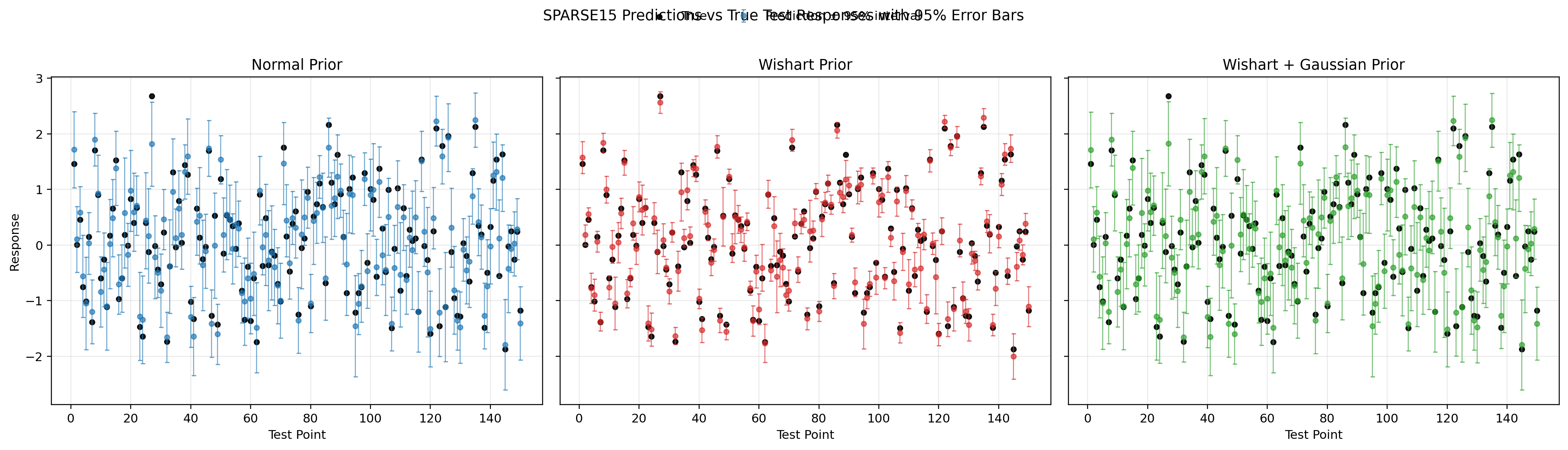}
        \caption{GP posterior predictive means against test-point index.}
    \end{subfigure}\hfill
    \begin{subfigure}[t]{0.49\textwidth}
        \centering
        \includegraphics[width=\linewidth]{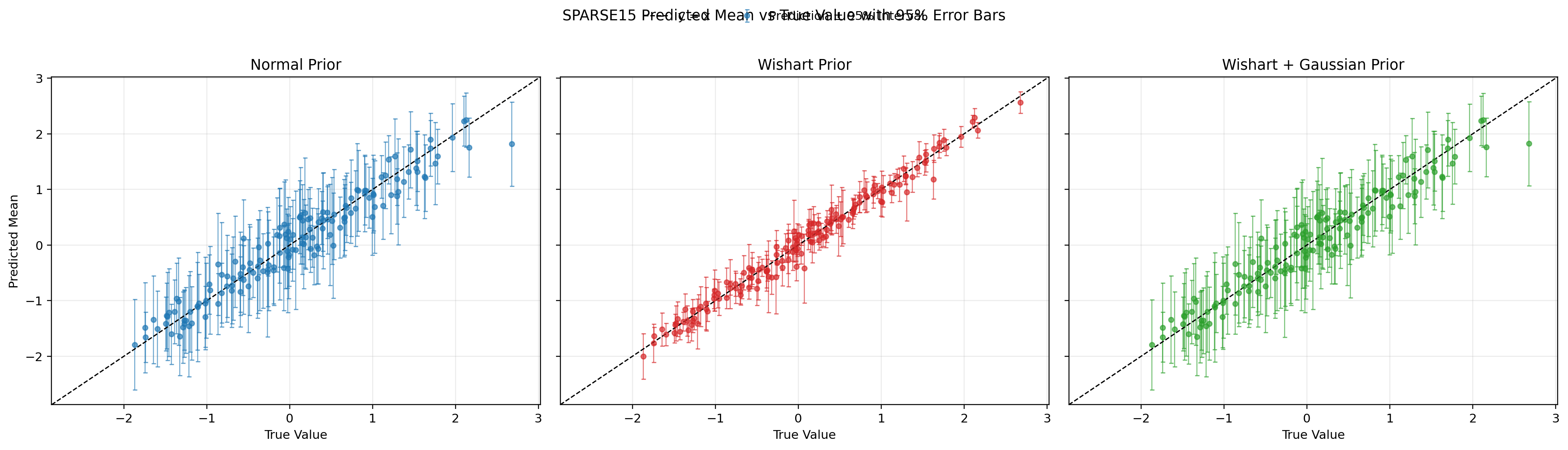}
        \caption{GP predicted mean versus true test response.}
    \end{subfigure}

    \vspace{0.8em}

    \begin{subfigure}[t]{0.35\textwidth}
        \centering
        \includegraphics[width=\linewidth]{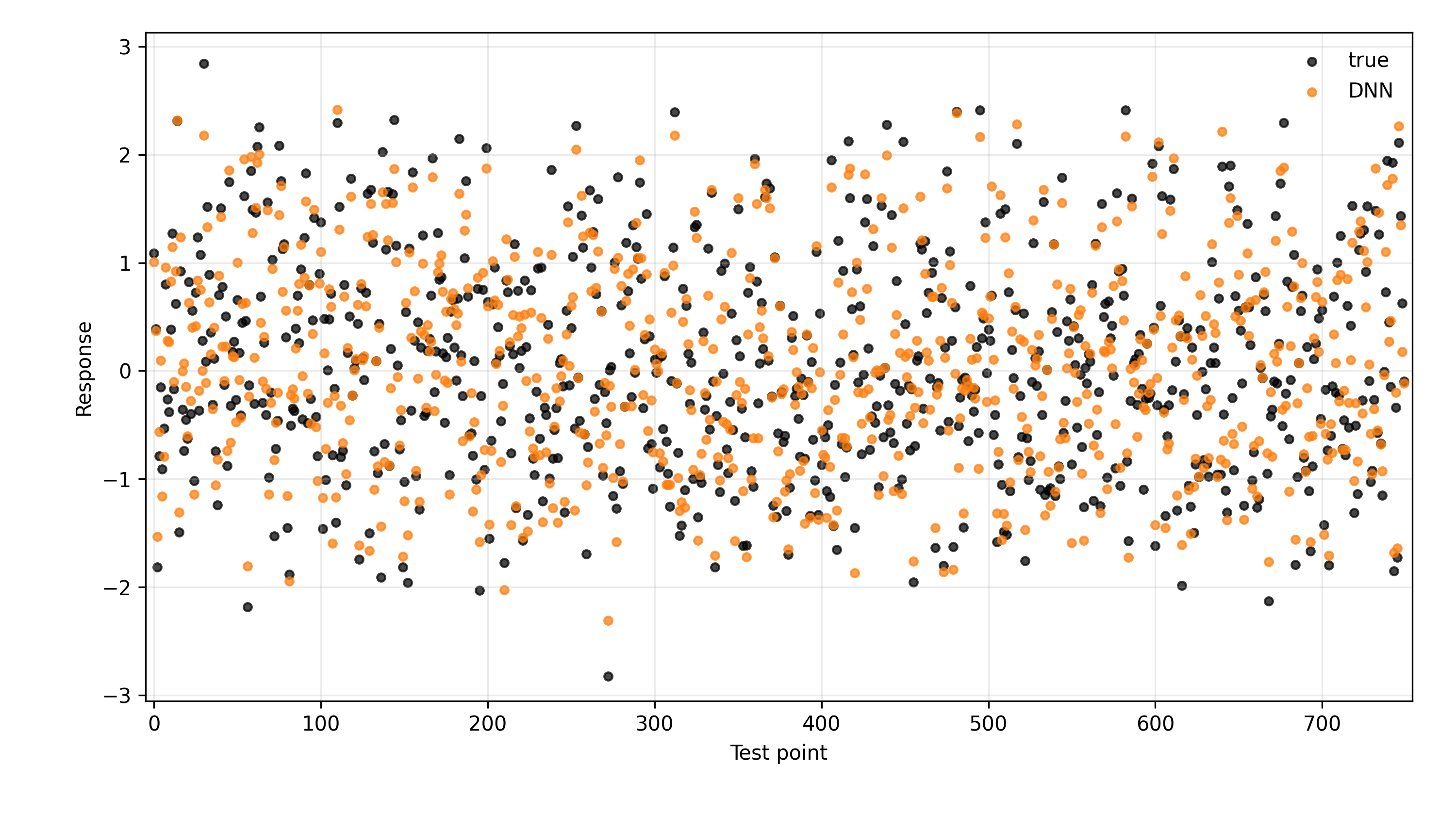}
        \caption{DNN predictions against test-point index.}
    \end{subfigure}\hfill
    \begin{subfigure}[t]{0.35\textwidth}
        \centering
        \includegraphics[width=\linewidth]{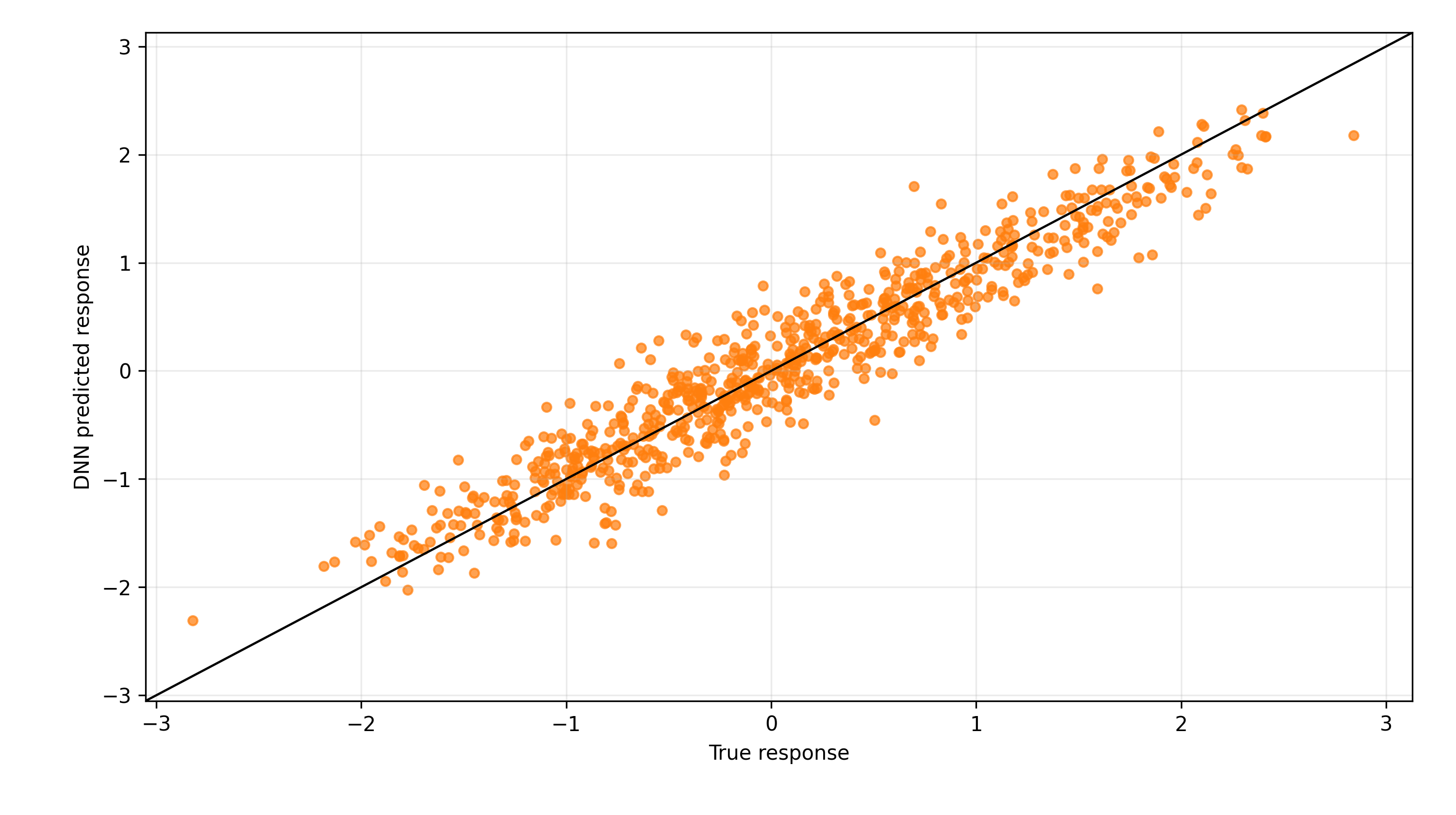}
        \caption{DNN predicted response versus true test response.}
    \end{subfigure}

    \caption{Baseline 15-dimensional synthetic experiment. Top row: post-burn-in ARD lengthscale chains under the three GP prior specifications. Middle row: GP predictive summaries. Bottom row: DNN predictive summaries. The DNN is included only as a predictive benchmark, since it does not have analogous ARD lengthscale chains.}
    \label{fig:baseline_synthetic_combined}
\end{figure}

Figure~\ref{fig:baseline_synthetic_combined} summarises the baseline experiment. The Wishart prior leaves the relevant inputs on a broadly comparable scale to the competing priors, but drives many nuisance-input lengthscales much higher. The predictive panels show the corresponding effect on performance: the Wishart predictive means lie much closer to the true responses, particularly away from the centre of the response range.

\FloatBarrier

\begin{figure}[!htbp]
    \centering

    \begin{subfigure}[t]{0.32\textwidth}
        \centering
        \includegraphics[width=\linewidth]{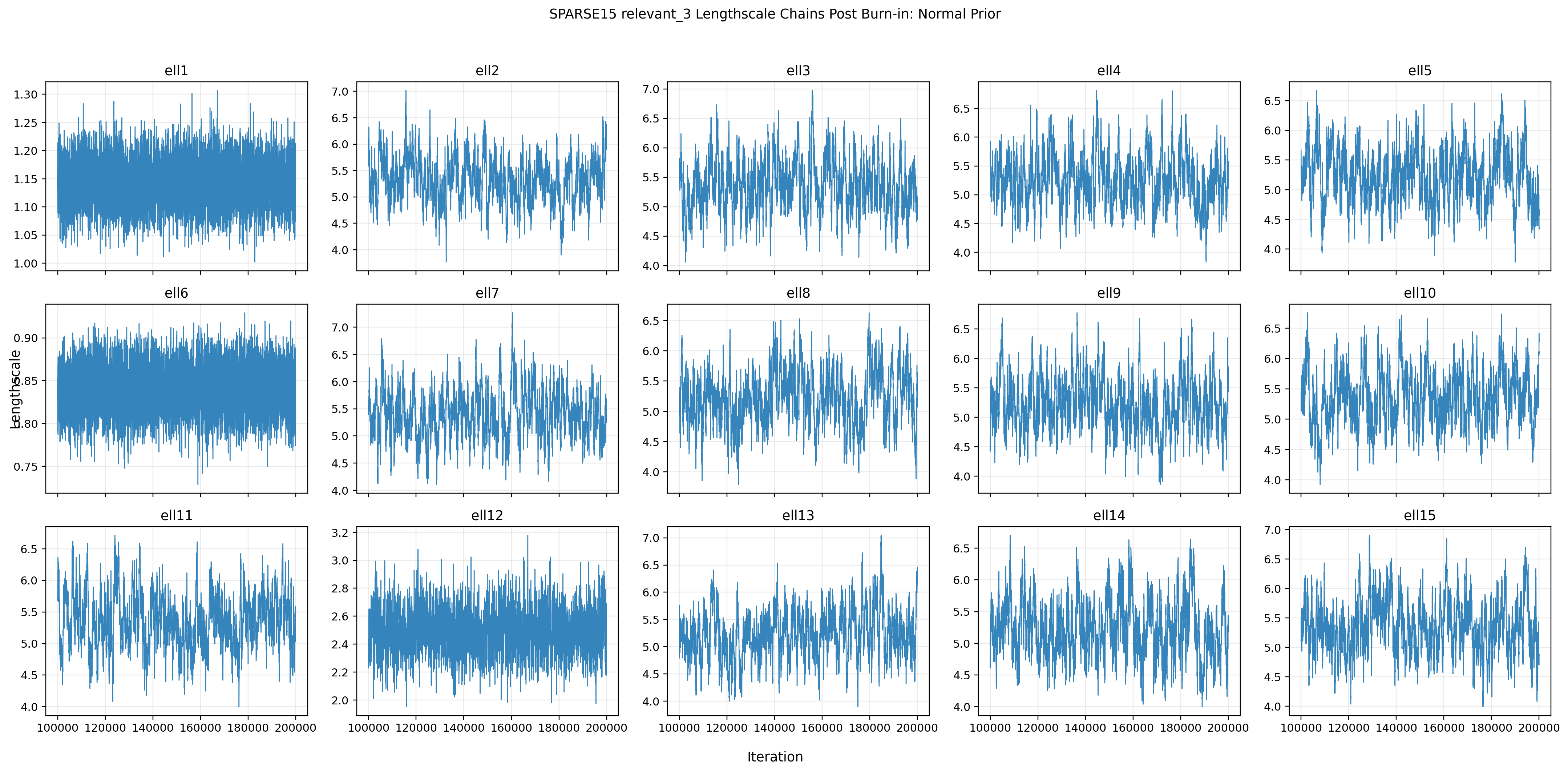}
        \caption{Normal prior.}
    \end{subfigure}\hfill
    \begin{subfigure}[t]{0.32\textwidth}
        \centering
        \includegraphics[width=\linewidth]{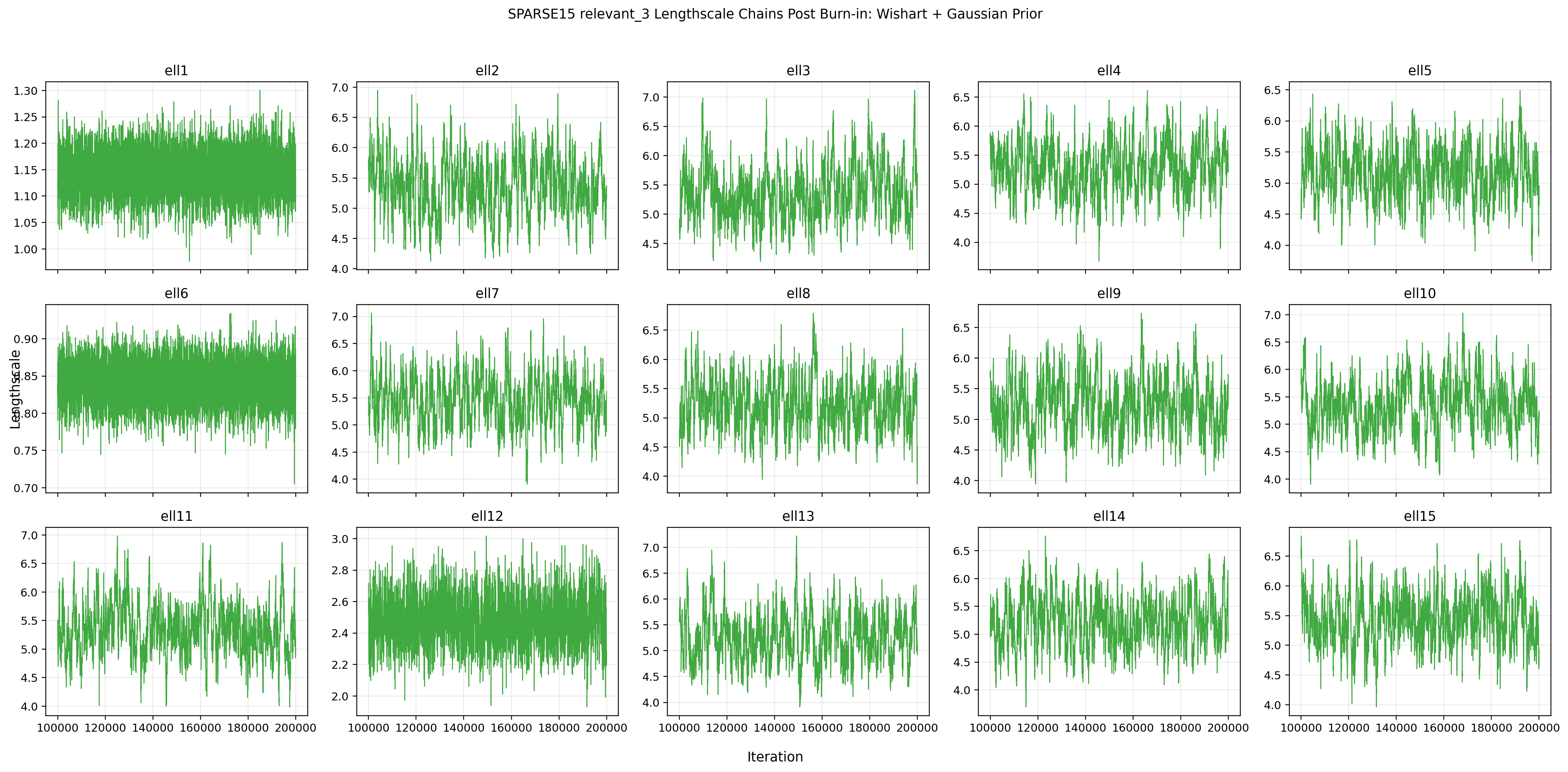}
        \caption{Wishart + Normal prior.}
    \end{subfigure}\hfill
    \begin{subfigure}[t]{0.32\textwidth}
        \centering
        \includegraphics[width=\linewidth]{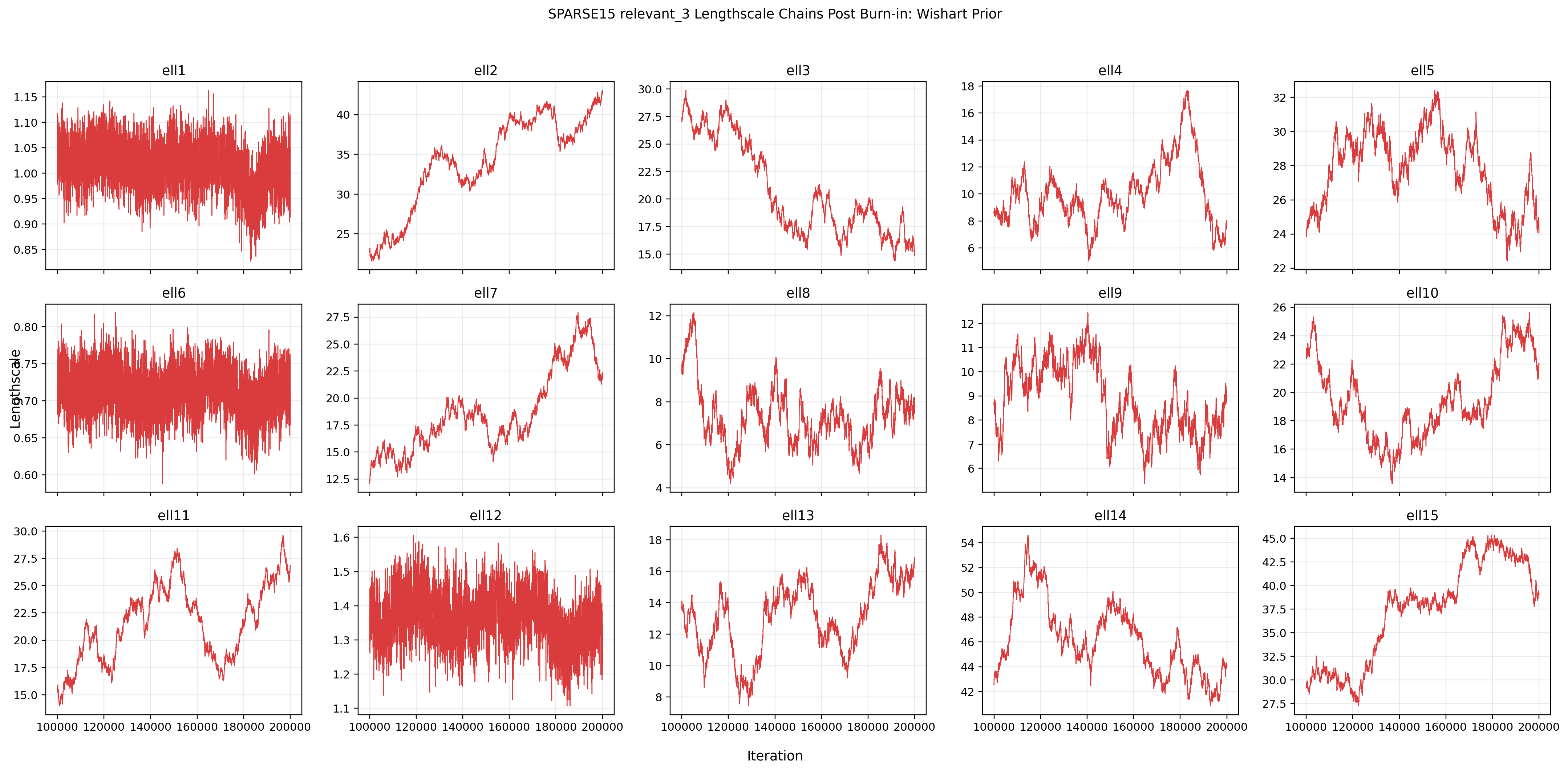}
        \caption{Wishart prior.}
    \end{subfigure}

    \vspace{0.8em}

    \begin{subfigure}[t]{0.49\textwidth}
        \centering
        \includegraphics[width=\linewidth]{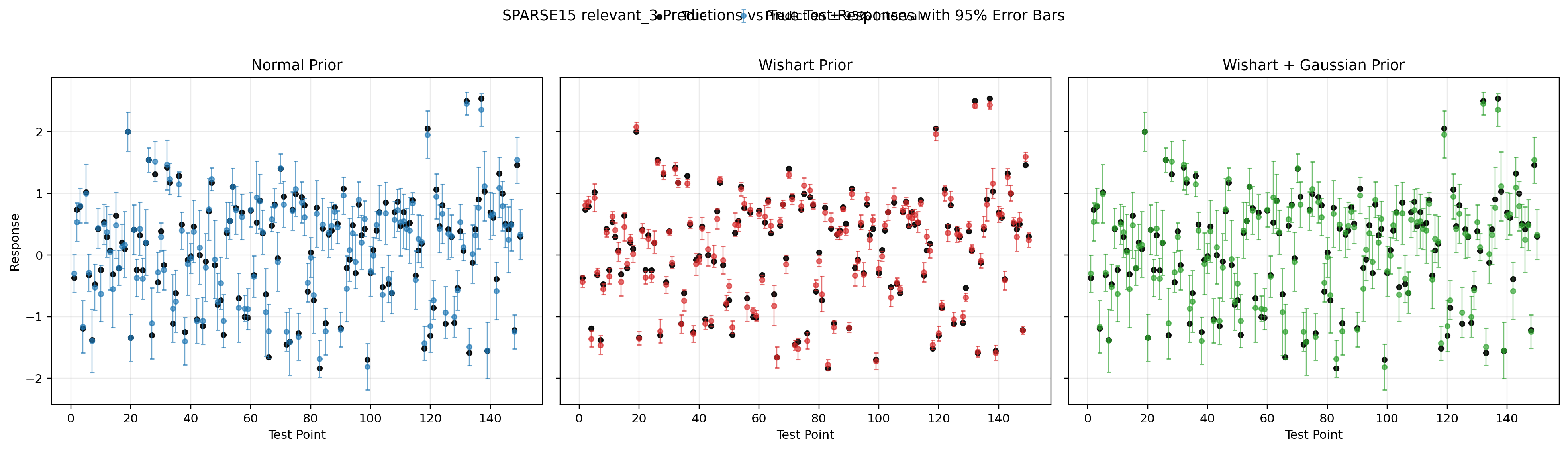}
        \caption{GP posterior predictive means against test-point index.}
    \end{subfigure}\hfill
    \begin{subfigure}[t]{0.49\textwidth}
        \centering
        \includegraphics[width=\linewidth]{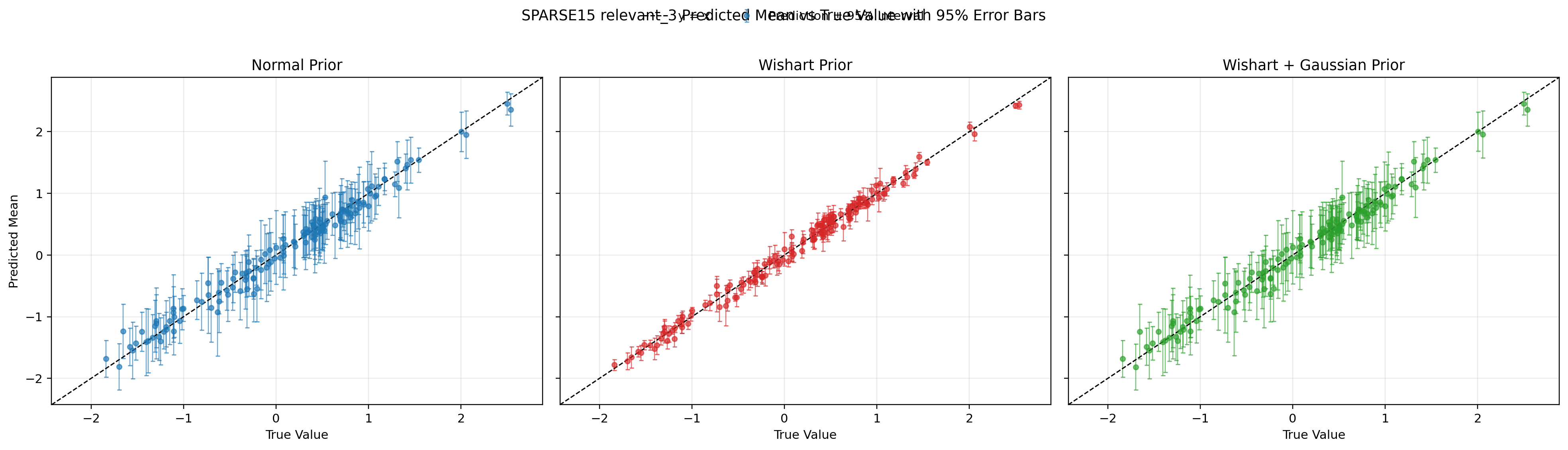}
        \caption{GP predicted mean versus true test response.}
    \end{subfigure}

    \vspace{0.8em}

    \begin{subfigure}[t]{0.35\textwidth}
        \centering
        \includegraphics[width=\linewidth]{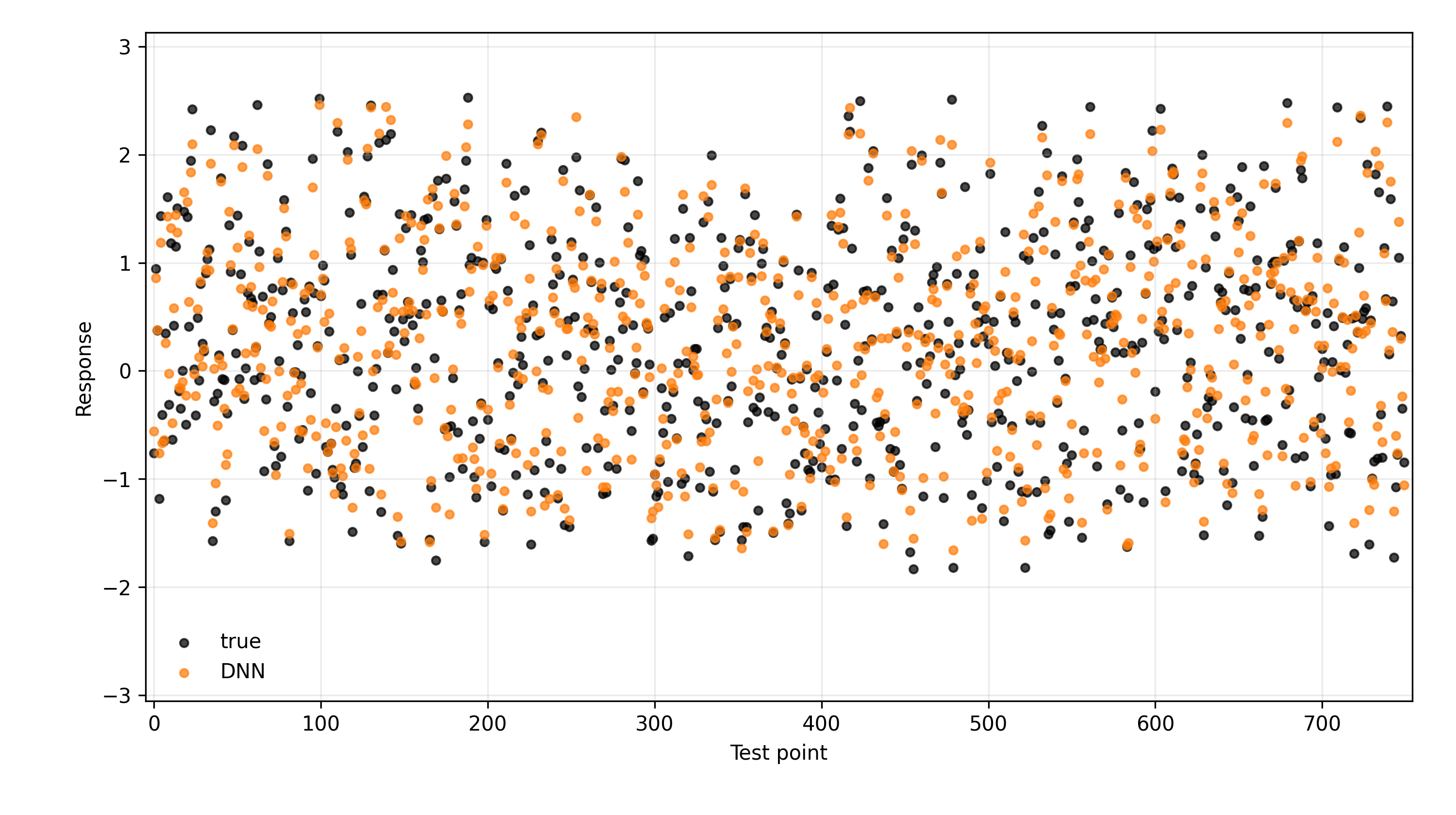}
        \caption{DNN predictions against test-point index.}
    \end{subfigure}\hfill
    \begin{subfigure}[t]{0.35\textwidth}
        \centering
        \includegraphics[width=\linewidth]{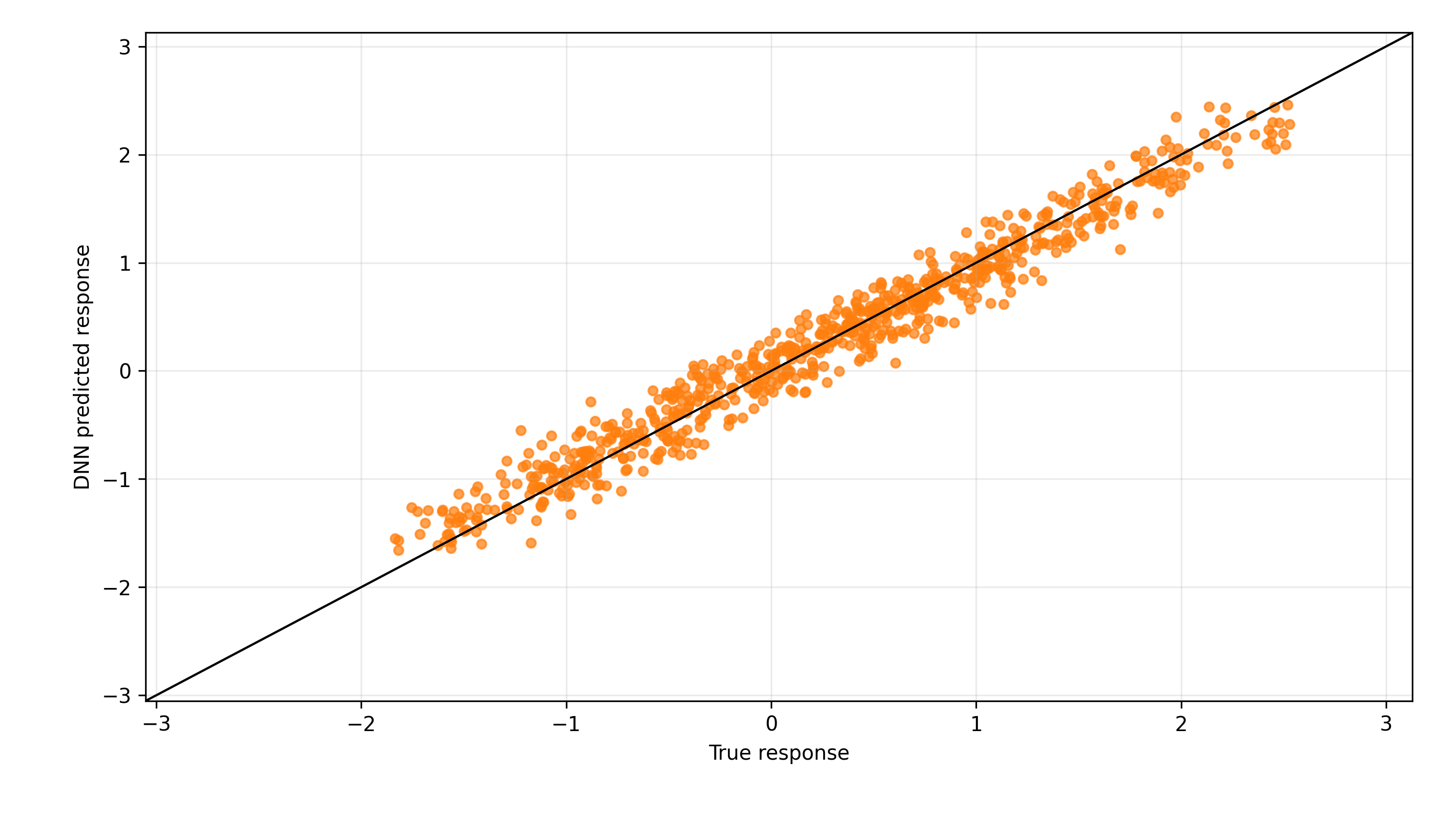}
        \caption{DNN predicted response versus true test response.}
    \end{subfigure}

    \caption{Synthetic experiment with 3 relevant inputs. Top row: post-burn-in ARD lengthscale chains under the three GP prior specifications. Middle row: GP predictive summaries. Bottom row: DNN predictive summaries. The DNN is included only as a predictive benchmark, since it does not have analogous ARD lengthscale chains.}
    \label{fig:rel3_synthetic_combined}
\end{figure}

Figure~\ref{fig:rel3_synthetic_combined} shows the 3-input additive case. The same separation mechanism is visible in a simpler setting: the Wishart prior again pushes nuisance inputs much farther out than the competing priors, and the predictive panels show a clear gain in point prediction, albeit with less conservative uncertainty intervals.

\FloatBarrier

\begin{figure}[!htbp]
    \centering

    \begin{subfigure}[t]{0.32\textwidth}
        \centering
        \includegraphics[width=\linewidth]{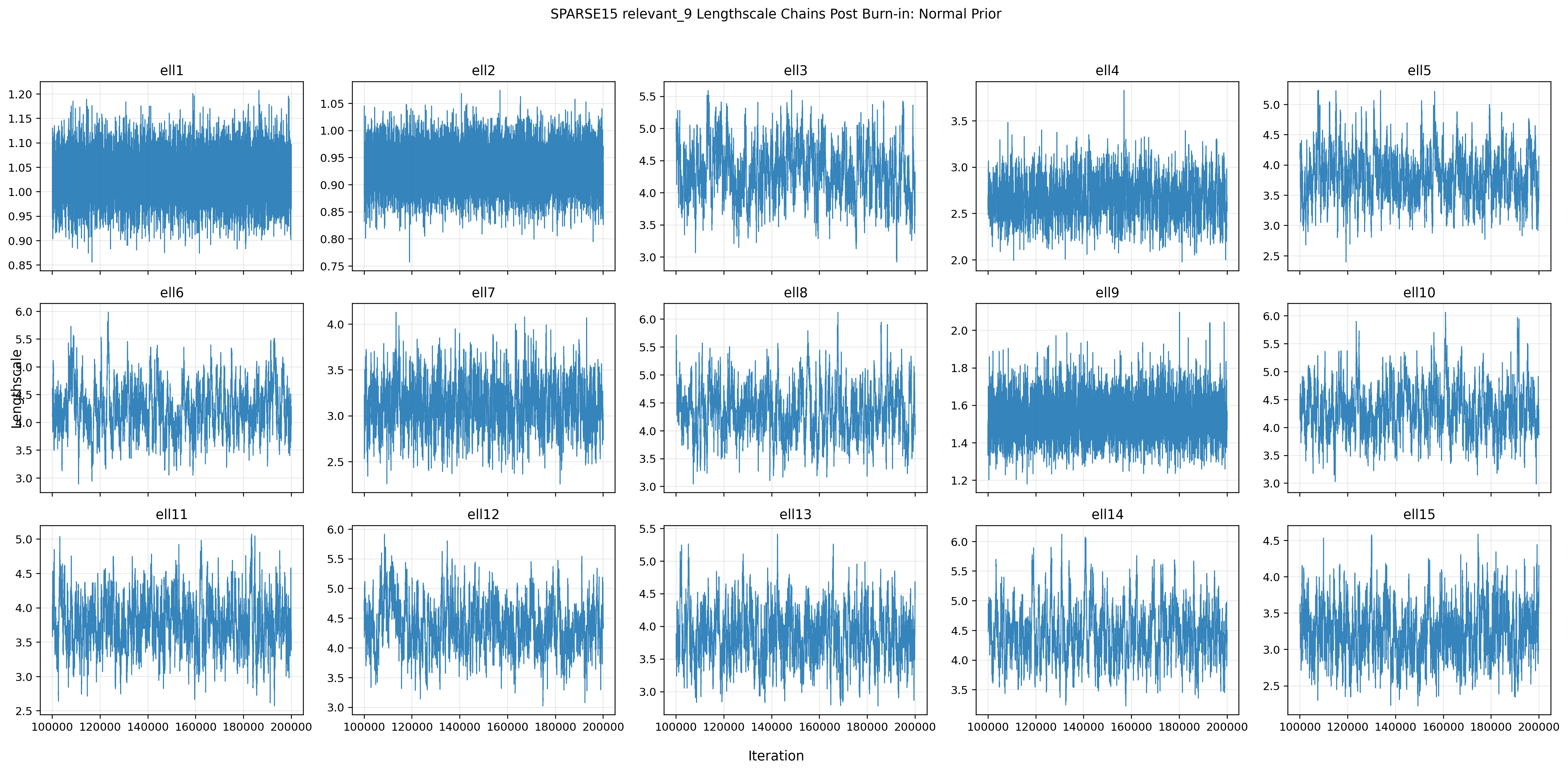}
        \caption{Normal prior.}
    \end{subfigure}\hfill
    \begin{subfigure}[t]{0.32\textwidth}
        \centering
        \includegraphics[width=\linewidth]{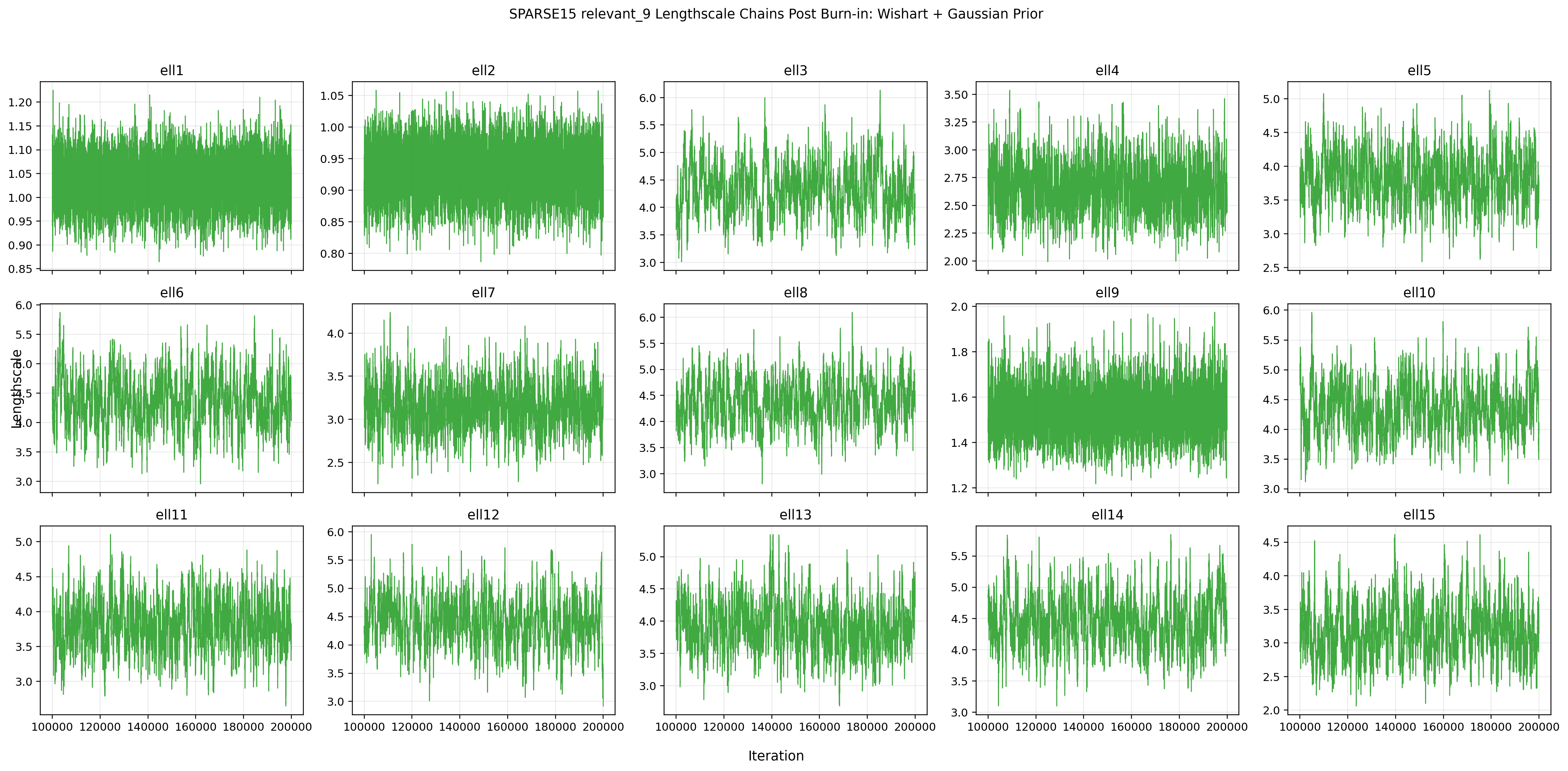}
        \caption{Wishart + Normal prior.}
    \end{subfigure}\hfill
    \begin{subfigure}[t]{0.32\textwidth}
        \centering
        \includegraphics[width=\linewidth]{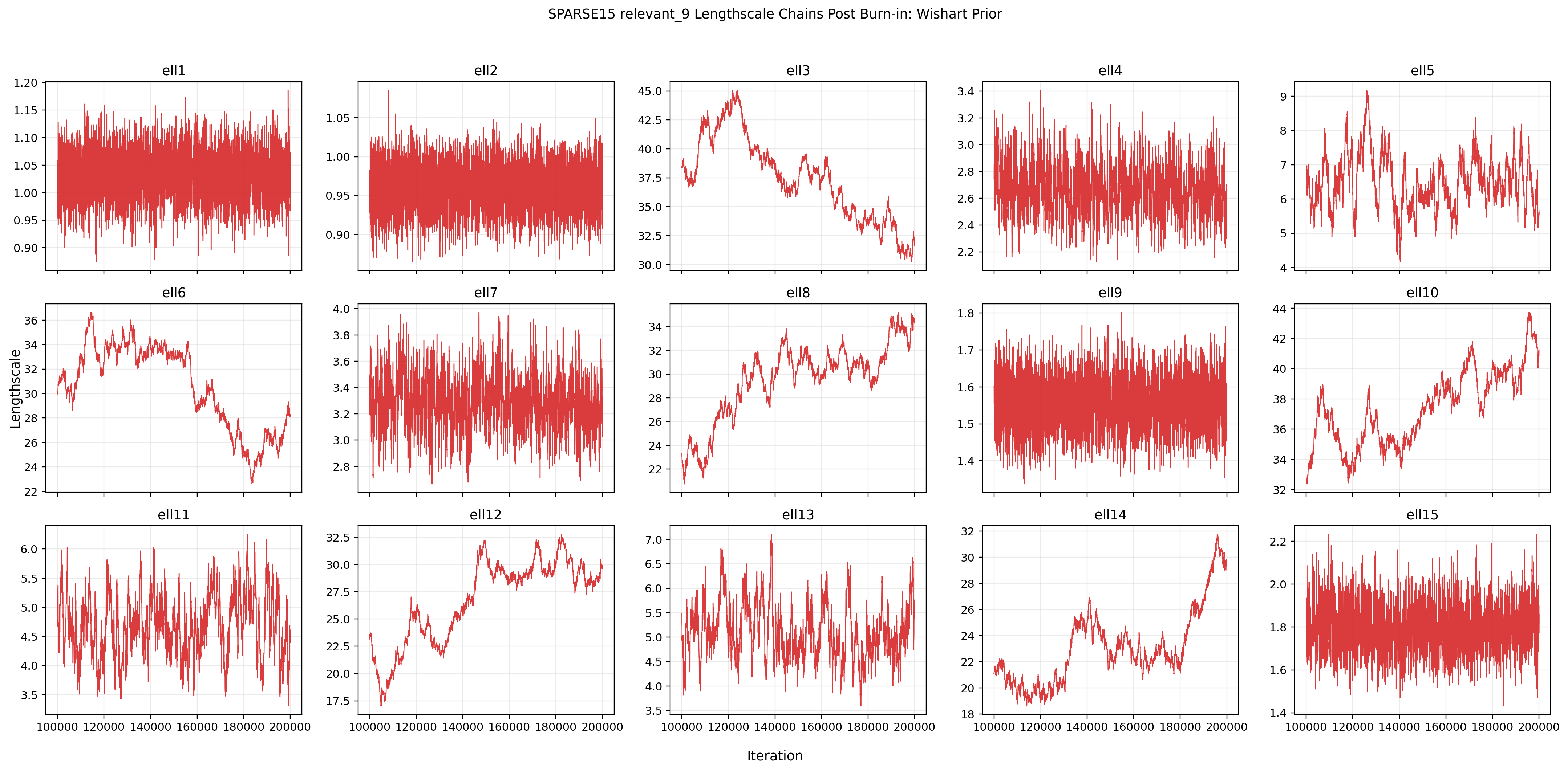}
        \caption{Wishart prior.}
    \end{subfigure}

    \vspace{0.8em}

    \begin{subfigure}[t]{0.49\textwidth}
        \centering
        \includegraphics[width=\linewidth]{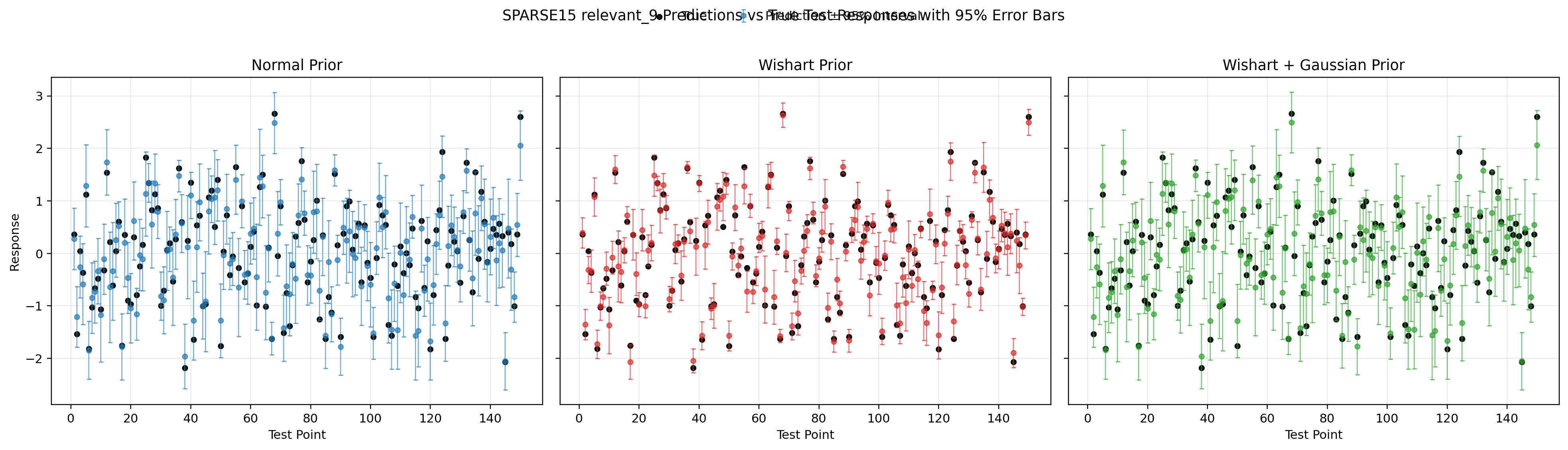}
        \caption{GP posterior predictive means against test-point index.}
    \end{subfigure}\hfill
    \begin{subfigure}[t]{0.49\textwidth}
        \centering
        \includegraphics[width=\linewidth]{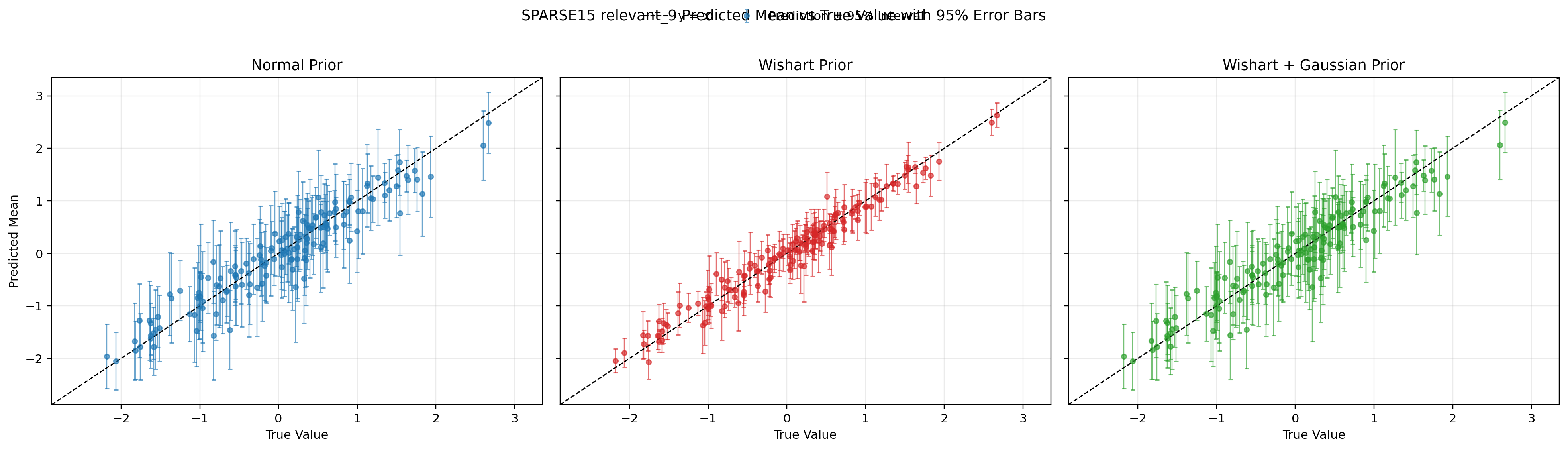}
        \caption{GP predicted mean versus true test response.}
    \end{subfigure}

    \vspace{0.8em}

    \begin{subfigure}[t]{0.35\textwidth}
        \centering
        \includegraphics[width=\linewidth]{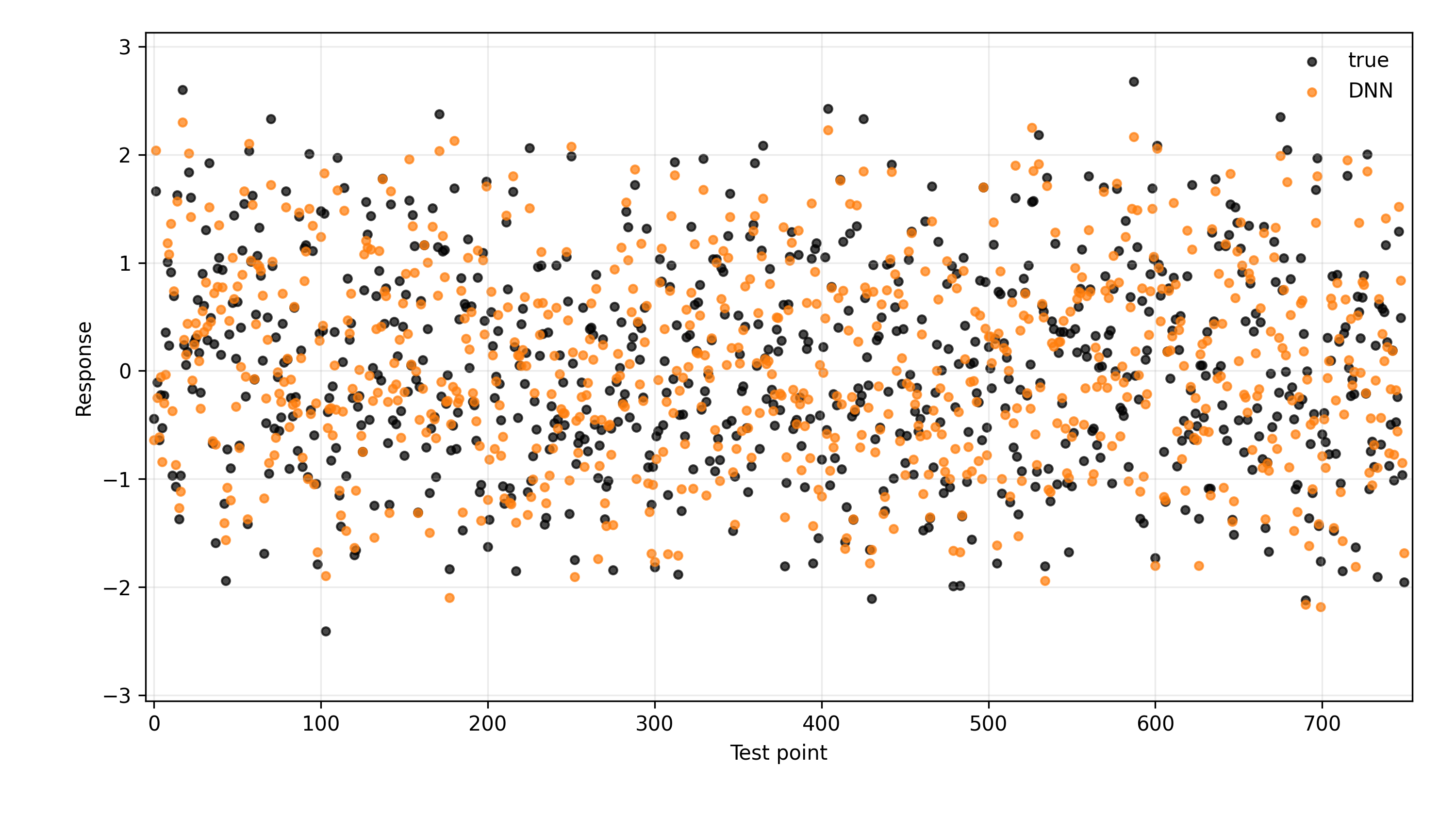}
        \caption{DNN predictions against test-point index.}
    \end{subfigure}\hfill
    \begin{subfigure}[t]{0.35\textwidth}
        \centering
        \includegraphics[width=\linewidth]{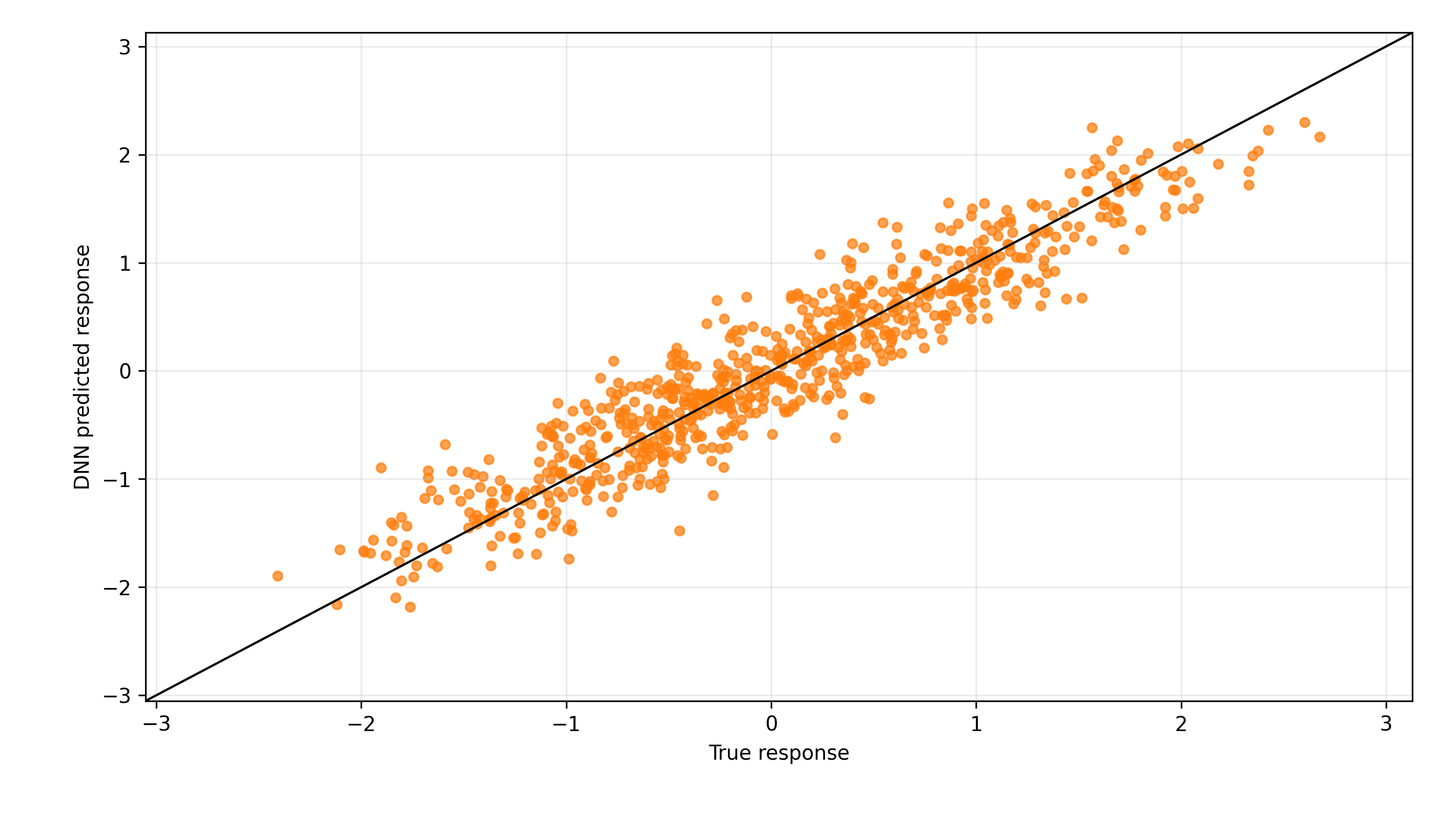}
        \caption{DNN predicted response versus true test response.}
    \end{subfigure}

    \caption{Synthetic experiment with 9 relevant inputs. Top row: post-burn-in ARD lengthscale chains under the three GP prior specifications. Middle row: GP predictive summaries. Bottom row: DNN predictive summaries. The DNN is included only as a predictive benchmark, since it does not have analogous ARD lengthscale chains.}
    \label{fig:rel9_synthetic_combined}
\end{figure}

Figure~\ref{fig:rel9_synthetic_combined} shows that the same pattern persists when the problem is less sparse. The Wishart prior still suppresses nuisance inputs more strongly, and its predictive means remain closer to the true responses than those of the competing priors.

\FloatBarrier

\begin{figure}[!htbp]
    \centering

    \begin{subfigure}[t]{0.32\textwidth}
        \centering
        \includegraphics[width=\linewidth]{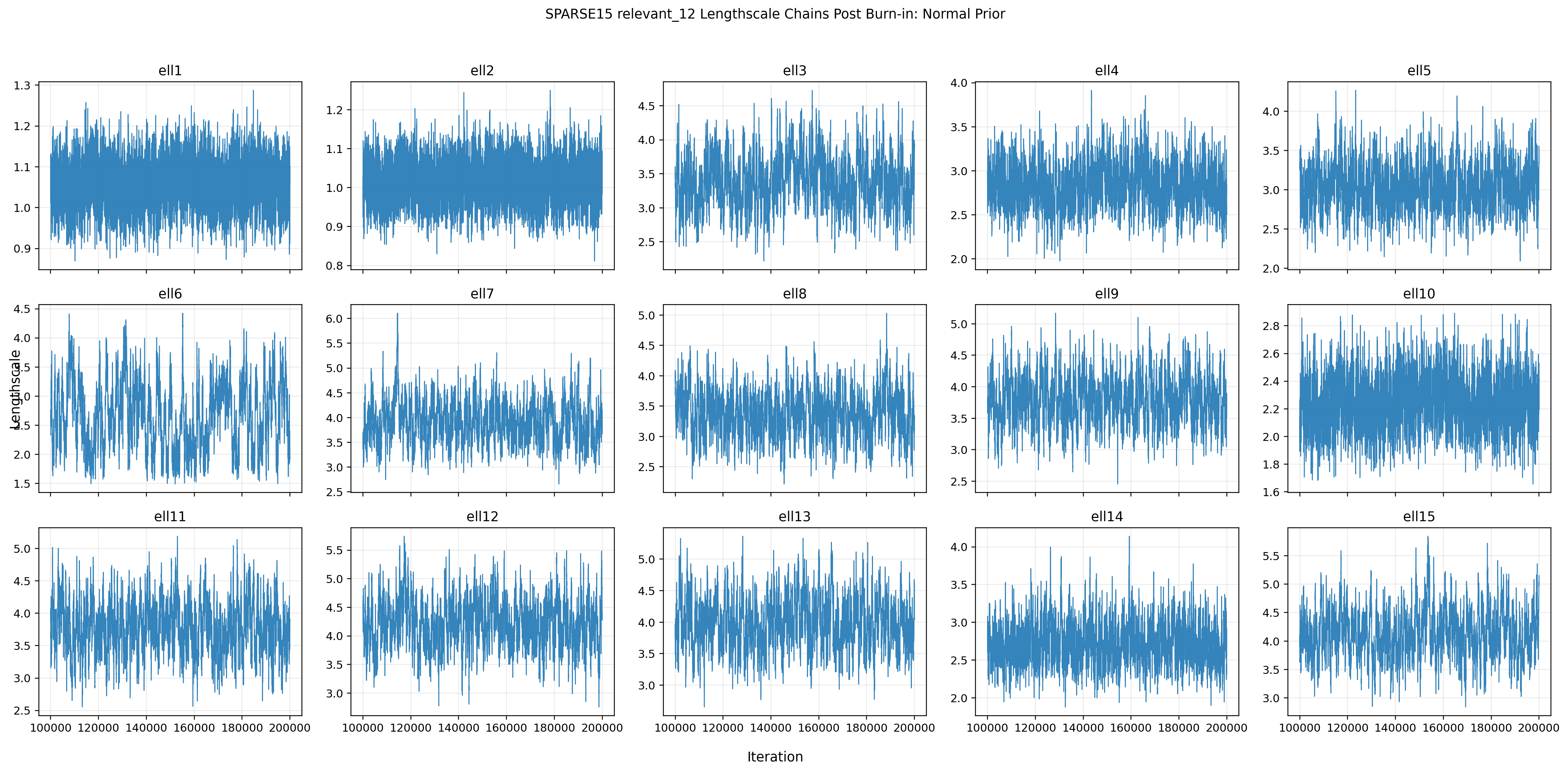}
        \caption{Normal prior.}
    \end{subfigure}\hfill
    \begin{subfigure}[t]{0.32\textwidth}
        \centering
        \includegraphics[width=\linewidth]{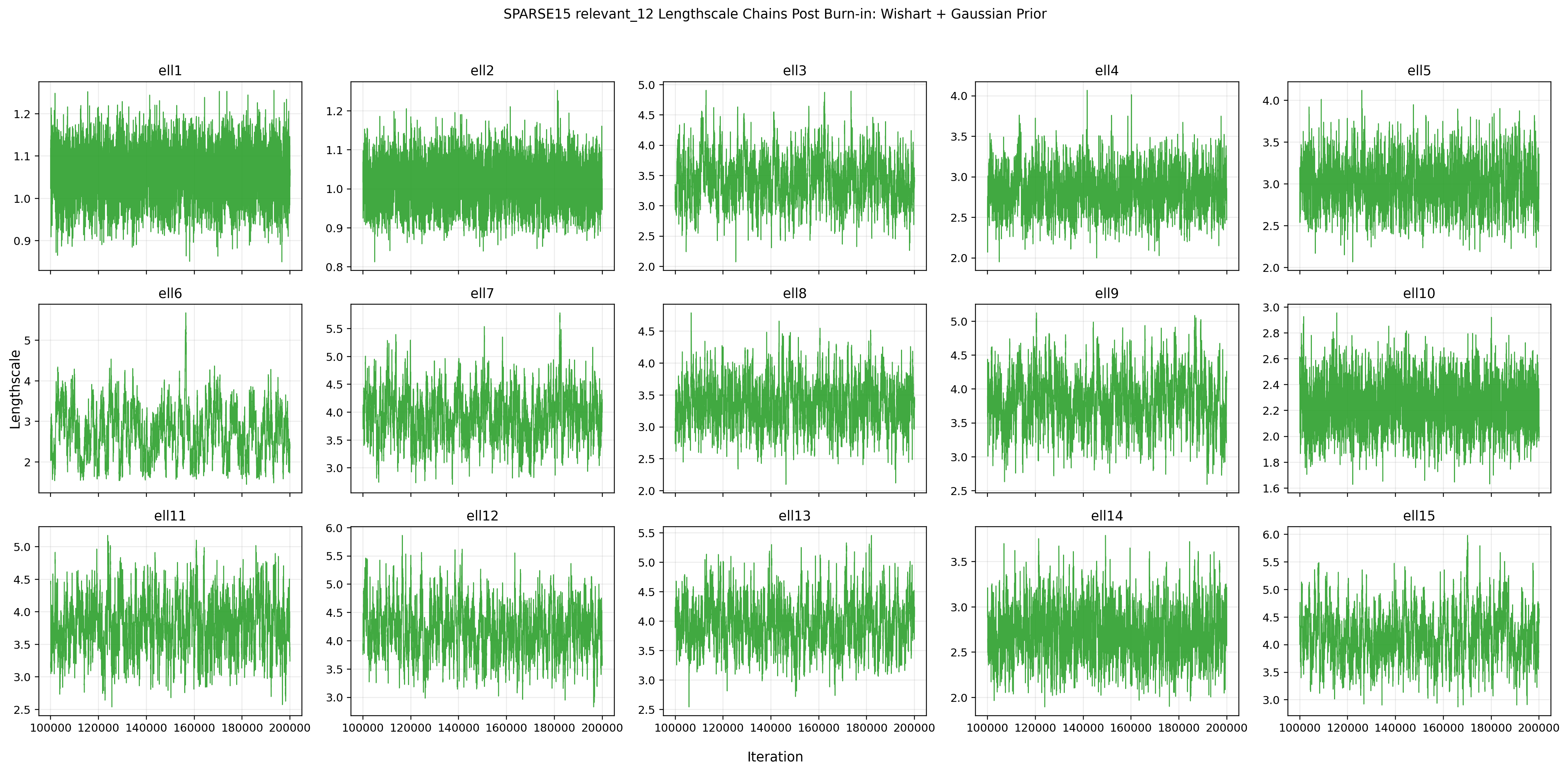}
        \caption{Wishart + Normal prior.}
    \end{subfigure}\hfill
    \begin{subfigure}[t]{0.32\textwidth}
        \centering
        \includegraphics[width=\linewidth]{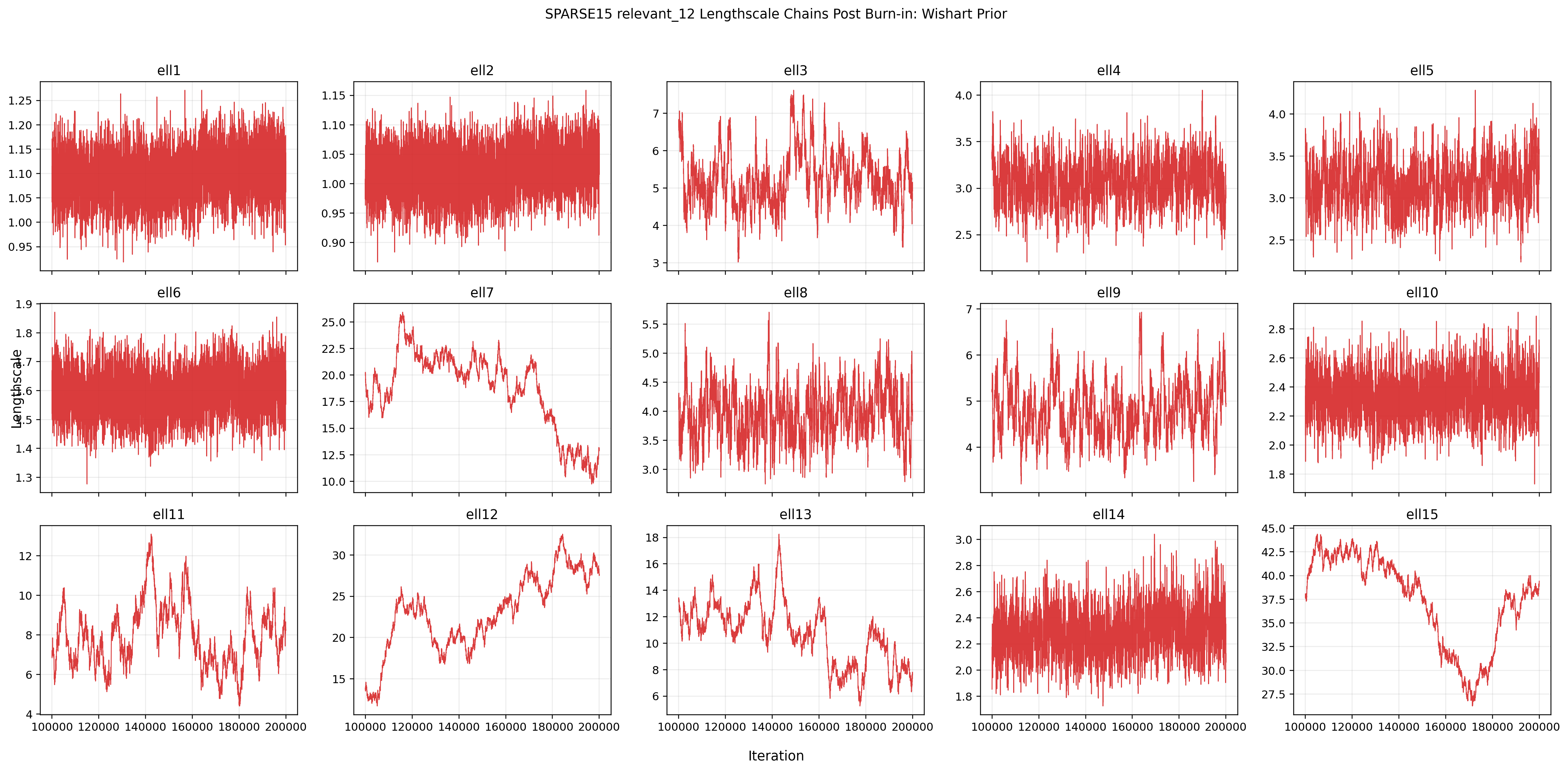}
        \caption{Wishart prior.}
    \end{subfigure}

    \vspace{0.8em}

    \begin{subfigure}[t]{0.49\textwidth}
        \centering
        \includegraphics[width=\linewidth]{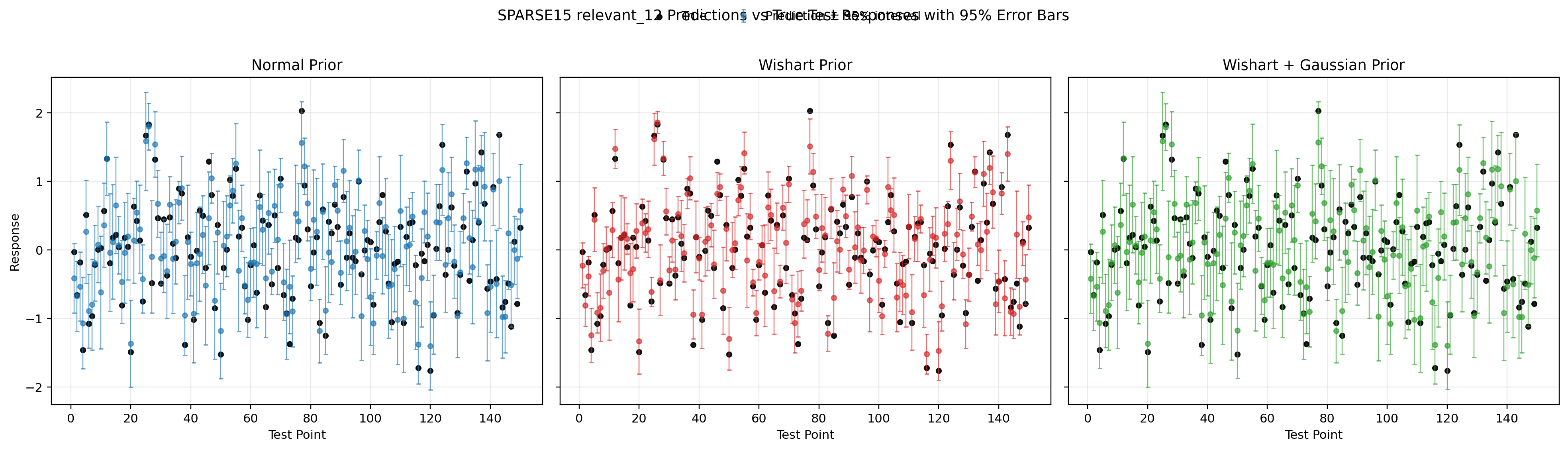}
        \caption{GP posterior predictive means against test-point index.}
    \end{subfigure}\hfill
    \begin{subfigure}[t]{0.49\textwidth}
        \centering
        \includegraphics[width=\linewidth]{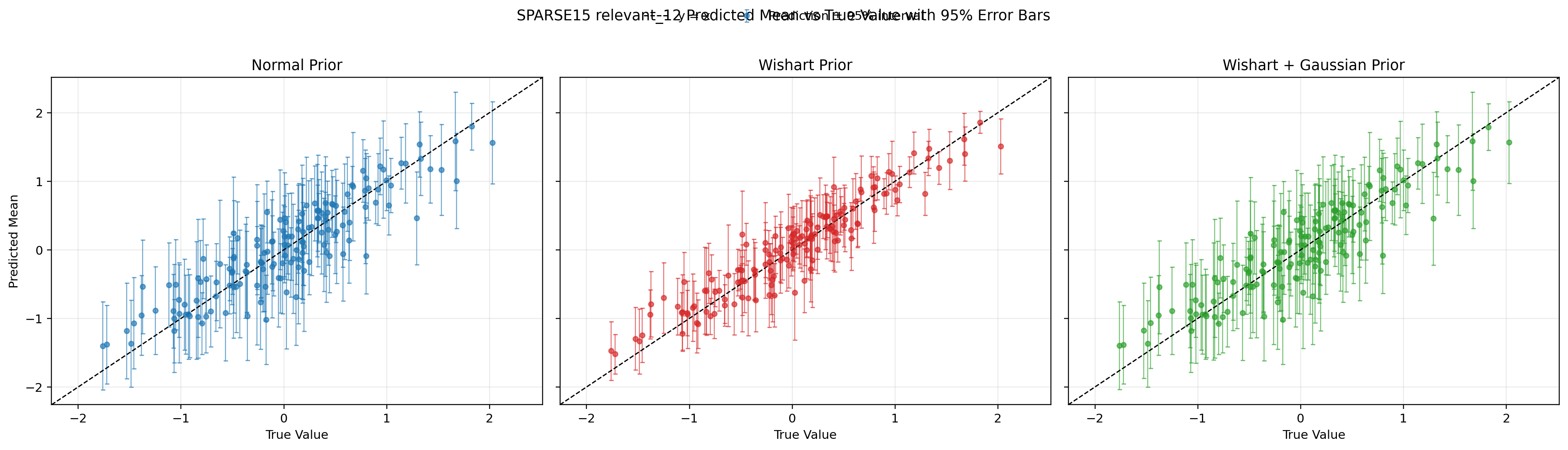}
        \caption{GP predicted mean versus true test response.}
    \end{subfigure}

    \vspace{0.8em}

    \begin{subfigure}[t]{0.35\textwidth}
        \centering
        \includegraphics[width=\linewidth]{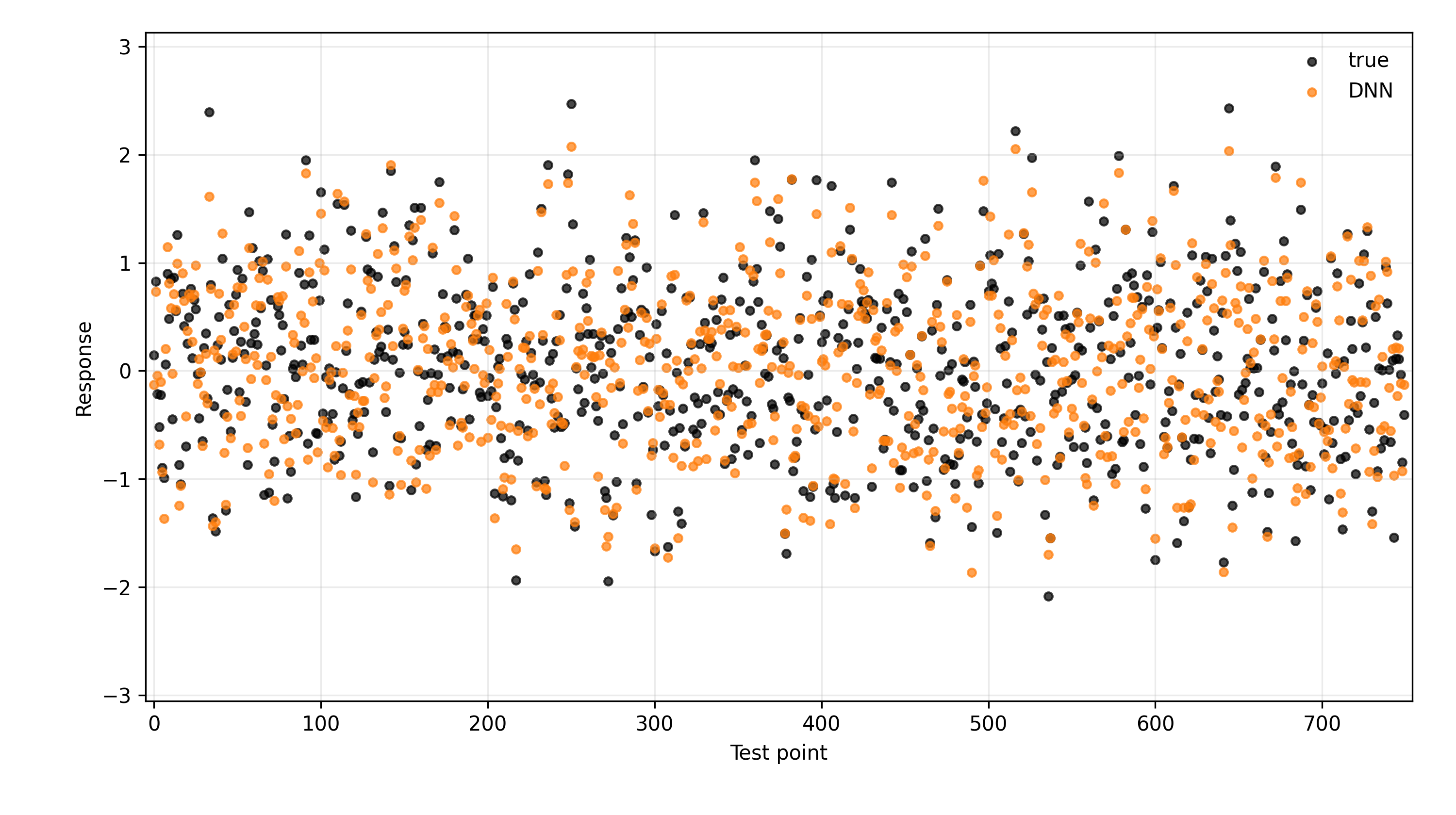}
        \caption{DNN predictions against test-point index.}
    \end{subfigure}\hfill
    \begin{subfigure}[t]{0.35\textwidth}
        \centering
        \includegraphics[width=\linewidth]{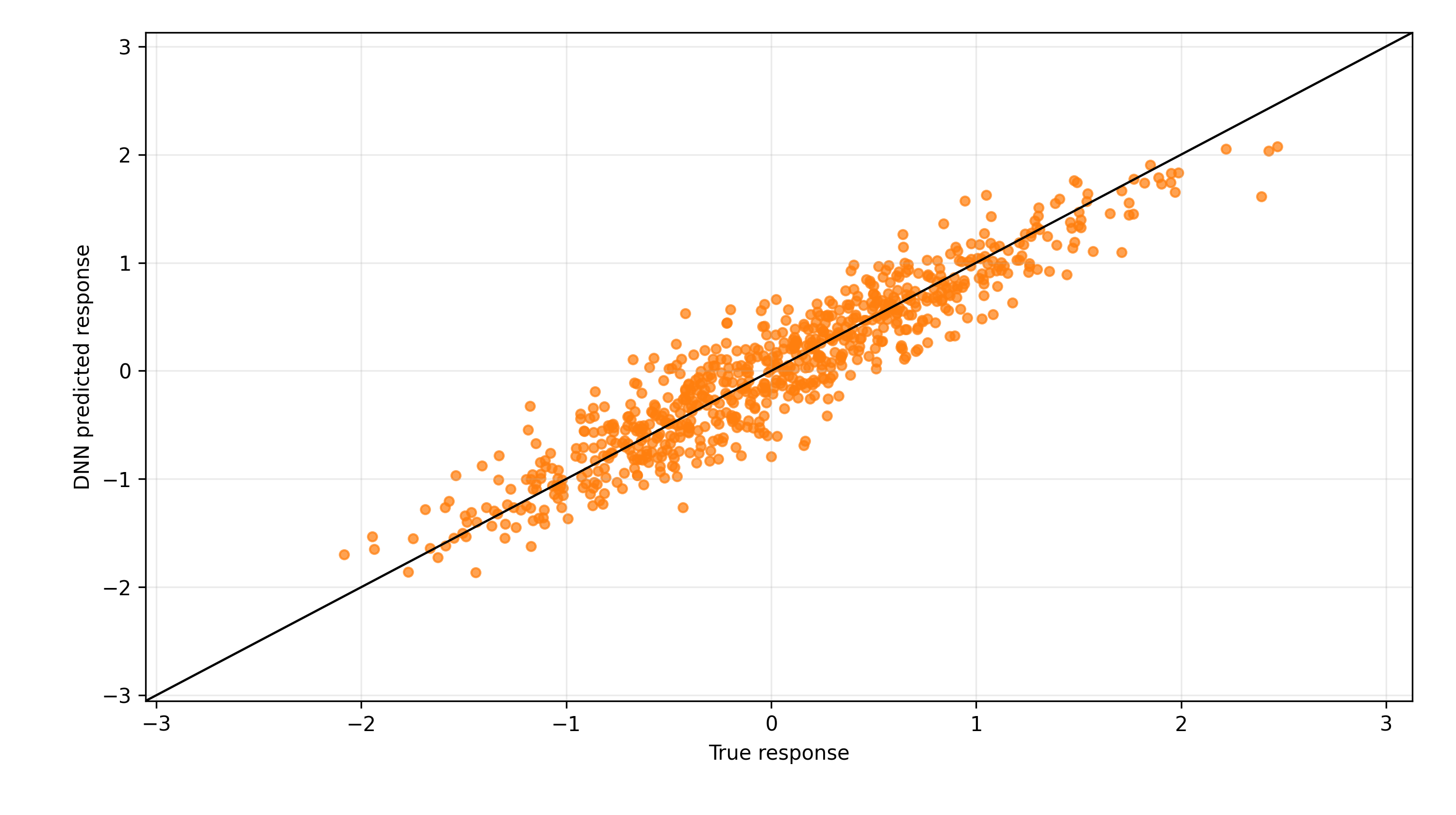}
        \caption{DNN predicted response versus true test response.}
    \end{subfigure}

    \caption{Synthetic experiment with 12 relevant inputs. Top row: post-burn-in ARD lengthscale chains under the three GP prior specifications. Middle row: GP predictive summaries. Bottom row: DNN predictive summaries. The DNN is included only as a predictive benchmark, since it does not have analogous ARD lengthscale chains.}
    \label{fig:rel12_synthetic_combined}
\end{figure}

Figure~\ref{fig:rel12_synthetic_combined} shows the 12-input setting. The Wishart prior still has the strongest predictive relationship to the true values, although the gain is more modest than in the more sparse cases. This matches the summary table.

\FloatBarrier

\begin{figure}[!htbp]
    \centering

    \begin{subfigure}[t]{0.32\textwidth}
        \centering
        \includegraphics[width=\linewidth]{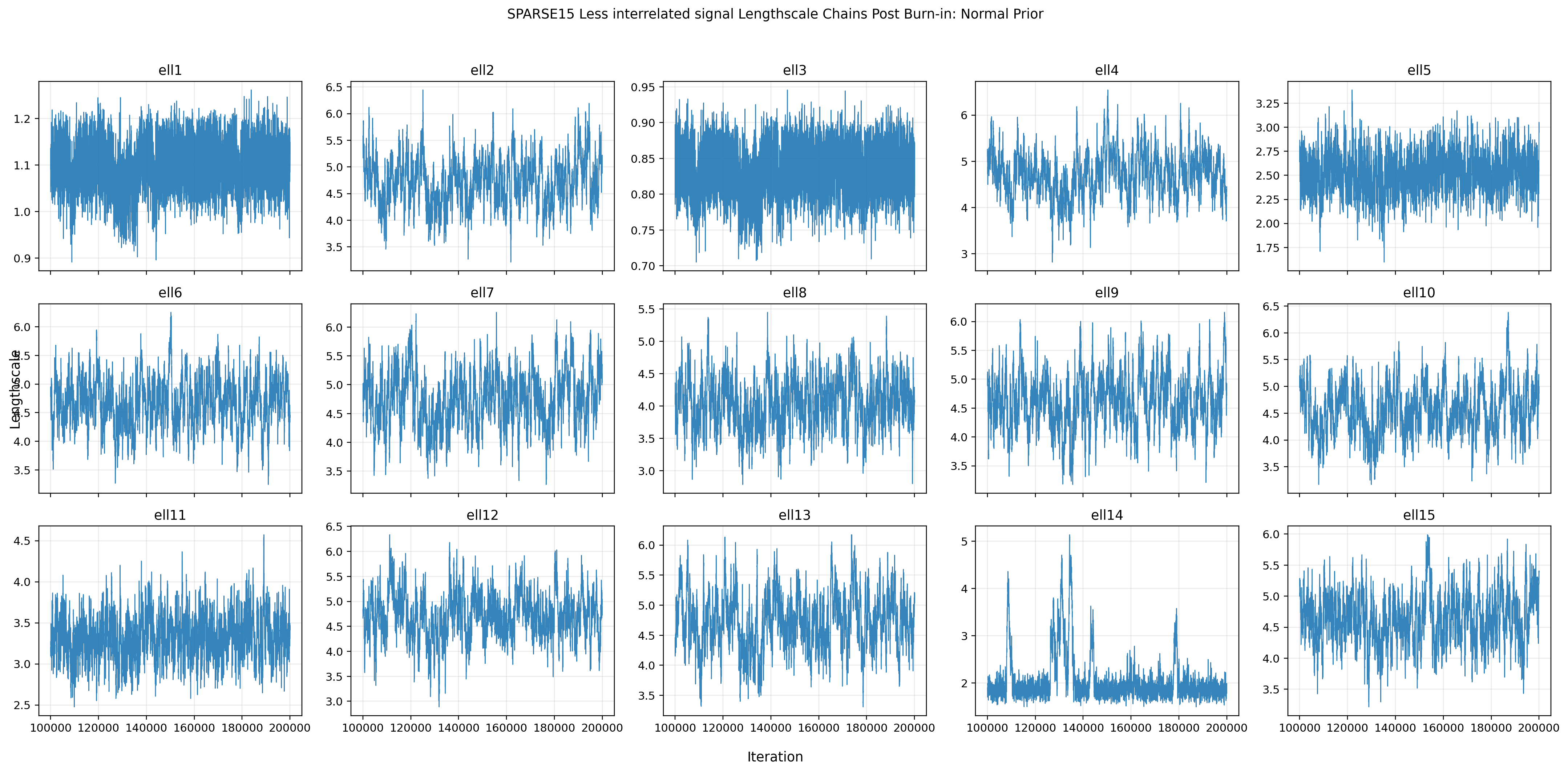}
        \caption{Normal prior.}
    \end{subfigure}\hfill
    \begin{subfigure}[t]{0.32\textwidth}
        \centering
        \includegraphics[width=\linewidth]{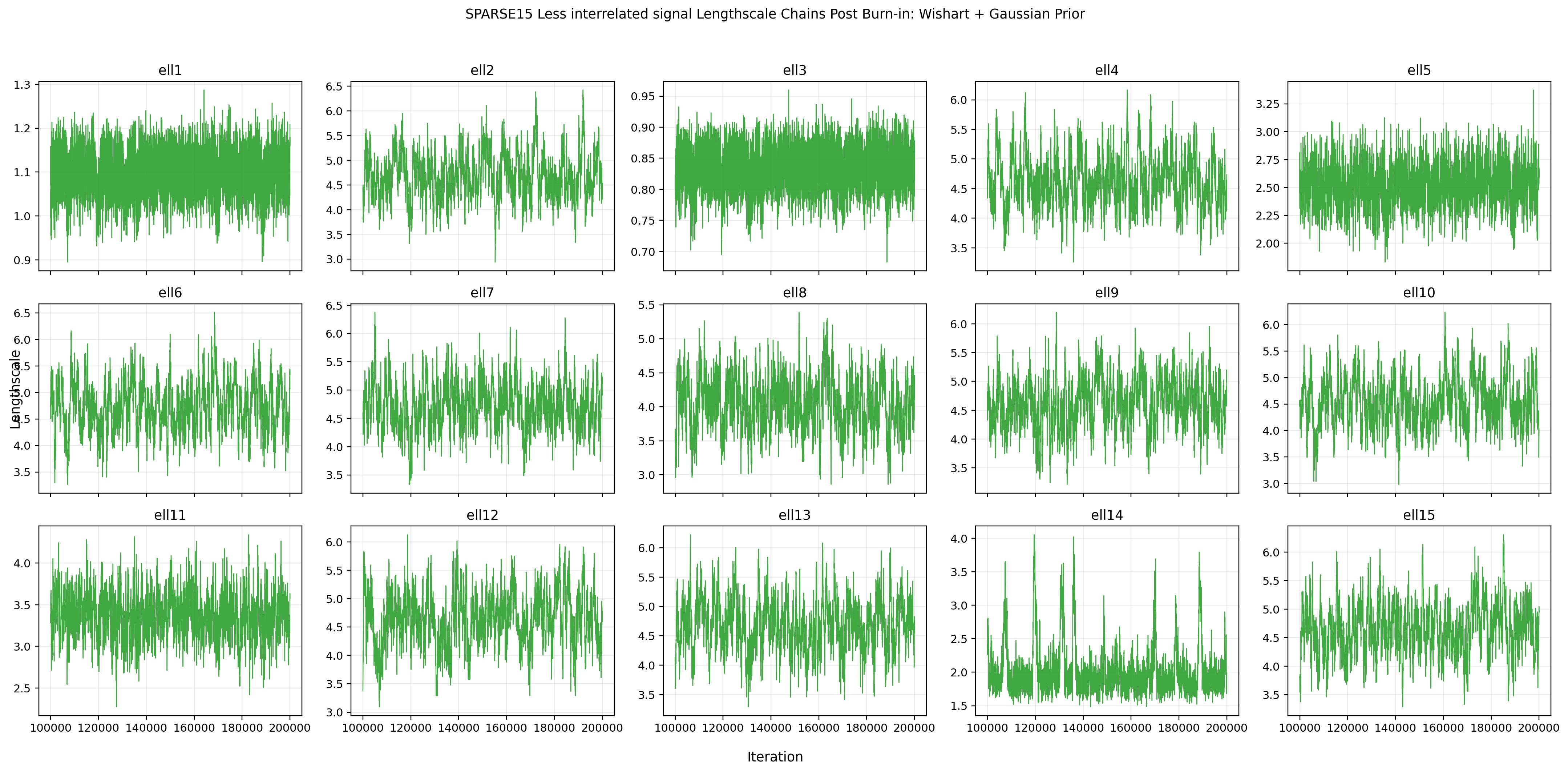}
        \caption{Wishart + Normal prior.}
    \end{subfigure}\hfill
    \begin{subfigure}[t]{0.32\textwidth}
        \centering
        \includegraphics[width=\linewidth]{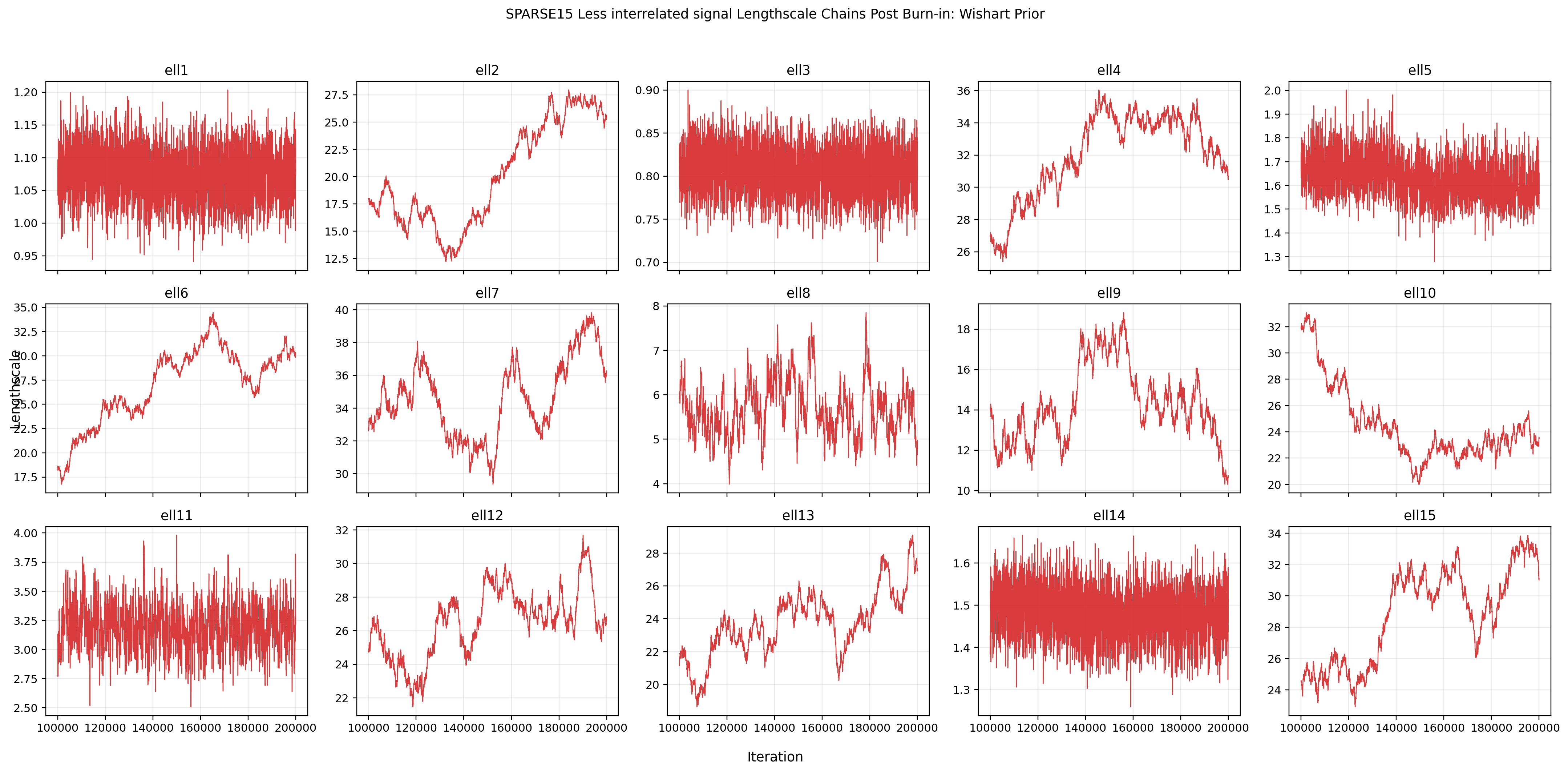}
        \caption{Wishart prior.}
    \end{subfigure}

    \vspace{0.8em}

    \begin{subfigure}[t]{0.49\textwidth}
        \centering
        \includegraphics[width=\linewidth]{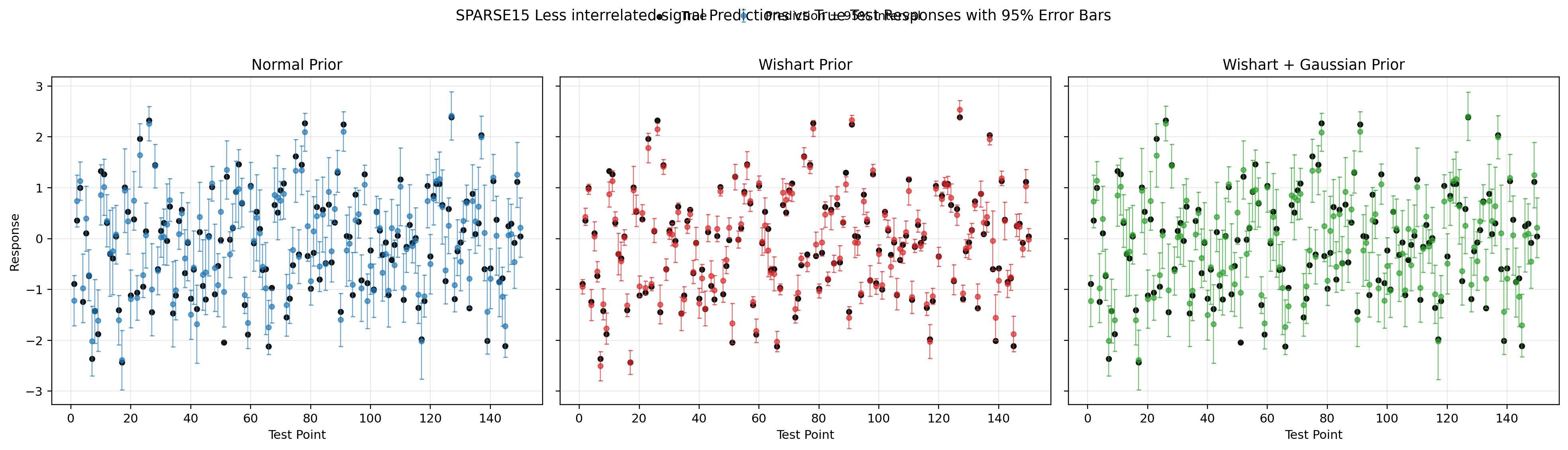}
        \caption{GP posterior predictive means against test-point index.}
    \end{subfigure}\hfill
    \begin{subfigure}[t]{0.49\textwidth}
        \centering
        \includegraphics[width=\linewidth]{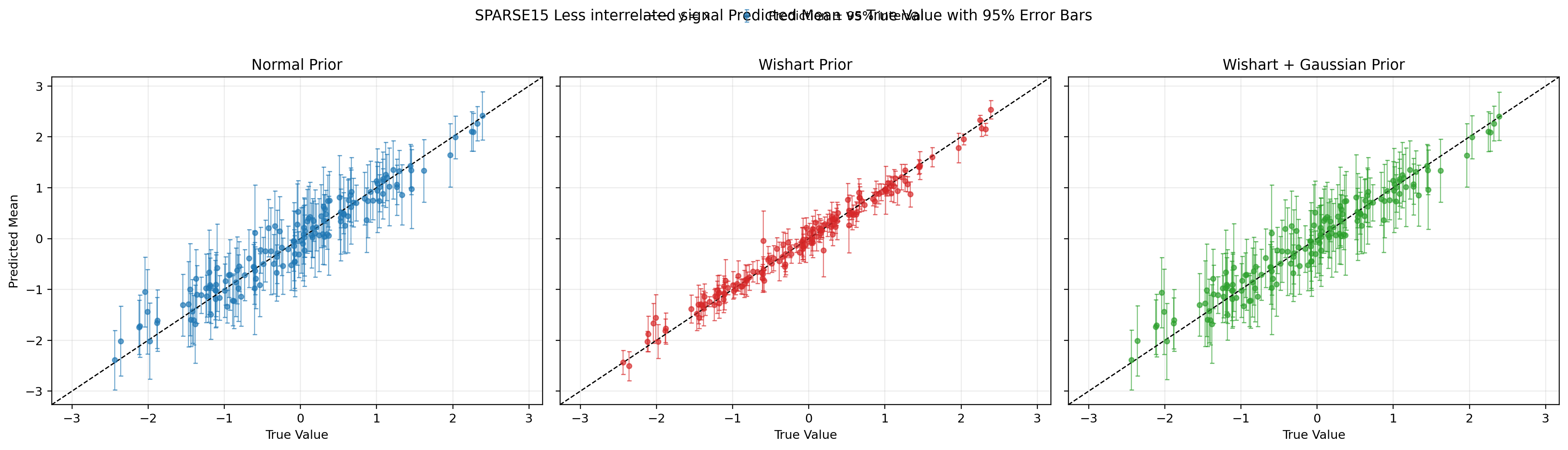}
        \caption{GP predicted mean versus true test response.}
    \end{subfigure}

    \vspace{0.8em}

    \begin{subfigure}[t]{0.35\textwidth}
        \centering
        \includegraphics[width=\linewidth]{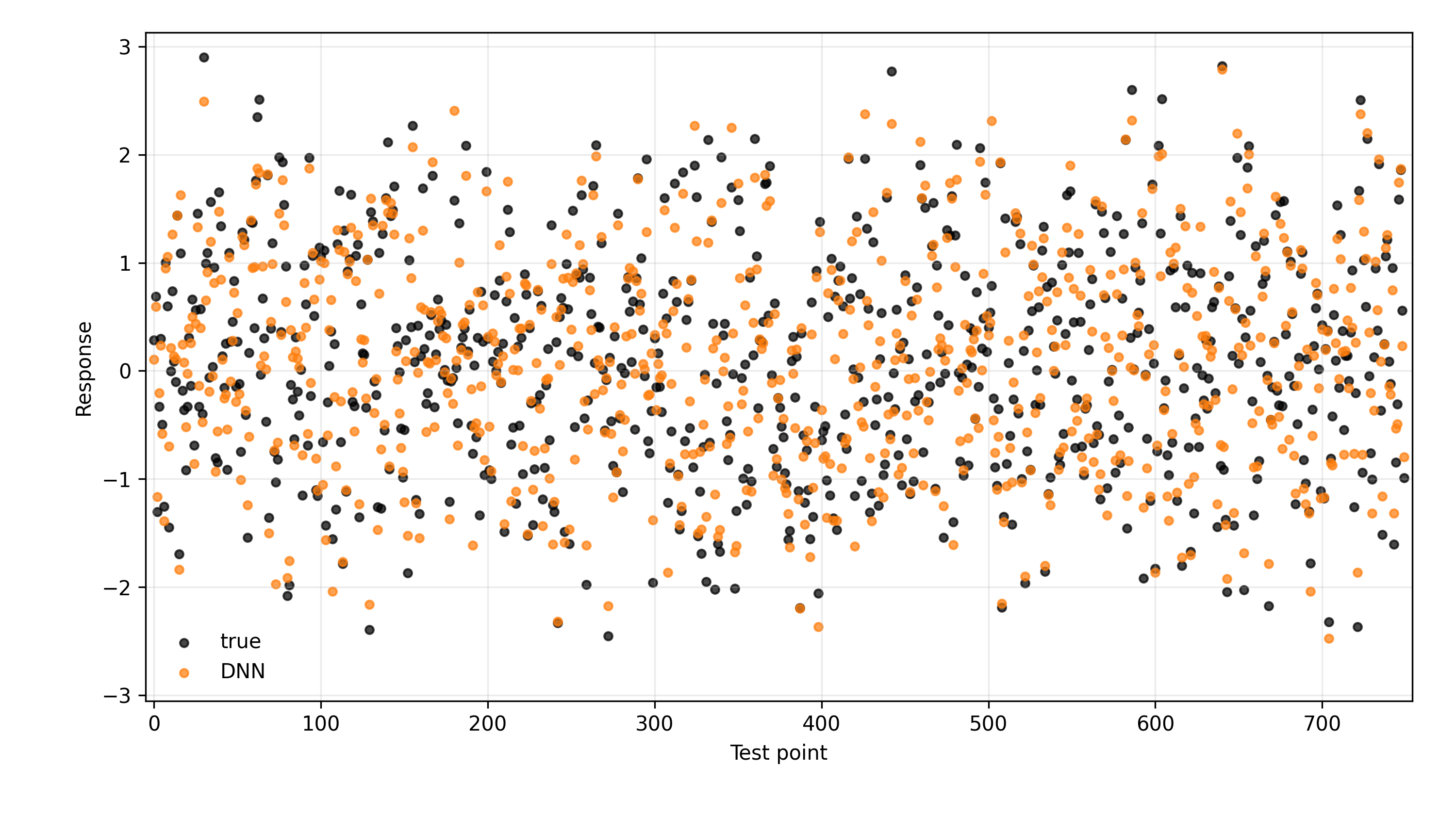}
        \caption{DNN predictions against test-point index.}
    \end{subfigure}\hfill
    \begin{subfigure}[t]{0.35\textwidth}
        \centering
        \includegraphics[width=\linewidth]{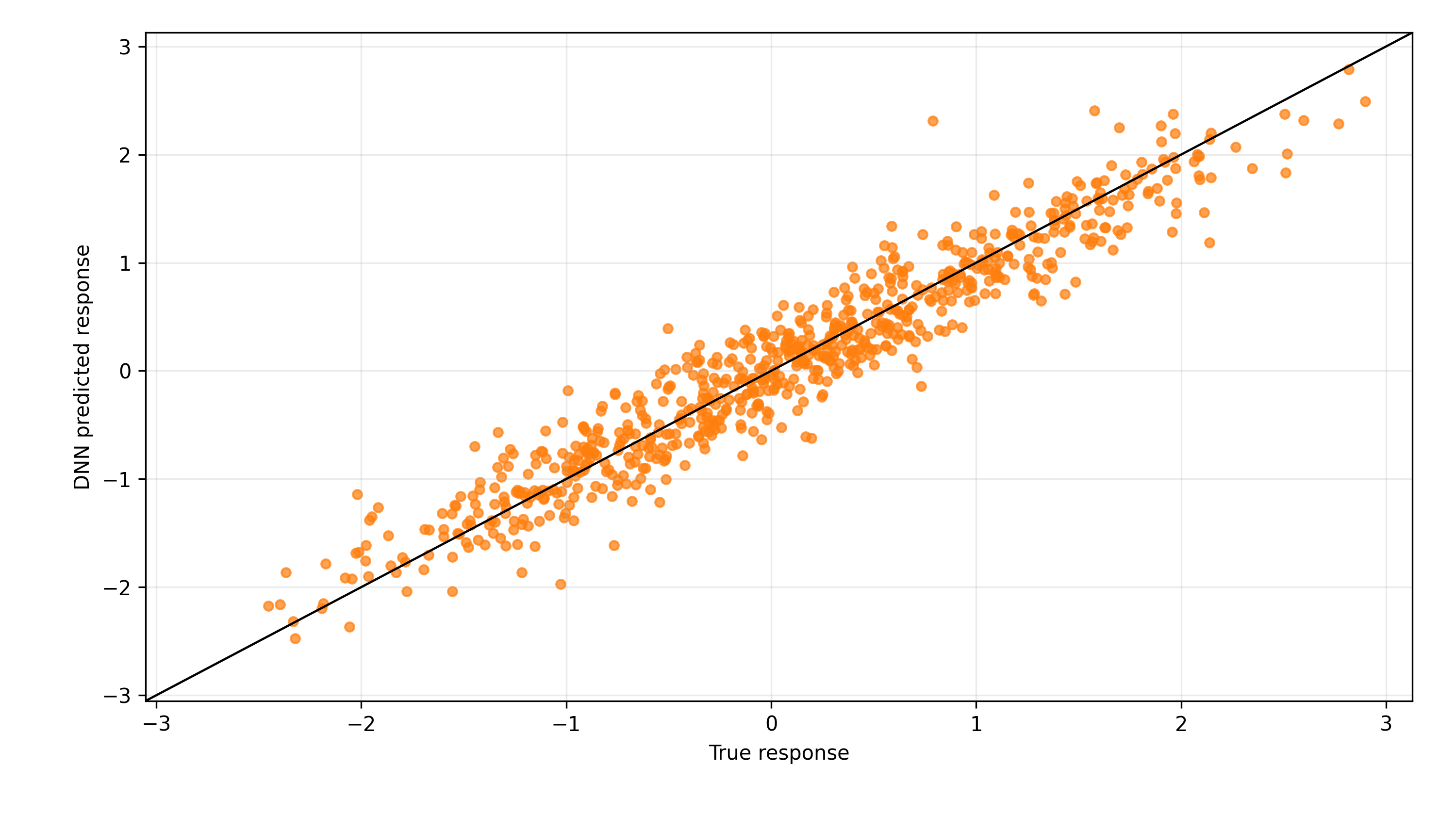}
        \caption{DNN predicted response versus true test response.}
    \end{subfigure}

    \caption{Less interrelated 6-input synthetic experiment. Top row: post-burn-in ARD lengthscale chains under the three GP prior specifications. Middle row: GP predictive summaries. Bottom row: DNN predictive summaries. The DNN is included only as a predictive benchmark, since it does not have analogous ARD lengthscale chains.}
    \label{fig:less_synthetic_combined}
\end{figure}

Figure~\ref{fig:less_synthetic_combined} is particularly useful because it shows that the Wishart prior is not only helping in the most interaction-heavy version of the problem. It still gives much stronger relevance separation and substantially better point prediction when the same sparse setup is made structurally simpler. At the same time, the gain is somewhat smaller than in the original baseline case, suggesting that signal complexity also plays a role.

\FloatBarrier

One feature of the nuisance-input lengthscale traces is that they need not show the same type of convergence as the relevant-input lengthscales. This is not interpreted here as a failure of the Wishart prior. The Wishart prior is placed on the induced covariance matrix, rather than directly on each individual lengthscale. Hence its role is to favour covariance structures that suppress uninformative inputs, not to force each nuisance hyperparameter to concentrate around a particular scalar value. Since the nuisance inputs carry essentially no signal for the response, the likelihood provides little information about their individual lengthscales once they are sufficiently large to remove the corresponding input from the covariance structure. As a result, different large nuisance lengthscale values can imply very similar predictive covariance behaviour, and the associated chains may continue to move within this weakly identified region. By contrast, the apparent convergence of nuisance lengthscales under the Normal and Wishart + Normal specifications is partly driven by the marginal hyperparameter prior itself: the prior supplies an individual scalar target for each lengthscale, even when the data provide little information about that nuisance input.

Overall, the synthetic experiments point in the same broad direction. All three GP prior specifications recover some relevance structure, but the Wishart prior does so much more strongly. Across all settings, its main effect is not to make the relevant inputs dramatically smaller than under the competing methods, but to push the nuisance inputs to much larger values. That sharper separation is then reflected in prediction. The DNN panels provide an additional predictive benchmark: they show that a flexible non-GP model can often track the response well, but without providing the same ARD lengthscale-based interpretation. The strongest gains do not appear to depend on sparsity alone: they vary with the complexity of the signal as well. The most natural interpretation is therefore that the Wishart prior helps most when the GP must simultaneously suppress weakly informative inputs and learn a more demanding covariance geometry.

\section{Real-data application: T\'etouan City energy data}

We now turn to a practical prediction problem using the T\'etouan City energy data. The aim in this section is not to revisit the controlled questions addressed by the synthetic experiments, but to see whether the covariance-level prior is useful in a realistic multi-input GP setting.

For both the full 5-input and reduced 4-input analyses, the Wishart-based specifications use the same look-back window, \(n=205\). The training set contains \(p=203\) observations, so this choice satisfies the non-degeneracy condition for the Wishart density while keeping the look-back window close to the number of training observations. The same implementation choice is used throughout the real-data application.

Our starting point is the fuller predictor set consisting of temperature, humidity, wind speed, general flows, and diffuse flows. We first fit the model using all five inputs, treating this as the initial predictive specification for the real-data problem. What makes this application interesting is that the covariance-level prior does not simply improve or degrade predictive accuracy uniformly: in the fuller analysis, it appears to single out one input as contributing relatively little useful smooth predictive structure under the assumed GP covariance model. This then motivates a second-stage analysis in which that input is removed and prediction is reassessed in a reduced 4-input model.

The real-data section therefore follows an applied modelling workflow. We begin with the fuller input set that one would naturally use for prediction, examine what the covariance-level prior does in that setting, and then use that behaviour to motivate a reduced-input analysis. In this sense, the T\'etouan application is useful not only because it provides a real-data test of predictive performance, but also because it shows that the covariance-level prior can act as a practical diagnostic device within GP learning.

\subsection{Full 5-input analysis}

We begin with the fuller predictor set
\[
\mathbf{x}=(X_1,X_2,X_3,X_4,X_5)^\top,
\]
where the five inputs are temperature, humidity, wind speed, general flows, and diffuse flows. As before, we use an ARD squared exponential kernel, so that each input component receives its own lengthscale. We compare three prior specifications:
\begin{enumerate}[leftmargin=1.5em]
    \item a Normal prior on the lengthscales only,
    \item a Wishart prior on the induced covariance matrix only,
    \item a combined Wishart + Normal prior.
\end{enumerate}

Table~\ref{tab:tetouan-all5-metrics} summarises predictive performance on the test set.

\begin{table}[!htbp]
\centering
\begin{tabular}{lccc}
\toprule
Prior & RMSE & MAE & 95\% coverage \\
\midrule
Normal & 8793.45 & 6573.86 & 0.88 \\
Wishart & 8126.84 & 6428.59 & 0.98 \\
Wishart + Normal & 8535.04 & 6777.72 & 0.98 \\
\bottomrule
\end{tabular}
\caption{Predictive performance on the T\'etouan City test set in the full 5-input analysis.}
\label{tab:tetouan-all5-metrics}
\end{table}

In the full 5-input problem, the Wishart prior gives the strongest predictive performance. It attains the lowest RMSE and the lowest MAE, while also achieving the highest empirical coverage, tied with the combined Wishart + Normal specification. This provides an initial indication that the covariance-level prior is practically useful in the real-data setting, rather than merely reproducing the qualitative behaviour seen in the synthetic experiments.

\begin{figure}[!htbp]
    \centering
    \begin{subfigure}[t]{0.32\textwidth}
        \centering
        \includegraphics[width=\linewidth]{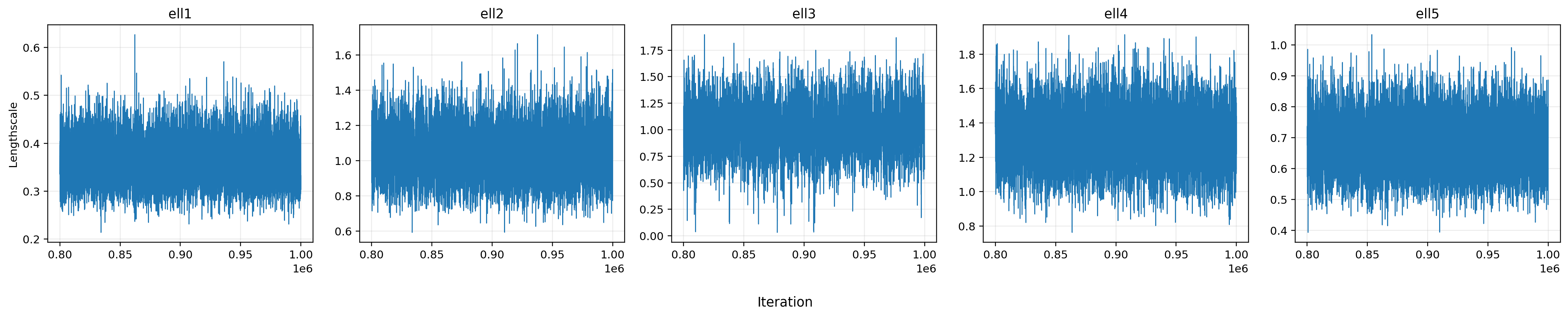}
        \caption{Normal prior.}
    \end{subfigure}\hfill
    \begin{subfigure}[t]{0.32\textwidth}
        \centering
        \includegraphics[width=\linewidth]{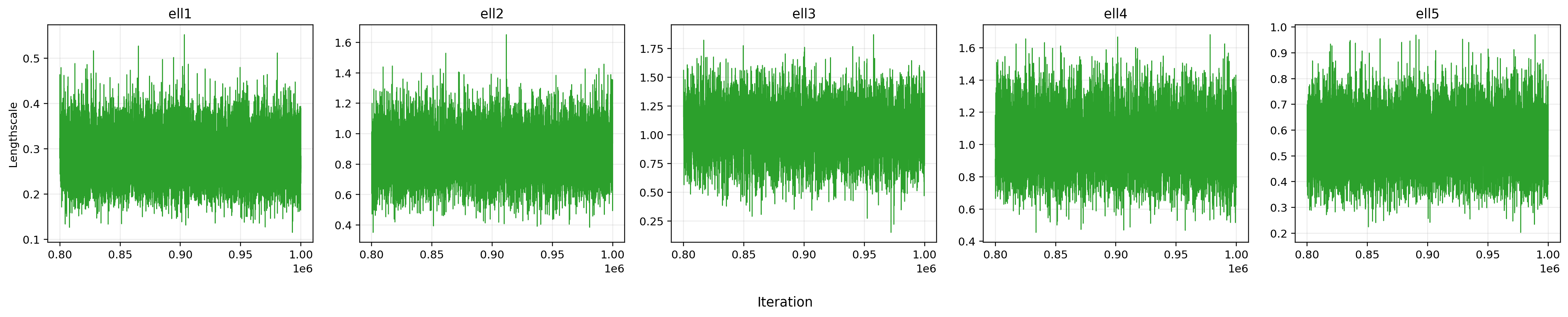}
        \caption{Wishart + Normal prior.}
    \end{subfigure}\hfill
    \begin{subfigure}[t]{0.32\textwidth}
        \centering
        \includegraphics[width=\linewidth]{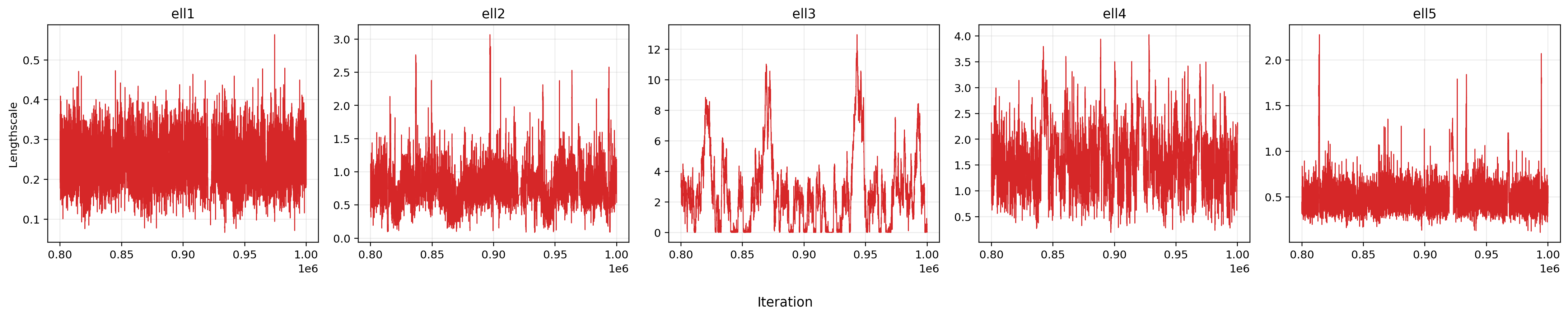}
        \caption{Wishart prior.}
    \end{subfigure}

    \vspace{0.8em}

    \begin{subfigure}[t]{0.49\textwidth}
        \centering
        \includegraphics[width=\linewidth]{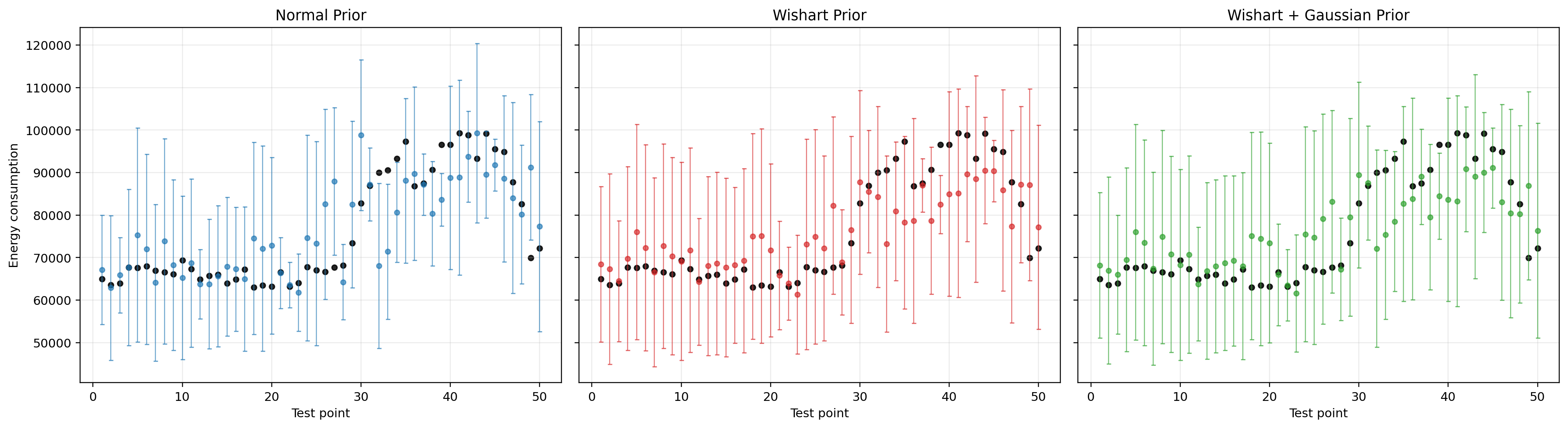}
        \caption{Posterior predictions against test-point index.}
    \end{subfigure}\hfill
    \begin{subfigure}[t]{0.49\textwidth}
        \centering
        \includegraphics[width=\linewidth]{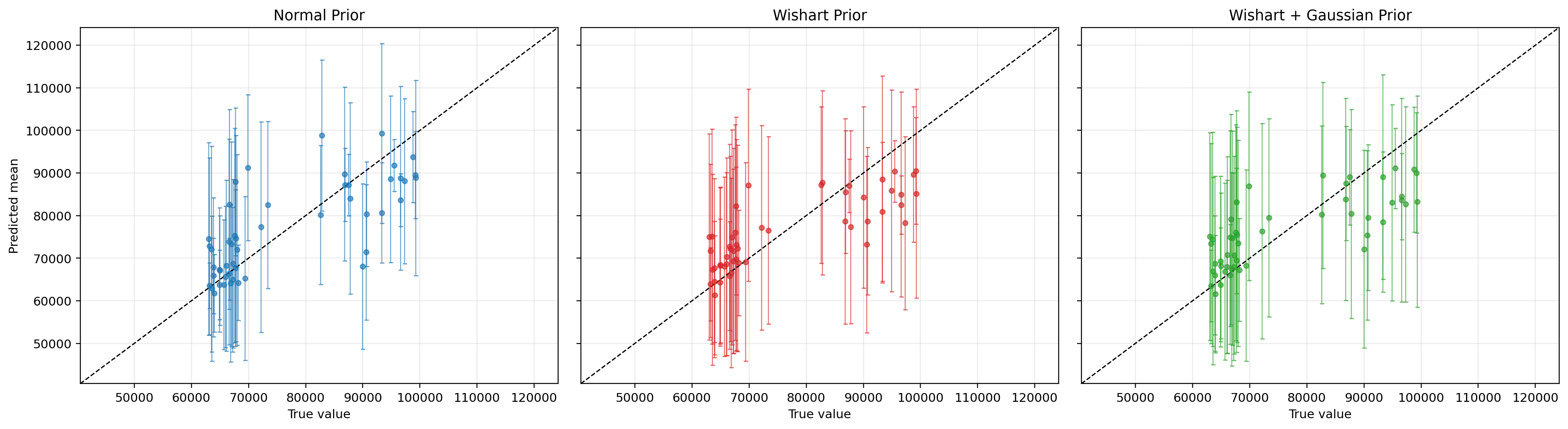}
        \caption{Predicted mean versus true test response.}
    \end{subfigure}

    \caption{Full 5-input T\'etouan analysis. Top row: post-burn-in ARD lengthscale chains under the three prior specifications. Bottom row: predictive summaries for the same analysis.}
    \label{fig:tetouan-all5-combined}
\end{figure}

Figure~\ref{fig:tetouan-all5-combined} summarises the full 5-input analysis. The top row shows the post-burn-in ARD lengthscale chains under the three prior specifications, while the bottom row shows the two predictive summaries. Under the Normal and Wishart + Normal priors, the learnt lengthscales remain comparatively concentrated and broadly similar in scale. Under the Wishart prior, by contrast, one component is treated much more distinctly, with substantially larger excursions and much weaker concentration than the others. This suggests that the covariance-level prior is reacting sharply to an input that is only weakly informative under the assumed smooth GP covariance structure. The predictive panels are consistent with the same interpretation: the Wishart prior follows the higher-response part of the test set more closely than the other two specifications and appears more closely aligned with the diagonal in the predicted-versus-true comparison.

\FloatBarrier

Taken together, the full 5-input results suggest that the covariance-level prior is doing more than simply shrinking all inputs in a uniform way. Instead, it appears to be identifying an input that contributes relatively little useful smooth predictive structure and downweighting it much more strongly than a hyperparameter-level prior alone. This observation motivates the reduced-input analysis that follows.

\subsection{Reduced 4-input analysis}

The full 5-input analysis suggests that one input is only weakly informative under the assumed smooth GP covariance structure. This motivates a second-stage analysis in which that input is removed, leaving temperature, humidity, general flows, and diffuse flows:
\[
\mathbf{x}=(X_1,X_2,X_4,X_5)^\top.
\]
As before, we use an ARD squared exponential kernel so that each input component is assigned its own lengthscale.

Table~\ref{tab:tetouan-reduced-metrics} summarises predictive performance under the three prior specifications. In this reduced-input analysis, coverage is reported through \(1\sigma\) and \(2\sigma\) predictive compatibility rather than empirical 95\% interval coverage.

\begin{table}[!htbp]
  \centering
  \small
  \begin{tabular}{lcccc}
    \toprule
    Prior & RMSE & MAE & \(\mathrm{cov}_{1\sigma}(\%)\) & \(\mathrm{cov}_{2\sigma}(\%)\) \\
    \midrule
    Wishart & 8510.840 & 6344.578 & 70.00 & 92.00 \\
    Wishart + Normal & 7851.099 & 5973.490 & 72.00 & 96.00 \\
    Normal & 7915.140 & 6006.304 & 72.00 & 94.00 \\
    \bottomrule
  \end{tabular}
  \caption{Predictive performance on the T\'etouan City test set in the reduced 4-input analysis. Coverage is the percentage of test points whose observed value lies within \(\hat{y}\pm k\hat{\sigma}\) for \(k\in\{1,2\}\).}
  \label{tab:tetouan-reduced-metrics}
\end{table}

The reduced-input results tell a different story from the full 5-input analysis. Once the weakly informative input has been removed, the Wishart-only prior is no longer the strongest practical specification. Instead, the combined Wishart + Normal prior gives the strongest overall predictive performance, with the lowest RMSE and MAE together with the highest \(2\sigma\) predictive compatibility. The Normal prior remains competitive, but the combined prior gives the best overall balance across the reported metrics.

This makes the reduced-input results especially useful for interpretation. The full 5-input problem suggested that the covariance-level prior was identifying and downweighting a weakly informative input. Once that input is removed, the combined prior becomes the most attractive practical specification: it retains covariance-level structure while avoiding the less regular behaviour of the Wishart-only model.

\begin{figure}[!htbp]
    \centering
    \begin{subfigure}[t]{0.32\textwidth}
        \centering
        \includegraphics[width=\linewidth]{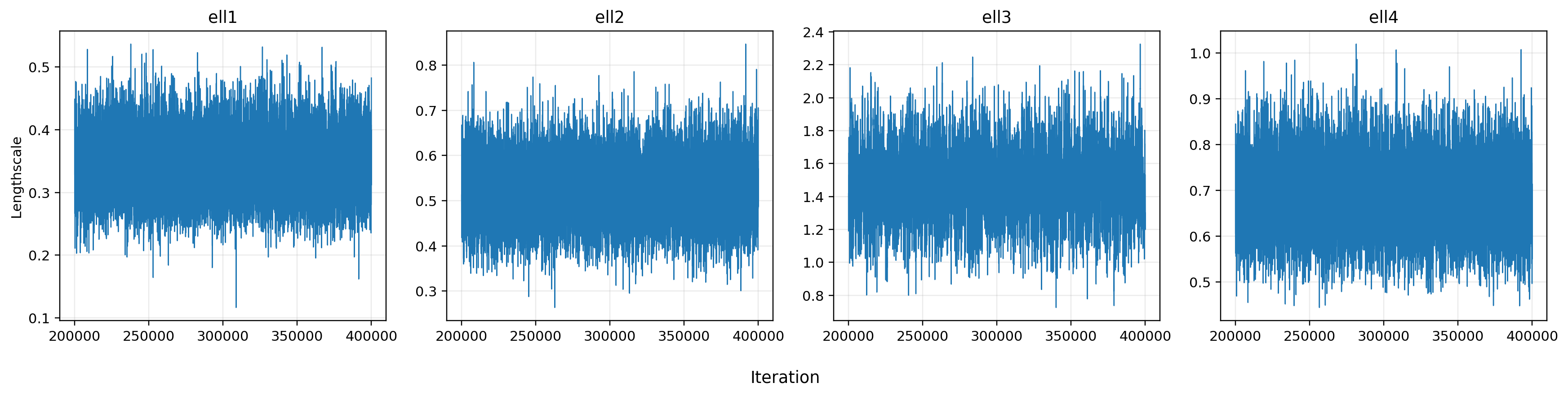}
        \caption{Normal prior.}
    \end{subfigure}\hfill
    \begin{subfigure}[t]{0.32\textwidth}
        \centering
        \includegraphics[width=\linewidth]{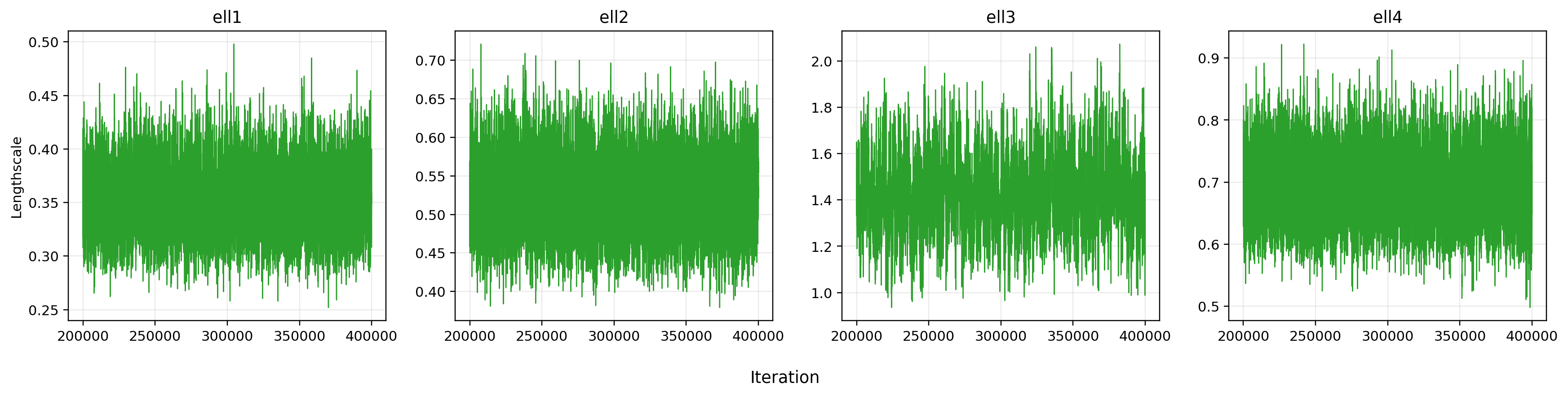}
        \caption{Wishart + Normal prior.}
    \end{subfigure}\hfill
    \begin{subfigure}[t]{0.32\textwidth}
        \centering
        \includegraphics[width=\linewidth]{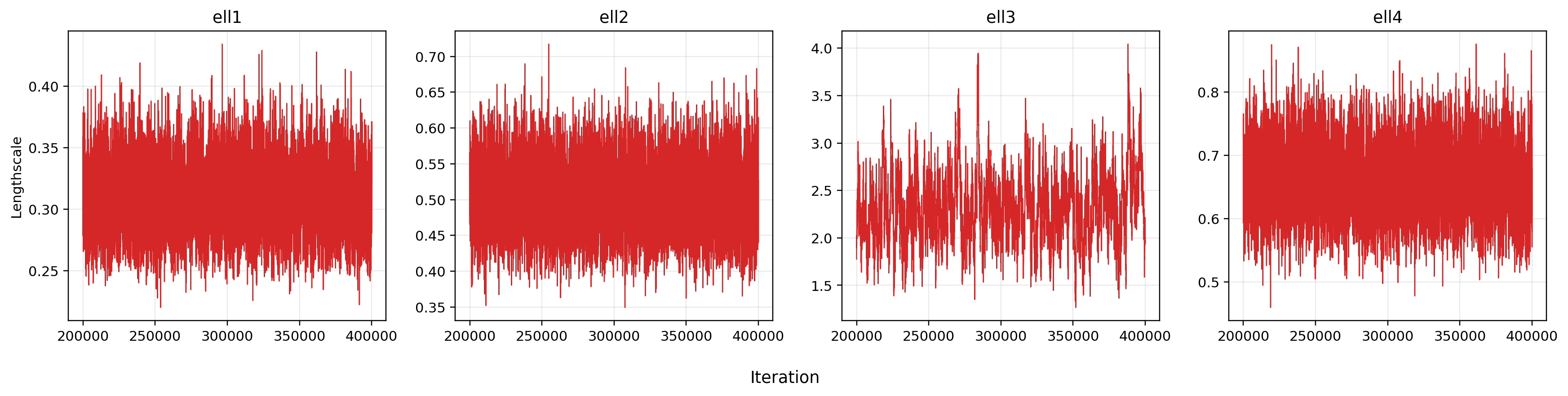}
        \caption{Wishart prior.}
    \end{subfigure}

    \vspace{0.8em}

    \begin{subfigure}[t]{0.49\textwidth}
        \centering
        \includegraphics[width=\linewidth]{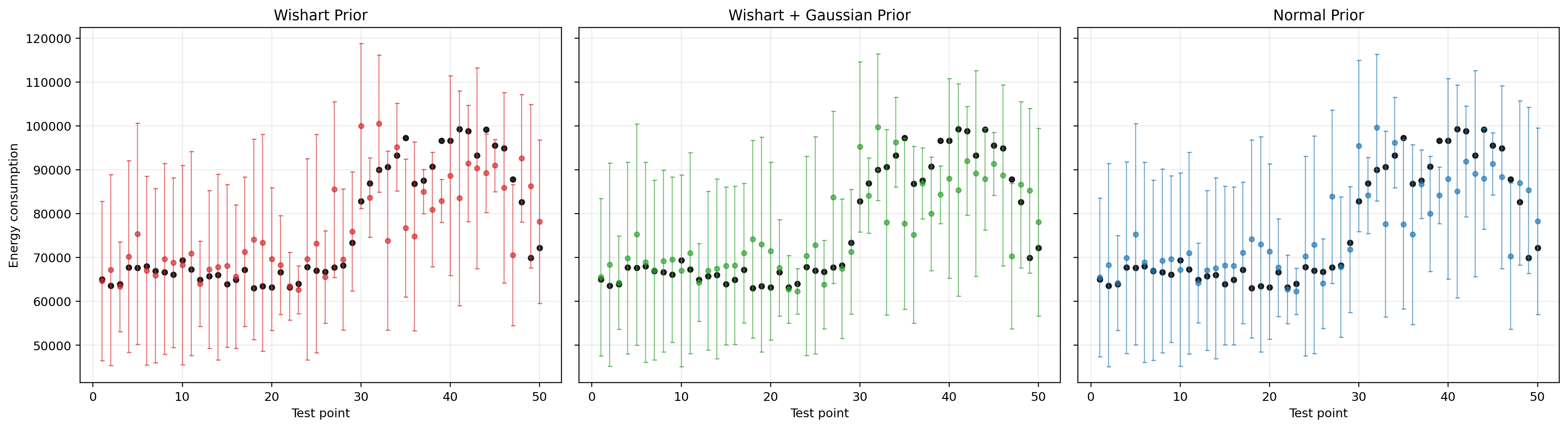}
        \caption{Posterior predictions against test-point index.}
    \end{subfigure}\hfill
    \begin{subfigure}[t]{0.49\textwidth}
        \centering
        \includegraphics[width=\linewidth]{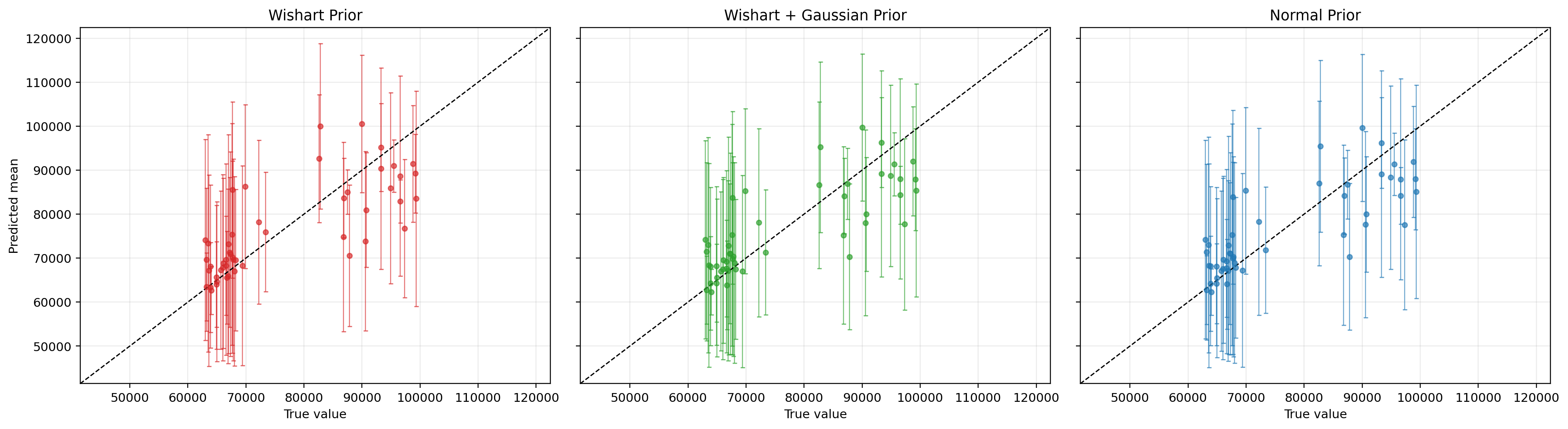}
        \caption{Predicted mean versus true test response.}
    \end{subfigure}

    \caption{Reduced 4-input T\'etouan analysis. Top row: post-burn-in ARD lengthscale chains under the three prior specifications. Bottom row: predictive summaries for the same analysis.}
    \label{fig:tetouan-all4-combined}
\end{figure}

Figure~\ref{fig:tetouan-all4-combined} summarises the reduced-input analysis. The predictive panels are consistent with the same story. The combined specification tracks the higher-response region more convincingly than the Wishart-only fit, while remaining competitive with the Normal prior. This is especially clear in the predicted-versus-true comparison, where the combined model gives the strongest overall behaviour once the weakly informative input has been removed.

\FloatBarrier

\subsection{Interpretation}

The two T\'etouan analyses play different roles. In the full 5-input problem, the Wishart prior appears to act as a diagnostic device, sharply downweighting an input that is only weakly informative under the smooth GP covariance structure and producing the strongest predictive performance. In the reduced 4-input problem obtained after removing that input, predictive performance improves across specifications, and the combined Wishart + Normal prior becomes the most attractive overall model.

This distinction is important. The real-data results do not suggest that the Wishart-only prior is universally the best final specification. Rather, they suggest that covariance-level prior structure can be useful in two related ways. First, it can reveal inputs that are weakly informative under the assumed GP covariance geometry. Second, once such inputs are removed, combining covariance-level structure with a hyperparameter-level prior can produce a more stable and more practically effective model for the remaining inputs.

\section{Conclusion}

This paper has developed a self-assembled Wishart prior for GP hyperparameter learning. The motivation was that, although priors in GP models are usually placed directly on kernel hyperparameters, the object that governs likelihood evaluation, numerical stability, and prediction is the covariance matrix induced by those hyperparameters at the observed design points. The proposed construction therefore places prior structure directly on this induced covariance matrix, while retaining the usual kernel-parametric form of the GP model.

The prior is self-assembled in the sense that its scale matrix is not fixed in advance, but is built from a look-back window of recently accepted hyperparameter values. This produces an iteration-dependent covariance-level prior structure: each proposed hyperparameter value induces a candidate covariance matrix, and the Wishart prior is evaluated directly on that matrix. The construction therefore provides a way to introduce covariance-level information into Bayesian GP hyperparameter inference without replacing the kernel model by an unrestricted covariance estimator.

The synthetic experiments show that this covariance-level prior can behave very differently from priors placed only on individual hyperparameters. Across sparse nonlinear problems with automatic relevance determination, the Wishart prior produced substantially stronger separation between relevant and irrelevant input dimensions. The main mechanism was not dramatic contraction of relevant-input lengthscales, but substantial inflation of nuisance-input lengthscales. This sharper separation was accompanied by improved predictive accuracy across the synthetic settings considered here, while repeated-seed experiments showed that the baseline pattern was stable. Additional synthetic variants indicated that the gain is not explained by sparsity alone, but also depends on the complexity of the underlying signal and the covariance geometry that the GP must learn.

The T\'etouan City energy application showed a complementary role for the same covariance-level behaviour. In the full 5-input analysis, the Wishart prior gave the strongest predictive performance and appeared to identify an input that was only weakly informative under the assumed smooth GP covariance structure. After removing that input, predictive performance improved across specifications, and the combined Wishart + Normal prior gave the strongest overall behaviour in the reduced-input problem. These results suggest that covariance-level prior structure can be useful not only for suppressing nuisance inputs, but also as a diagnostic tool for identifying inputs that contribute little useful smooth predictive structure.

Taken together, the results support the view that prior specification directly on the induced covariance matrix is a useful complement to conventional hyperparameter-level prior specification in higher-dimensional GP learning. The approach does not seek greater flexibility by changing the kernel class itself; instead, it retains the kernel-parametric model while changing where prior information is imposed. This distinction is important in multi-input settings, where many hyperparameters must be learned simultaneously and where weakly informative inputs can distort the covariance structure that ultimately determines inference and prediction.

There are several natural directions for future work. One is to investigate alternative choices of look-back window, degrees of freedom, and scale-matrix construction, since these choices determine how strongly the prior responds to recently accepted covariance structures. Another is to explore the method beyond the squared exponential ARD setting, including other kernel families and larger input spaces. Finally, the covariance-level behaviour observed here suggests that the approach may be useful in applications where input relevance, nuisance-input suppression, and covariance-level diagnostics are important for model reduction. Overall, the empirical results indicate that acting directly on the induced covariance matrix can provide a practical and interpretable route to more effective GP hyperparameter learning.

\clearpage
\bibliographystyle{plainnat}
\bibliography{references}

\end{document}